%% file: neurips_2024.tex
\documentclass{article}

 \PassOptionsToPackage{numbers, compress}{natbib}
     
\usepackage[final]{neurips_2024}

\usepackage[utf8]{inputenc} 
\usepackage[T1]{fontenc}    
\usepackage[hidelinks]{hyperref}       
\usepackage{url}            
\usepackage{booktabs}       
\usepackage{amsfonts}       
\usepackage{nicefrac}       
\usepackage{microtype}      
\usepackage{xcolor}         
\usepackage{bm} 
\usepackage{algorithm}
\usepackage{algpseudocode}
\usepackage{algorithmicx}
\usepackage{amsmath,amsfonts,amssymb,amsthm}

\usepackage{wrapfig,lipsum,booktabs}
\usepackage{subcaption}

\usepackage{lmodern}
\usepackage{adjustbox}
\usepackage{multicol}

\usepackage{todonotes}

\usetikzlibrary{bayesnet}
\usetikzlibrary{arrows.meta}

\input{figures/tikz_graphs}

\usepackage{multirow}
\usepackage{geometry}
\usepackage{colortbl}  
\usepackage{makecell}  
\usepackage{booktabs}  
\definecolor{headercolor}{rgb}{0.8, 0.9, 1.0} 

\allowdisplaybreaks

\usepackage{thmtools}
\usepackage{thm-restate}
\usetikzlibrary{decorations.pathreplacing}

\usepackage{booktabs}

\newtheorem{definition}{Definition}[section]

\newcommand{\infdiv}[2]{D_{\text{KL}}\left(#1 \, \| \, #2 \right)}

\title{Adaptive World Models: Learning Behaviors by Latent Imagination Under Non-Stationarity}

\author{
  Emiliyan Gospodinov\textsuperscript{\normalfont 1}\thanks{Corresponding author. Email to <gospodinov.emilian@gmail.com>} \hspace{2pt}
  Vaisakh Shaj\textsuperscript{\normalfont 1}  \hspace{2pt}
  Philipp Becker\textsuperscript{\normalfont 1}  \hspace{2pt} 
  Stefan Geyer\textsuperscript{\normalfont 2}\\
  \textbf{Gerhard Neumann}\textsuperscript{\normalfont 1}\\
  \textsuperscript{\normalfont 1}Karlsruhe Institute of Technology (KIT), Germany \\
  \textsuperscript{\normalfont 2}Institute for Artificial Intelligence, Stuttgart, Germany \\
}

\begin{document}

\maketitle

\begin{abstract}
Developing foundational world models is a key research direction for embodied intelligence, with the ability to adapt to non-stationary environments being a crucial criterion. In this work, we introduce a new formalism, Hidden Parameter-POMDP, designed for control with adaptive world models. We demonstrate that this approach enables learning robust behaviors across a variety of non-stationary RL benchmarks. Additionally, this formalism effectively learns task abstractions in an unsupervised manner, resulting in structured, task-aware latent spaces.  
\end{abstract}

\section{Introduction}

Recent advances in foundational models have achieved remarkable success in NLP and vision tasks~\cite{openai2024gpt4technicalreport, rombach2022highresolutionimagesynthesislatent}. Still, they fall short in addressing the complexities faced by embodied agents in dynamic, real-world environments. For embodied intelligence, we argue that it is essential to develop foundational world models~\cite{gupta2024essential,shaj2024learning} that capture the causal nature of the world we live in and can make counterfactual predictions. Furthermore, these models should adapt dynamically to non-stationary environments.

Current state-of-the-art approaches for Model-Based Reinforcement Learning (MBRL)~\cite{hafner2019dream,hafner2019learning,hansen2022temporal} often use probabilistic state-space models~\cite{krishnan2015deep,becker2019recurrent,shaj2021action} as a backbone. They learn behaviors by making counterfactual predictions in the latent space of world models. Often these approaches focus on agents mastering a specific, narrow task. Throughout this work, a "\textbf{task}" refers to a particular schema of environment dynamics or a specific reward function. In real-world settings, however, tasks are frequently non-stationary and subject to change over time. Thus, a truly intelligent agent must (1) understand the current task and (2) dynamically adapt its perception, model, and behavior to new tasks with minimal interaction.

We identified a gap in the literature regarding MBRL in latent spaces that address multitask learning and adaptation under non-stationarity. This work makes two key contributions: (1) highlighting the limitations of current state-of-the-art model-based agents in non-stationary settings, and (2) proposing a new formalism that models non-stationarity as an additional causal latent variable, resulting in robust policies.

\section{Non-Stationary RL Formalisms In Latent Spaces}

\subsection{POMDP formalism}
Existing state-of-the-art MBRL agents~\cite{hafner2019dream,hafner2019learning,hafner2021mastering,hansen2022temporal,hansen2024tdmpc} that learn in latent spaces typically rely on the partially observable Markov decision process (POMDP) formalism. In this framework, incoming sensory signals are used to update the agent’s belief about the hidden state of the environment, enabling the agent to make decisions under uncertainty. Theoretically, the POMDP formalism could handle non-stationarity by treating slowly changing, unobserved tasks as part of the latent states~\cite{xie2020deep}. Here the assumption is that the underlying environment is assumed to be stationary, but the agent has an incomplete view of it~\cite{khetarpal2022towards}. Consequently, single-task frameworks that rely on latent dynamics models for learning should, in theory, be applicable in streaming settings. However, an unstructured latent state, without inductive biases, may hinder the learning of sample-efficient adaptive policies. 

\subsection{HiP-POMDP formalism}
A complementary but more popular view in literature for non-stationary RL is that the components of the RL (transition, reward, observation functions, action space, etc.) may depend upon time~\cite{khetarpal2022towards}. We build our formalism, the HiP-POMDP, upon this non-stationary function view~\cite{doshi2016hidden,zhang2020learning,xie2020deep,shaj2022hidden,costen2023planning,shaj2023multi}. We start by providing a formal definition of a HiP-POMDP and demonstrate that this simple modification, along with a scalable variational inference scheme, enables learning adaptive policies across a wide range of non-stationary scenarios where the changing tasks are unknown to the agent.
\begin{definition}
A HiP-POMDP is given by a tuple $$\lbrace \mathcal{S}, \mathcal{A}, \mathcal{O}, \mathcal{C}, \mathcal{L}, p_s(\bm{s}_{t+1} | \bm{a}_{t}, \bm{s}_{t}, \bm{l}), p_o(\bm{o}_t| \bm{s}_t, \bm{l}), r(\bm{s}_t,\bm{a}_t,\bm{l}), p_c(\mathcal{C}_l | \bm{l}) \rbrace,$$ 
where $\mathcal{S}$, $\mathcal{A}$, and $\mathcal{O}$ are the standard state, action and observation spaces. 
Additionally, we introduce a space of latent task variables $\mathcal{L}$, where $\bm{l} \in \mathcal{L}$, and a space of context observations $\mathcal{C}$, where $\mathcal{C}_l \in \mathcal{C}$. Context observations $\mathcal{C}_l$  are generated from $\bm{l}$ according to $p(\mathcal{C}_l | \bm{l})$. Finally, the transition model $p_s$, observation model $p_o$, and the reward function $r$ all depend on the latent task $l$.   
\end{definition}

This general definition does not specify how exactly the context can be observed. 
Throughout this work, we assume $\mathcal{C}_l = \{ \left( \bm{o}, \bm{a}, r, \bm{o}^{\prime} \right)_n \}_{n=1}^{N}$, i.e., a set of N recent transitions. However, more expressive $\mathcal{C}_l$ including temporal embeddings, task metadata, or any other available information about the task could be used in the future. 
In a  HiP-POMDP, the agent's objective is to infer a latent task distribution $p(\bm{l} \mid \bm{C}_{l})$ based on the context observations $\bm{C}_{l}$ and learn a latent task conditional policy $\pi(\bm{a}_t \mid \bm{s}_t, \bm{l})$ that maximizes the expected cumulative discounted reward, $
\mathbb{E}_{\pi}\left[\sum_{t=0}^{T-1} \gamma^t r_{t+1}\right],$
where $T$ is the total number of time steps and $\gamma \in [0, 1]$ is the discount factor.

\section{Adaptive Latent Space Models for HiP-POMDPs}
In line with standard practices in model-based reinforcement learning (MBRL), we alternate between representation learning, behavior learning, and environment interactions to learn policies in the latent space of a world model. However, unlike existing approaches, we make each of these stages adaptive by conditioning them on an inferred task representation or abstraction. Thus, our work goes in the direction of building foundational multi-task world models and subsequent behavior policies.
For efficient learning and inference, we adopt a two-phase approach, where we first infer the latent task, which we then use to condition the model, actor, and critic. 

\subsection{Inferring Latent Task Abstractions via Aggregation}
\label{subsec:task-infer}
As stated, throughout this work, we choose $\bm{C_l}$ to be a collection of N recently observed transition tuples $\{ (\bm{o},  \bm{a}, r, \bm{o}^{\prime})_n\}_{n=1}^N$, practically implemented as a FIFO buffer.
Note that we chose these tuples of observations, actions, rewards, and next observations because they worked well in practice, capturing sufficient task-relevant statistics as shown in Section 4. 

Now, to form the posterior belief over the latent task variable $\bm{l}$, we first extract encoded representations $\bm{x}_n$ with associated variances $\bm{\sigma}_n$ from each transition tuple in the context set using a set encoder network with shared parameters. 
We assume the latent representation is distributed according to $ \mathcal{N}\left( \bm{x}_n \mid \bm{l}, \textrm{diag}\left( \bm{\sigma}_l \right) \right)$.
This assumption allows us to form Gaussian beliefs $\mathcal{N}(\bm{\mu}_l, \bm{\sigma}_l)$ over $\bm{l}$ using Bayes rule. As shown in \cite{volpp2021bayesian,shaj2022hidden}, the beliefs over $\bm{l}$ can be computed in closed form, given a Gaussian prior \( p_0(\mathbf{l}) \). The update rules and their properties are detailed in Appendix \ref{appen:implementation}.
\begin{figure}[t]
    \centering
    \includegraphics[width=0.9\linewidth]{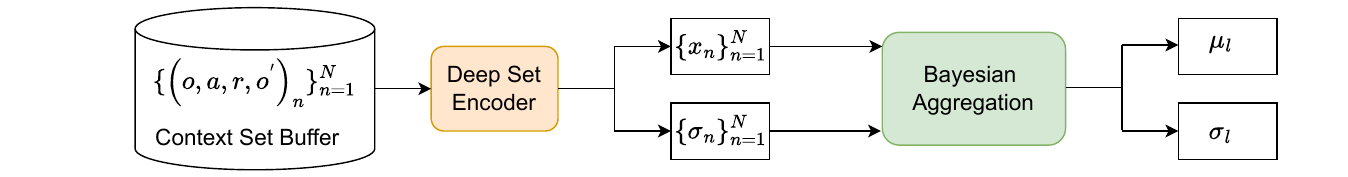}
    \caption{Given a set of N transitions, the deep set encoder emits a latent representation for each of the observations and their corresponding uncertainty. The set of latent representations is then aggregated via Bayesian aggregation to infer $p \left( \bm{l} \mid \bm{C_l} \right)$.}
    \label{fig:infer_latent_representation_pipeline}
\end{figure}

\subsection{Learning Adaptive Representations}
\label{subsec:learning_adaptive_representations}
\begin{wrapfigure}[21]{r}{0.42\textwidth}
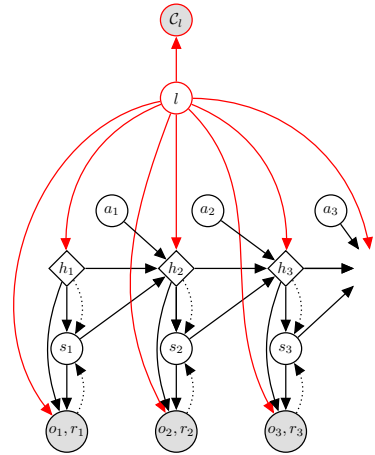

    \centering
    \scalebox{.6}{\tikzHiPDreamer}
    \caption{Hidden Parameter RSSM: The latent task variable is inferred from context $\bm{C}_{l}$ via Bayesian aggregation. Solid lines indicate the generative process and dashed lines the inference model. Modifications from \cite{hafner2019dream} are shown in red.}
    \label{fig:hip_rssm}
\end{wrapfigure}

In this stage, we learn representations of generative world models that can make counterfactual predictions of the world states based on imagined actions. We make these learned representations adaptive to the task at hand based on the generative model shown in Figure \ref{fig:hip_rssm}. We achieve this by maximizing the conditional data log-likelihood and subsequently deriving an evidence lower bound, as in Equation \ref{eq:context_elbo}. A detailed derivation can be found in Appendix \ref{subsec:hip_rssm_full_objective_derivation}.

\vspace{-0.7cm}
\begin{align}
    & \ln p\left(\bm{o}_{1: T}, r_{1: T} \mid \bm{a}_{1: T}, \bm{C}_l \right) \nonumber 
    \geq 
    \sum_{t=1}^{T} 
    \underbrace{
        \mathbb{E}_{p\left(\bm{l} \mid \bm{C}_l\right)q\left(\bm{s}_t \mid \bm{o}_{\leq t}, \bm{a}_{<t}, \bm{l}\right)}
        \left[ 
            \ln p\left(\bm{o}_t, r_t \mid \bm{s}_t, \bm{l}\right)
        \right]
    }_{\textbf{Reconstruction Term}} \nonumber \\
    & \quad +
    \underbrace{
        \mathbb{E}_{p\left(\bm{l} \mid \bm{C}_l\right)q\left(\bm{s}_{t-1} \mid \bm{o}_{\leq t-1}, \bm{a}_{<t-1}, \bm{l}\right)}
        \left[ 
            \mathrm{D}_{\mathrm{KL}}\left(
                q\left(\bm{s}_t \mid  \bm{o}_{\leq t},  \bm{a}_{<t}, 
                \bm{l}\right)
                \,\|\, 
                p\left(\bm{s}_t \mid \bm{s}_{t-1}, \bm{a}_{t-1}, \bm{l}\right)
            \right)
        \right]
    }_{\textbf{Regularization Term}} \label{eq:context_elbo}
\end{align}

The outer expectation can be estimated using a reparameterized sample from the latent task posterior $p\left( \bm{l} \mid \bm{C}_l \right)$. The practical implementation of this builds upon the RSSMs used in the popular Dreamer series of models~\cite{hafner2019dream,hafner2019learning,hafner2021mastering, hafner2024masteringdiversedomainsworld}, where an additional deterministic path (using a GRU) is used in addition to the stochastic SSM for long-term predictions. The subsequent Hidden Parameter-RSSM generative model is shown in Figure \ref{fig:hip_rssm}.

\paragraph{Discussion:}Though we use the generative model from \cite{hafner2019dream,hafner2019learning}, the HiP-POMDP formalism can be used in conjunction with any model-based RL framework in latent spaces~\cite{becker2022uncertainty,hafner2021mastering,hafner2024masteringdiversedomainsworld,hansen2022temporal,hansen2023td,samsami2024mastering} for multitask learning. 

\subsection{Learning Adaptive Behaviors}
The agent optimizes long-term rewards using a context-sensitive actor-critic approach~\cite{sutton2018reinforcement,hafner2019dream}, conditioning both actor and critic on the latent task representation \( \bm{l} \). The actor \( \pi_{\bm{\phi}}(\bm{a}_t \mid \bm{s}_t, \bm{l}) \) selects actions to maximize expected values along imagined trajectories, while the critic \( v_{\bm{\psi}}(\bm{s}_\tau, \bm{l}) \) regresses those estimates:
\[
\begin{aligned}
\max_{\bm{\phi}} \ \mathbb{E}_{q_{\bm{\theta}}, \pi_{\bm{\phi}}}\left( \sum_{\tau=t}^{t+H} \mathrm{V}_\lambda(\bm{s}_\tau, \bm{l}) \right), \quad
\min_{\bm{\psi}} \ \mathbb{E}_{q_{\bm{\theta}}, \pi_{\bm{\phi}}}\left( \sum_{\tau=t}^{t+H} \frac{1}{2} \left\| v_{\bm{\psi}}(\bm{s}_\tau, \bm{l}) - \mathrm{V}_\lambda(\bm{s}_\tau, \bm{l}) \right\|^2 \right).
\end{aligned}
\]
Here, \( \mathrm{V}_\lambda(\bm{s}_\tau, \bm{l}) \) is the \( \lambda \)-return~\cite{sutton2018reinforcement,hafner2019dream}, a smoothed estimate of the cumulative reward that balances short- and long-term returns using the discount factor \( \lambda \). We compute analytic gradients through the learned dynamics to optimize the actor via stochastic backpropagation through time.

\vspace{-0.2cm}
\paragraph{Discussion} As shown in Figure 2, during a short imagination rollout the latent task $\bm{l}$ remains fixed. This is a reasonable assumption for such short horizons. However, during environment interaction, the context buffer is updated continuously. This enables the agent to re-infer the task and adapt to both inter and intra-episodic task changes.

\section{Evaluation}
In this section, we evaluate the performance of two competing formalisms—POMDP and HiP-POMDP—in handling non-stationarity within an episodic evaluation setting. We focus on three broad categories of non-stationarity: (1) changing transition functions, (2) changing rewards, and (3) a combination of both. For each category, we further consider two scenarios:

\begin{itemize}
    \item \textbf{Inter-Episodic Non-Stationarity:} Changes remain fixed within an episode but vary between episodes.
    \item \textbf{Intra-Episodic Non-Stationarity:} Non-stationary changes can occur within a single episode.
\end{itemize}

A more detailed description of these scenarios is provided in Appendix C. In all experiments, proprioceptive sensors are used as the source of observations.

\paragraph{Algorithms Compared:} We use the Dreamer \cite{hafner2019dream} as our baseline for the POMDP formalism. For the HiP-POMDP, we modify Dreamer by incorporating latent task abstractions, ensuring a fair comparison between the two approaches. Additionally, we include an "Oracle" baseline where the task is assumed to be directly observed. In this setup, the known task replaces the inferred latent task variable $\bm{l}$, serving as an upper bound on performance. This helps illustrate the potential gains if perfect task information were available.

We evaluate the agents in all experiments by calculating the mean return from 10 trajectories every 25 epochs, each with randomly sampled environmental changes. The performance curves are computed by averaging the results over 10 different random seeds. Our evaluation answers the following questions: 

\paragraph{Can HiP-POMDP agents handle changing dynamics?}
To evaluate the effectiveness of HiP-POMDP formalism under changing dynamics, here we introduce two tasks: 1) We modify the standard HalfCheetah agent by adding joint perturbations of varying magnitudes randomly, and 2) the Hopper agent by randomly changing the body mass and inertia for random number of body parts. Additional evaluation is introduced in Appendix \ref{appen:evaluation_dynamical_changes}.

\begin{figure}
    \centering
    \includegraphics[width=0.95\linewidth]{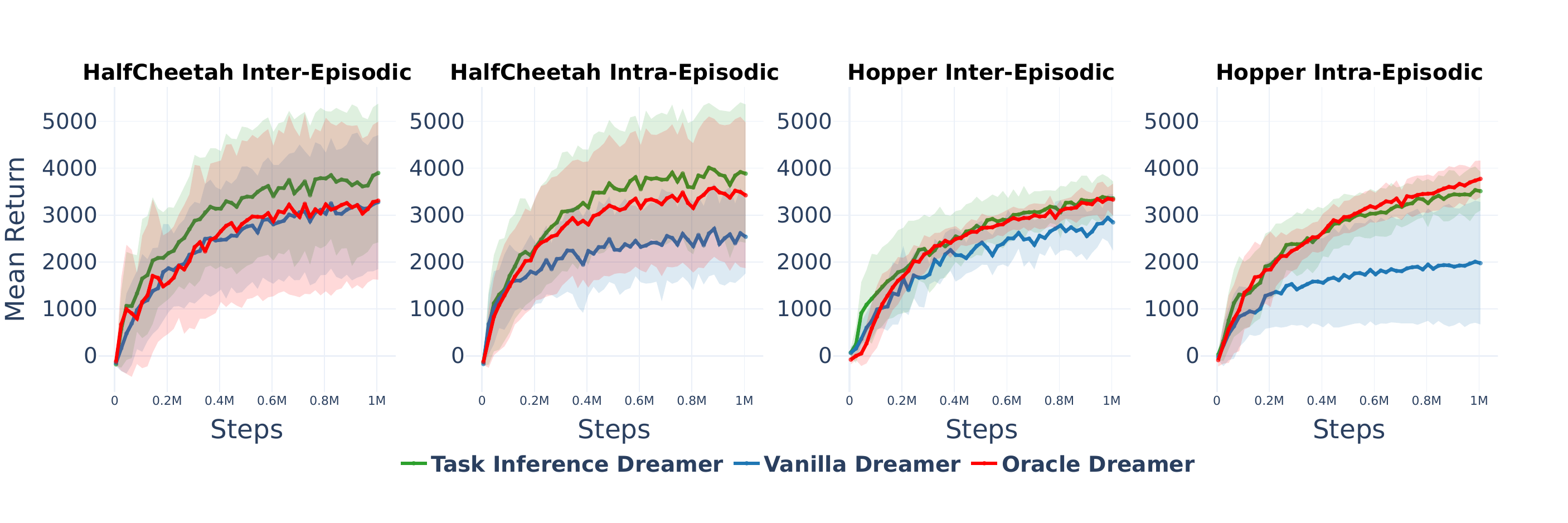}
    \vspace{-15pt}
    \caption{Performance of HalfCheetah and Hopper agents under changing dynamics caused by joint perturbations and body mass inertia variations, respectively.}
    \label{fig:inter_intra_halfcheetah_hopper_dynamical_changes}
\end{figure}

As seen in Figure \ref{fig:inter_intra_halfcheetah_hopper_dynamical_changes} HiP-POMDP agent results in robust performance gains, especially under challenging intra-episodic changes and even competing with the Oracle. 

\paragraph{Can HiP-POMDP agents handle changing objectives?}
\label{para:evaluation_objective_changes_main}To create non-stationarity with changing reward functions/objectives, we modify the standard HalfCheetah such that a target velocity needs to be reached which changes randomly.
Additionally, we evaluate the agents on custom-designed multi-task benchmarks using pre-defined tasks from \cite{hansen2024tdmpc}, where each task requires the agent to perform different skills (e.g., standing, running, flip) in various directions. As such multi-skill objective changes are more challenging, the experiments are run over 5M steps.

\begin{figure}
    \centering
    \includegraphics[width=0.95\linewidth]{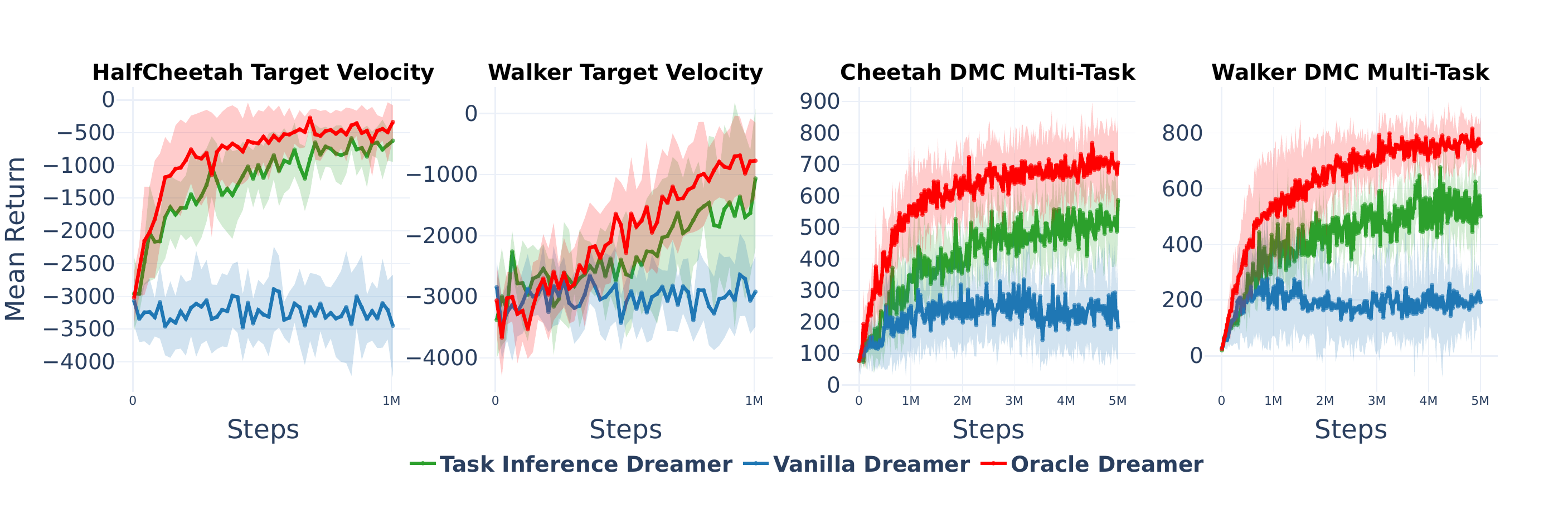}
    \vspace{-15pt}
    \caption{Performance comparison of Half Cheetah and Walker agents under different changing reward scenarios (changing target velocities and skills). }
\label{fig:inter_halfcheetah_walker_objective_changes}
\end{figure}

As seen in Figure \ref{fig:inter_halfcheetah_walker_objective_changes} the vanilla POMDP agent fails to deal with objective changes in all cases. On the other hand side, the HiP-POMDP agent with inferred task abstractions resolves the issue to a large extent. 
Further investigation as well as evaluation under combined changes can be found in Appendix \ref{appen:objective_changes} and \ref{appen:combined_changes} respectively.

\paragraph{Does the HiP-POMDP agents learn meaningful latent representations?}
    \begin{figure}[ht]
        \centering
        \begin{minipage}[b]{0.27\textwidth}
            \centering
            \scalebox{0.6}{\parbox{1.5\linewidth}{\centering \textbf{Task Inference Dreamer \\ Latent Task Space}}}
            \vspace{-0.5em}  
            \includegraphics[width=\textwidth]{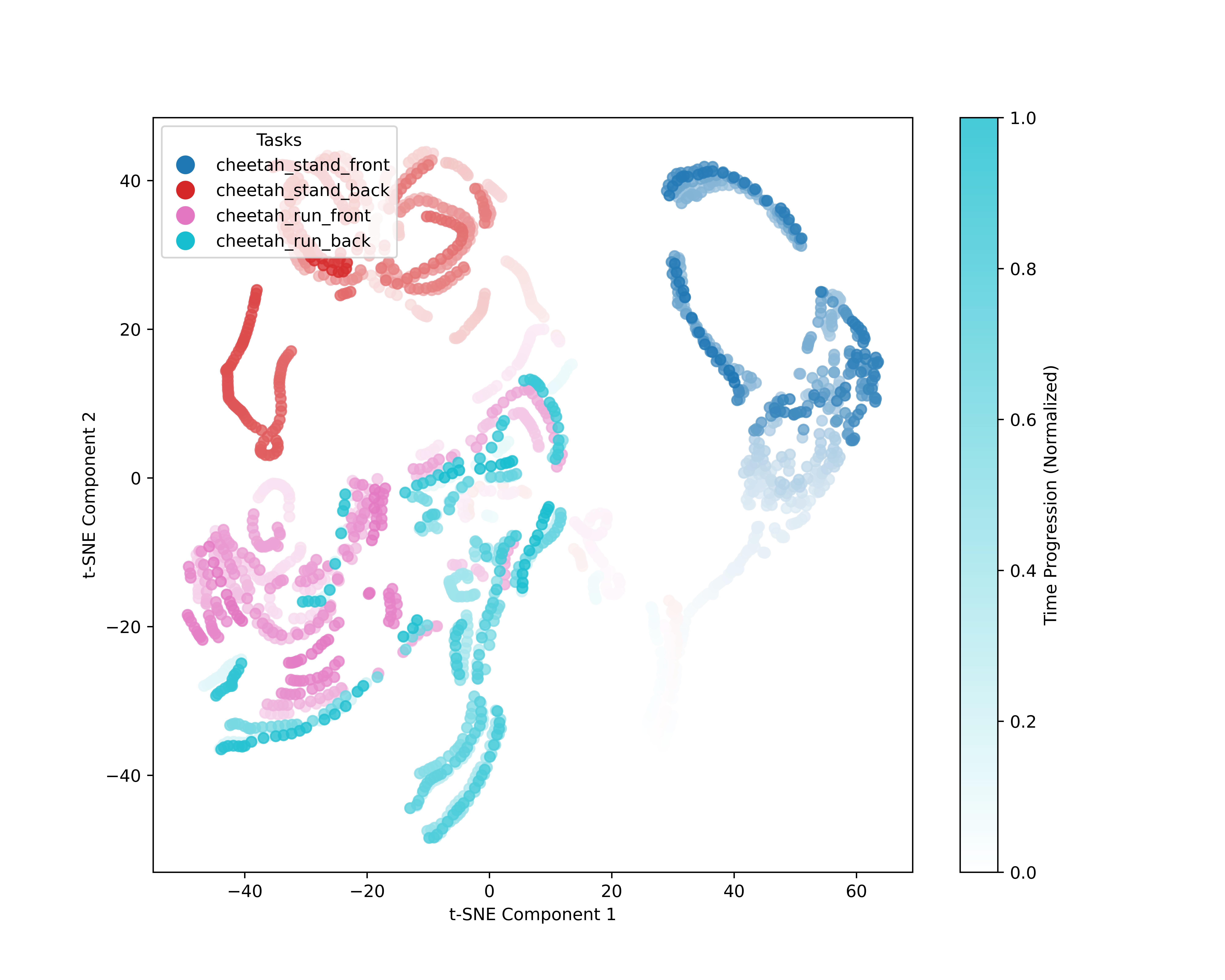}
        \end{minipage}
        \hspace{-18pt}
        \begin{minipage}[b]{0.27\textwidth}
            \centering
            \scalebox{0.6}{\parbox{1.5\linewidth}{\centering \textbf{Task Inference Dreamer \\ Latent State Space}}}
            \vspace{-0.5em}  
            \includegraphics[width=\textwidth]{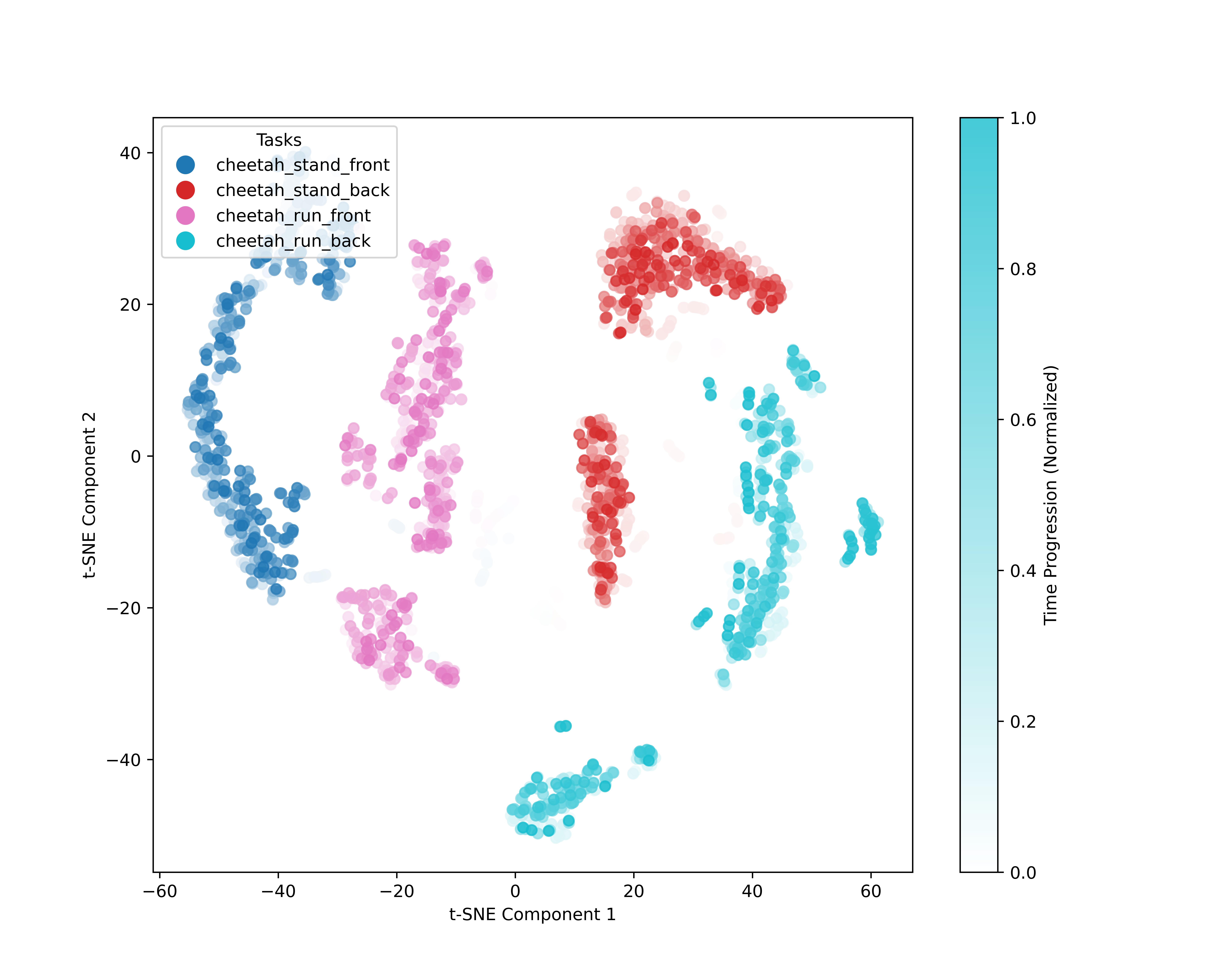}
        \end{minipage}
        \hspace{-18pt}
        \begin{minipage}[b]{0.27\textwidth}
            \centering
            \scalebox{0.6}{\parbox{1.5\linewidth}{\centering \textbf{Vanilla Dreamer \\ Latent State Space}}}
            \vspace{-0.5em}  
            \includegraphics[width=\textwidth]{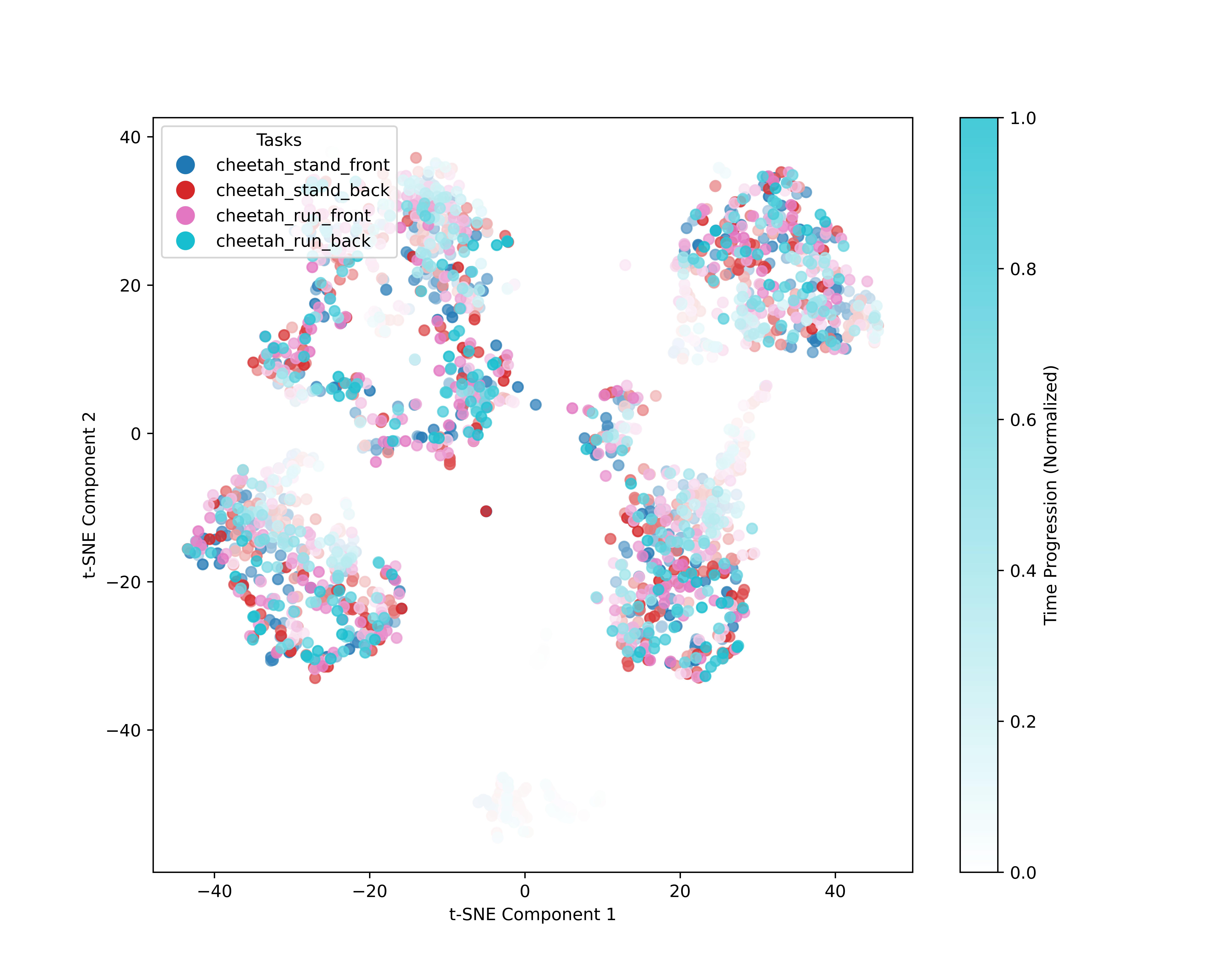}
        \end{minipage}
        \hspace{-18pt}
        \begin{minipage}[b]{0.27\textwidth}
            \centering
            \scalebox{0.6}{\parbox{1.5\linewidth}{\centering \textbf{Oracle Dreamer \\ Latent State Space}}}
            \vspace{-0.5em}  
            \includegraphics[width=\textwidth]{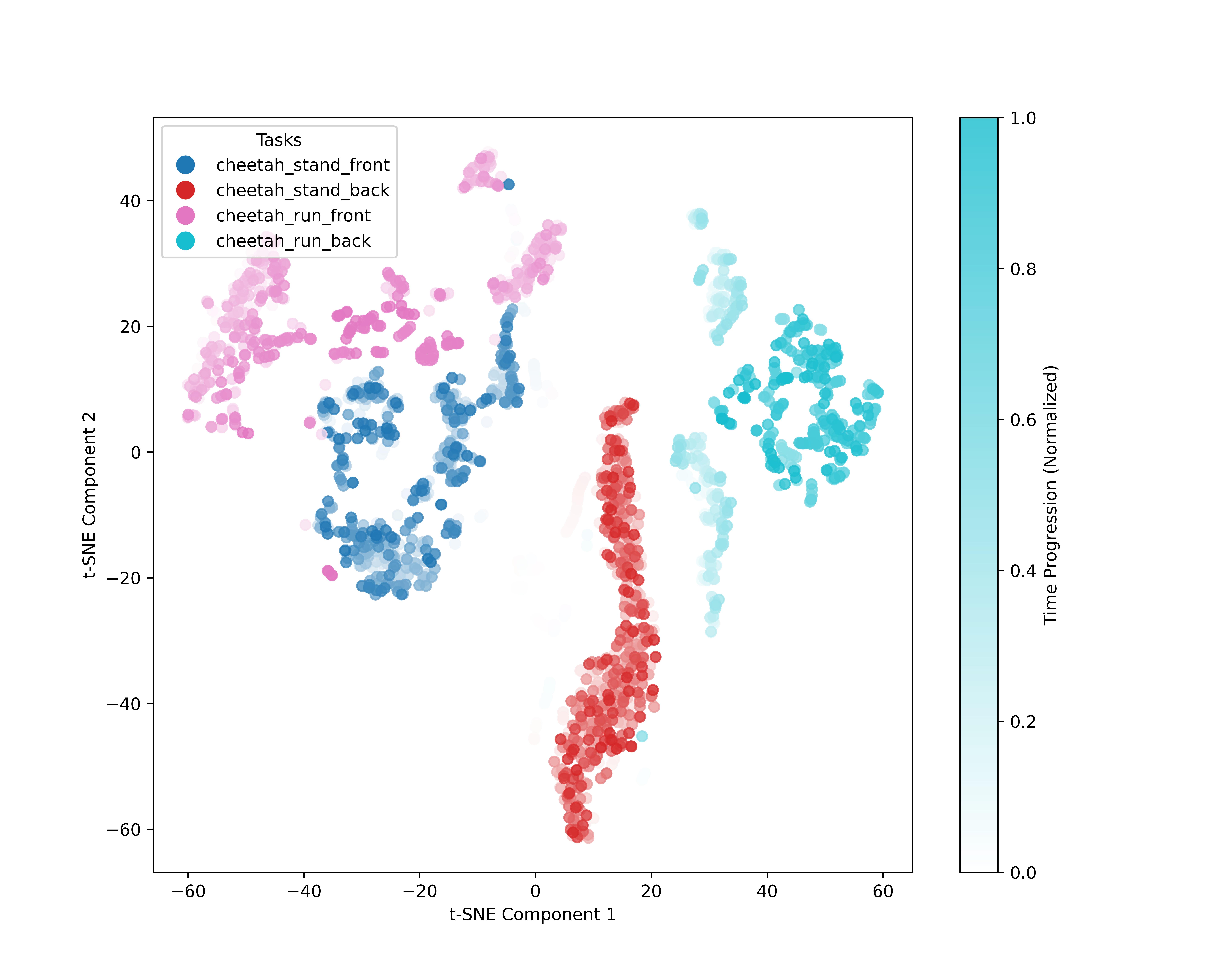}
        \end{minipage}
        \caption{2d projections of learned latent state spaces on DMC Cheetah learning 4 skills, Table \ref{table:dmc_multi_task_benchmarks}.}
        \label{fig:cheetah_easy_rssm_latent_state_space}
    \end{figure}
    
Finally, we compared the learned state-space representations (the concatenation of $\bm{s_t}$ and $\bm{h_t}$) of the world model (RSSM) in both POMDP and HiP-POMDP settings. As shown in Figure \ref{fig:cheetah_easy_rssm_latent_state_space}, the task abstractions in HiP-POMDP shape a more structured and disentangled latent space that aligns with the inferred tasks, unlike the POMDP setting. This disentanglement was also observed in the latent task space representation ($\bm{l}$) within the HiP-POMDP setup.

\section{Conclusion}
In this work, we introduced the HiP-POMDP formalism to learn adaptive world models and behavior policies in latent state spaces. The formalism resulted in algorithms that learn meaningful task abstractions and improved performance on a variety of non-stationary benchmarks. Future work would extend these models to more high-dimensional sensory inputs like images and point clouds. 

\section*{Acknowledgments}
The authors acknowledge support by the state of Baden-Württemberg through bwHPC, as well as the HoreKa supercomputer funded by the Ministry of Science, Research and the Arts Baden-Württemberg and by the German Federal Ministry of Education and Research.

\iflanguage{english}{%
  \bibliographystyle{abbrvnat}%
}{%
  \bibliographystyle{dinat}%
}
\bibliography{bibliography}

\appendix
\section{Objective Derivation}
\label{subsec:hip_rssm_full_objective_derivation}
    \begin{align*}\label{eq:hip_rssm_full_objective_derivation}
        & \ln p\left(\bm{o}_{1: T}, r_{1: T} \mid \bm{a}_{1: T}, \bm{C_{l}} \right) \\
        &= \ln
            \mathop{\mathbb{E}}_{p\left(\bm{l} \mid \bm{C_{l}}\right)}
                \left[
                    p\left(\bm{o}_{1: T}, r_{1: T} \mid \bm{a}_{1: T}, \bm{l} \right)
                \right] 
        \quad\text{(Data aggregation for latent task inference given the context $\boldsymbol{C_{l}}$)} \\
        &= \ln 
            \mathop{\mathbb{E}}_{p\left(\bm{l} \mid \bm{C_{l}}\right)}
                \left[
                    \mathop{\mathbb{E}}_{p\left(\bm{s}_{1: T} \mid \bm{a}_{1: T}, \bm{l}\right)}
                        \left[p\left(\bm{o}_{1: T}, r_{1: T} \mid \bm{s}_{1: T}, \bm{l}\right)\right]
                \right]
        \quad\parbox{15em}{(Posterior predictive log-likelihood \\ using Latent Variable Model)} \\
        &= \ln 
            \mathop{\mathbb{E}}_{p\left(\bm{l} \mid \bm{C_{l}}\right)}
                \left[
                    \int 
                        p\left(\bm{s}_{1: T} \mid \bm{a}_{1: T}, \bm{l}\right) 
                        p\left(\bm{o}_{1: T}, r_{1: T} \mid \bm{s}_{1: T}, \bm{l}\right)
                    \,d\bm{s}_{1: T} 
                \right]
        \quad\text{(Expectation definition)} \\
        &= \ln 
            \mathop{\mathbb{E}}_{p\left(\bm{l} \mid \bm{C_{l}}\right)}
                \left[                        
                    \int 
                        \prod_{t=1}^{T} 
                            p\left(\bm{s}_{t} \mid \bm{s}_{t-1}, \boldsymbol{a}_{t-1}, \bm{l}\right) 
                            p\left(\bm{o}_{t}, r_t \mid \bm{s}_{t}, \bm{l}\right) 
                    \,d\bm{s}_{1: T}
                \right]
        \quad\text{(Factorization due to Markov property)} \\
        &= \ln 
            \mathop{\mathbb{E}}_{p\left(\bm{l} \mid \bm{C_{l}}\right)}
                \left[   
                    \int 
                        \prod_{t=1}^{T}
                            \frac
                                {q\left(\bm{s}_{t} \mid \bm{o}_{\leq t}, \bm{a}_{<t}, \bm{l}\right)}{q\left(\bm{s}_{t} \mid \bm{o}_{\leq t}, \bm{a}_{<t}, \bm{l}\right)}
                            p\left(\bm{s}_{t} \mid \bm{s}_{t-1}, \bm{a}_{t-1}, \bm{l}\right) 
                            p\left(\bm{o}_{t}, r_t \mid \bm{s}_{t}, \bm{l}\right) 
                    \,d\bm{s}_{1: T}
                \right]
        \quad\parbox{15em}{(Variational approximate \\ posterior)} \\      
        &= \ln 
             \mathop{\mathbb{E}}_{p\left(\bm{l} \mid \bm{C_{l}}\right)}
                \left[              
                    \mathop{\mathbb{E}}_{q\left(\bm{s}_{1: T} \mid \bm{o}_{1: T}, \bm{a}_{1: T}, \bm{l}\right)}
                        \left[
                            \prod_{t=1}^{T} 
                                \frac
                                    {p\left(\bm{o}_{t}, r_t \mid \bm{s}_{t}, \bm{l}\right) p\left(\bm{s}_{t} \mid \bm{s}_{t-1}, \bm{a}_{t-1}, \bm{l}\right)}
                                    {q\left(\bm{s}_{t} \mid \bm{o}_{\leq t}, \bm{a}_{<t}, \bm{l}\right)}
                        \right] 
                \right] 
        \quad\parbox{15em}{(Expectation w.r.t approximate posterior + exchange order \\ of nominator factors)} \\
        &\ge 
             \mathop{\mathbb{E}}_{p\left(\bm{l} \mid \bm{C_{l}}\right)}
                \left[ 
                    \ln              
                        \mathop{\mathbb{E}}_{q\left(\bm{s}_{1: T} \mid \bm{o}_{1: T}, \bm{a}_{1: T}. \bm{l}\right)}
                            \left[
                                \prod_{t=1}^{T} 
                                    \frac
                                        {p\left(\bm{o}_{t}, r_t \mid \bm{s}_{t}, \bm{l}\right) p\left(\bm{s}_{t} \mid \bm{s}_{t-1}, \bm{a}_{t-1}, \bm{l}\right)}
                                        {q\left(\bm{s}_{t} \mid \bm{o}_{\leq t}, \bm{a}_{<t}, \bm{l}\right)}
                            \right] 
                \right]
        \quad\parbox{15em}{(Jensen's inequality \\ w.r.t. outer expectation)} \\
        &\ge 
             \mathop{\mathbb{E}}_{p\left(\bm{l} \mid \bm{C_{l}}\right)}
                \left[ 
                    \mathop{\mathbb{E}}_{q\left(\bm{s}_{1: T} \mid \bm{o}_{1: T}, \bm{a}_{1: T}, \bm{l}\right)}
                            \left[
                                \ln              
                                    \prod_{t=1}^{T} 
                                        \frac
                                            {p\left(\bm{o}_{t}, r_t \mid \bm{s}_{t}, \bm{l}\right) p\left(\bm{s}_{t} \mid \bm{s}_{t-1}, \bm{a}_{t-1}, \bm{l}\right)}
                                            {q\left(\bm{s}_{t} \mid \bm{o}_{\leq t}, \bm{a}_{<t}, \bm{l}\right)}
                            \right] 
                \right]
        \quad\parbox{15em}{(Jensen's inequality \\ w.r.t. inner expectation)} \\
        &= 
             \mathop{\mathbb{E}}_{p\left(\bm{l} \mid \bm{C_{l}}\right)}
                \left[ 
                    \mathop{\mathbb{E}}_{q\left(\bm{s}_{1: T} \mid \bm{o}_{1: T}, \bm{a}_{1: T}, \bm{l}\right)}
                            \left[
                                \sum_{t=1}^{T} 
                                    \ln
                                        \frac
                                            {p\left(\bm{o}_{t}, r_t \mid \bm{s}_{t}, \bm{l}\right) p\left(\bm{s}_{t} \mid \bm{s}_{t-1}, \bm{a}_{t-1}, \bm{l}\right)}
                                            {q\left(\bm{s}_{t} \mid \bm{o}_{\leq t}, \bm{a}_{<t}, \bm{l}\right)}
                            \right] 
                \right]
        \quad\text{(Product rule of logarithms)} \\
        &= 
             \mathop{\mathbb{E}}_{p\left(\bm{l} \mid \bm{C_{l}}\right)}
                \left[ 
                    \mathop{\mathbb{E}}_{q\left(\bm{s}_{1: T} \mid \bm{o}_{1: T}, \bm{a}_{1: T}, \bm{l}\right)}
                            \left[
                                \sum_{t=1}^{T} 
                                    \ln 
                                        p\left(\bm{o}_{t}, r_t \mid \bm{s}_{t}, \bm{l}\right) 
                                        p\left(\bm{s}_{t} \mid \bm{s}_{t-1}, \bm{a}_{t-1}, \bm{l}\right) - 
                                        \ln {q\left(\bm{s}_{t} \mid \bm{o}_{\leq t}, \bm{a}_{<t}, \bm{l}\right)}
                            \right] 
                \right]
        \quad\parbox{15em}{(Quotient \\ logarithm \\ rule)} \\
        &= 
             \mathop{\mathbb{E}}_{p\left(\bm{l} \mid \bm{C_{l}}\right)}
                \left[ 
                    \mathop{\mathbb{E}}_{q\left(\bm{s}_{1: T} \mid \bm{o}_{1: T}, \bm{a}_{1: T}, \bm{l}\right)}
                            \left[
                                \sum_{t=1}^{T} 
                                    \ln p\left(\bm{o}_{t}, r_t \mid \bm{s}_{t}, \bm{l}\right) + 
                                    \ln p\left(\bm{s}_{t} \mid \bm{s}_{t-1}, \bm{a}_{t-1}, \bm{l}\right) - 
                                    \ln {q\left(\bm{s}_{t} \mid \bm{o}_{\leq t}, \bm{a}_{<t}, \bm{l}\right)}
                            \right] 
                \right]
        \quad\parbox{15em}{(Product \\ logarithm \\ rule)} \\
        &= 
            \sum_{t=1}^{T} 
                 \mathop{\mathbb{E}}_{p\left(\bm{l} \mid \bm{C_{l}}\right)}
                    \left[ 
                        \mathop{\mathbb{E}}_{q\left(\bm{s}_t \mid \bm{o}_{\leq t}, \bm{a}_{<t}, \bm{l}\right)}
                                \left[
                                        \ln p\left(\bm{o}_{t}, r_t \mid \bm{s}_{t}, \bm{l}\right) + 
                                        \ln p\left(\bm{s}_{t} \mid \bm{s}_{t-1}, \bm{a}_{t-1}, \bm{l}\right) - 
                                        \ln {q\left(\bm{s}_{t} \mid \bm{o}_{\leq t}, \bm{a}_{<t}, \bm{l}\right)}
                                \right] 
                    \right]
        \quad\parbox{15em}{(Linearity \\ of expectation)} \\
        &= 
            \sum_{t=1}^{T} 
                \mathop{\mathbb{E}}_{p\left(\bm{l} \mid \bm{C_{l}}\right)}
                    \left[ 
                        \mathop{\mathbb{E}}_{q\left(\bm{s}_t \mid \bm{o}_{\leq t}, \bm{a}_{<t}, \bm{l}\right)}
                                \left[
                                        \ln p\left(\bm{o}_{t}, r_t \mid \bm{s}_{t}, \bm{l}\right)
                                \right] 
                    \right] + \\
                 &\phantom{= \sum_{t=1}^{T}\ }
                 \mathop{\mathbb{E}}_{p\left(\bm{l} \mid \bm{C_{l}}\right)}
                    \left[ 
                        \mathop{\mathbb{E}}_{q\left(\bm{s}_t \mid \bm{o}_{\leq t}, \bm{a}_{<t}, \bm{l}\right)}
                                \left[
                                        \ln p\left(\bm{s}_{t} \mid \bm{s}_{t-1}, \bm{a}_{t-1}, \bm{l}\right) - 
                                        \ln {q\left(\bm{s}_{t} \mid \bm{o}_{\leq t}, \bm{a}_{<t}, \bm{l}\right)}
                                \right] 
                    \right]
        \quad\text{(Linearity of expectation)} \\
        &= 
            \sum_{t=1}^{T} 
                \mathop{\mathbb{E}}_{p\left(\bm{l} \mid \bm{C_{l}}\right)}
                    \left[ 
                        \mathop{\mathbb{E}}_{q\left(\bm{s}_{t} \mid \bm{o}_{\leq t}, \bm{a}_{<t}, \bm{l}\right)}
                                \left[
                                        \ln p\left(\bm{o}_{t}, r_t \mid \bm{s}_{t}, \bm{l}\right)
                                \right] 
                    \right] + \\
                 &\phantom{= \sum_{t=1}^{T}\ }
                 \mathop{\mathbb{E}}_{p\left(\bm{l} \mid \bm{C_{l}}\right)}
                    \left[ 
                        \mathop{\mathbb{E}}_{\substack{ q\left(\bm{s}_{t-1} \mid \bm{o}_{\leq t-1}, \bm{a}_{<t-1}, \bm{l}\right) \\ q\left(\bm{s}_{t} \mid \bm{o}_{\leq t}, \bm{a}_{<t}, \bm{l}\right)}}
                                \left[
                                        \ln p\left(\bm{s}_{t} \mid \bm{s}_{t-1}, \bm{a}_{t-1}, \bm{l}\right) - 
                                        \ln {q\left(\bm{s}_{t} \mid \bm{o}_{\leq t}, \bm{a}_{<t}, \bm{l}\right)}
                                \right] 
                    \right]
        \quad\parbox{15em}{(Expectation \\ is constant \\ for $t\not\in\{t, t-1\}$)} \\
        &= 
            \sum_{t=1}^{T} 
                \mathop{\mathbb{E}}_{p\left(\bm{l} \mid \bm{C_{l}}\right)}
                    \left[ 
                        \mathop{\mathbb{E}}_{q\left(\bm{s}_{t} \mid \bm{o}_{\leq t}, \bm{a}_{<t}, \bm{l}\right)}
                                \left[
                                        \ln p\left(\bm{o}_{t}, r_t \mid \bm{s}_{t}, \bm{l}\right)
                                \right] 
                    \right] + \\ 
                 &\phantom{= \sum_{t=1}^{T}\ }
                 \mathop{\mathbb{E}}_{p\left(\bm{l} \mid \bm{C_{l}}\right)}
                    \left[ 
                        \mathop{\mathbb{E}}_{q\left(\bm{s}_{t-1} \mid \bm{o}_{\leq t-1}, \bm{a}_{<t-1}, \bm{l}\right)}
                                \left[
                                    \mathop{\mathbb{E}}_{q\left(\bm{s}_{t} \mid \bm{o}_{\leq t}, \bm{a}_{<t}, \bm{l}\right)}
                                        \left[
                                            \ln p\left(\bm{s}_{t} \mid \bm{s}_{t-1}, \bm{a}_{t-1}, \bm{l}\right) - 
                                            \ln {q\left(\bm{s}_{t} \mid \bm{o}_{\leq t}, \bm{a}_{<t}, \bm{l}\right)}
                                        \right]
                                \right] 
                    \right] 
        \quad\parbox{15em}{(Expectation \\ split)} \\
        &= 
            \sum_{t=1}^{T} 
                \mathop{\mathbb{E}}_{p\left(\bm{l} \mid \bm{C_{l}}\right)}
                    \left[ 
                        \mathop{\mathbb{E}}_{q\left(\bm{s}_{t} \mid \bm{o}_{\leq t}, \bm{a}_{<t}, \bm{l}\right)}
                                \left[
                                        \ln p\left(\bm{o}_{t}, r_t \mid \bm{s}_{t}, \bm{l}\right)
                                \right] 
                    \right] + \\ 
                 &\phantom{= \sum_{t=1}^{T}\ }
                 \mathop{\mathbb{E}}_{p\left(\bm{l} \mid \bm{C_{l}}\right)}
                    \left[ 
                        \mathop{\mathbb{E}}_{q\left(\bm{s}_{t-1} \mid \bm{o}_{\leq t-1}, \bm{a}_{<t-1}, \bm{l}\right)}
                                \left[
                                    \mathop{\mathbb{E}}_{q\left(\bm{s}_{t} \mid \bm{o}_{\leq t}, \bm{a}_{<t}, \bm{l}\right)}
                                        \left[ 
                                            \ln
                                                \frac
                                                    {p\left(\bm{s}_{t} \mid \bm{s}_{t-1}, \bm{a}_{t-1}, \bm{l}\right)}
                                                    {q\left(\bm{s}_{t} \mid \bm{o}_{\leq t}, \bm{a}_{<t}, \bm{l}\right)}
                                        \right]
                                \right] 
                    \right]
        \quad\parbox{15em}{(Quotient rule \\ for logarithms)} \\
        &= 
            \sum_{t=1}^{T} 
                \mathop{\mathbb{E}}_{p\left(\bm{l} \mid \bm{C_{l}}\right)}
                    \left[ 
                        \mathop{\mathbb{E}}_{q\left(\bm{s}_{t} \mid \bm{o}_{\leq t}, \bm{a}_{<t}, \bm{l}\right)}
                                \left[
                                        \ln p\left(\bm{o}_{t}, r_t \mid \bm{s}_{t}, \bm{l}\right)
                                \right] 
                    \right] + \\ 
                 &\phantom{= \sum_{t=1}^{T}\ }
                 \mathop{\mathbb{E}}_{p\left(\bm{l} \mid \bm{C_{l}}\right)}
                    \left[ 
                        \mathop{\mathbb{E}}_{q\left(\bm{s}_{t-1} \mid \bm{o}_{\leq t-1}, \bm{a}_{<t-1}, \bm{l}\right)}
                                \left[
                                    \infdiv{q\left(\bm{s}_{t} \mid \bm{o}_{\leq t}, \bm{a}_{<t}, \bm{l}\right)}{p\left(\bm{s}_{t} \mid \bm{s}_{t-1}, \bm{a}_{t-1}, \bm{l}\right)}
                                \right] 
                    \right]
        \quad\parbox{15em}{(KL-Divergence \\ definition)} \\
        &= 
            \sum_{t=1}^{T} 
                \underbrace{
                    \underbrace{
                        \mathop{\mathbb{E}}_{\substack{p\left(\bm{l} \mid \bm{C_{l}}\right), \\ q\left(\bm{s}_{t} \mid \bm{o}_{\leq t}, \bm{a}_{<t}, \bm{l}\right)}}
                            \left[ 
                                \ln p\left(\bm{o}_{t}, r_t \mid \bm{s}_{t}, \bm{l}\right)
                            \right]
                    }_{\textbf{Reconstruction}} 
                    + 
                    \underbrace{
                        \mathop{\mathbb{E}}_{\substack{p\left(\bm{l} \mid \bm{C_{l}}\right), \\ q\left(\bm{s}_{t-1} \mid \bm{o}_{\leq t-1}, \bm{a}_{<t-1}, \bm{l}\right)}}                                \left[ 
                                \infdiv{q\left(\bm{s}_{t} \mid \bm{o}_{\leq t}, \bm{a}_{<t}, \bm{l}\right)}{p\left(\bm{s}_{t} \mid \bm{s}_{t-1}, \bm{a}_{t-1}, \bm{l}\right)}
                            \right]                    
                    }_{\textbf{Regularization}}
                }_{\textbf{Context-dependent Evidence Lower Bound}}
    \end{align*}

\section{Algorithm and Implementation Details}
\label{appen:implementation}

\subsection{Task Abstractions Via Bayesian Aggregation}
\begin{wrapfigure}[12]{r}{0.35\textwidth}
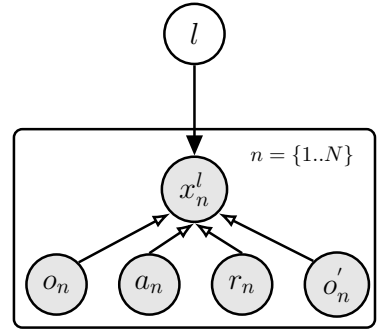

\centering
    \scalebox{.8}{\tikzLatentTaskInference}
    \caption{Generative model for the abstract latent task $\bm{l}$. The hollow arrows are deterministic transformations leading to implicit distribution $\bm{x}_{n}^l$ using a set encoder.}
\end{wrapfigure}
Given a set of encoded interaction tuples and their corresponding variances $\{\bm{x}_n^l, \bm{\sigma}_n^l \}_{n=1}^N $, using the prior and observation model assumptions in Section \ref{subsec:task-infer} of the main paper, we infer the latent task abstraction $p\left( \bm{l} \mid \bm{C}_l \right) = \mathcal{N}(\bm{\mu}_l, \bm{\Sigma}_{l})  = \mathcal{N}(\bm{\mu}_l, \textrm{diag}(\bm{\sigma}_{l}))$ as a Bayesian aggregation~\cite{volpp2021bayesian} of these using the following closed-form equations:
\begin{align*}
        \bm{\sigma_{l}}^2 & = \left( \left(\bm{\sigma}_{0}^2\right)^\ominus+ \sum_{n=1}^N \left(\left(\bm{\sigma}_n^{\bm{l}}\right)^2\right)^\ominus\right)^\ominus, \\ \bm{\mu}_{\bm{l}} & = \bm{\mu}_{0} + \bm{\sigma}_{\bm{l}}^2 \odot \sum_{n=1}^N \left(\bm{x}^{\bm{l}}_n - \bm{\mu}_{0}\right) \oslash \left(\bm{\sigma}_n^{\bm{l}}\right)^2
\end{align*}
where $\ominus$, $\odot$, and $\oslash$ denote element-wise inversion, product, and division, respectively. 

\paragraph{Discussion:} This closed-form solution represents Bayesian aggregation, which can be interpreted as a form of \textbf{probabilistic attention}. The uncertainty \( \bm{\sigma}^{\bm{l}}_{n} \) about each transition in the set encoder functions as attention weights, assigning higher weights to the most informative transitions. The derived update equations have only a linear computational complexity of $O(N)$, while similar deep set operations (self-attention) in transformers~\cite{vaswani2017attention} have a complexity of $O(N^2)$.
\nopagebreak
\subsection{HiP-Dreamer}
\begin{algorithm}[H]
   \label{alg:hip_dreamerv1}
    \caption{Hidden Parameter Dreamer (HiP-Dreamer)}
    \begin{algorithmic}[1]
    \Require
        \Statex
        \begin{tabular}{p{0.5\linewidth} p{0.5\linewidth}}
        \textbf{Hyperparameters} & \textbf{Model Parameters} \\
        Seed episodes: $S$ & Context set encoder: $p_{\bm{\theta}} \left( \bm{l} \mid \bm{C_{l}} \right)$ with \\
        Collect interval: $C$ & $\bm{C_{l}} = \{ \left( \bm{o}, \bm{a}, r, \bm{o\prime} \right)_{t} \}_{t=1}^N$ \\
        Batch size: $B$ & Representation: $p_{\bm{\theta}}\left(\bm{s}_t \mid \bm{s}_{t-1}, \bm{a}_{t-1}, \bm{o}_t, \bm{l}\right)$ \\
        Context size: $N$ & Observation: $q_{\bm{\theta}} \left( \bm{o}_t \mid \bm{s}_t, \bm{l} \right)$ \\
        Sequence length: $L$ & Transition: $q_{\bm{\theta}}\left(\bm{s}_t \mid \bm{s}_{t-1}, \bm{a}_{t-1}, \bm{l}\right)$ \\
        Imagination horizon: $H$ & Reward: $q_{\bm{\theta}}\left(r_t \mid \bm{s}_t, \bm{l}\right)$ \\
        Learning rate: $\alpha$ & Actor: $\pi_{\bm{\phi}}\left(\bm{a}_t \mid \bm{s}_t, \bm{l}\right)$ \\
        Trajectory length: $T$ & Critic: $v_{\bm{\psi}}\left(\bm{s}_t, \bm{l}\right)$\\
        Training epochs: $E$ & \\
        Action Repeat: $R$ & \\
        \end{tabular}
        \Statex
    \State \textbf{Initialize} dataset $\mathcal{D}$ with $S$ random seed episodes.
    \State \textbf{Initialize} neural network parameters $\bm{\theta}$, $\bm{\phi}$, $\bm{\psi}$ randomly.
    \While{not converged}
        \For{update step $c = 1$ to $C$}
            \State \textbf{// Dynamics learning}
            \State Draw $\mathcal{B}$ data sequences $\{\left(\bm{o}, \bm{a}, r, \bm{o\prime} \right)_{t}\}_{t=k}^{k+L} \sim \mathcal{D}$ uniformly at random from
            \State the dataset $\mathcal{D}$, with random start index k within an episode.
            \State Draw $\mathcal{B}$ consecutive context chunks $\bm{C_l} = \left\{ \left( \bm{o}, \bm{a}, r, \bm{o\prime} \right)_{t}\right\}_{t=k-N}^{k-1} \sim \mathcal{D}$  
            \State of previous N interaction transitions.
            \State Infer latent task posterior $p_{\bm{\theta}} \left( \bm{l} \mid \bm{C_{l}} \right)$ and sample $\bm{l}$ using reparameterization. 
            \State Use the same latent task sample $\bm{l}$ for the entire sequence length L.
            \State Compute model states $\bm{s}_t \sim p_{\bm{\theta}} (\bm{s}_t \mid \bm{s}_{t-1}, \bm{a}_{t-1}, \bm{o}_t, \bm{l})$.
            \State Update model parameters using derived objective in Section \ref{subsec:learning_adaptive_representations}.
            \State \textbf{// Behavior learning}
            \State Imagine trajectories $\{(\bm{s}_\tau, \bm{a}_\tau)\}_{\tau=t}^{t+H}$ from each $\bm{s}_t$ conditioned on derived task $\bm{l}$.
            \State Use the same latent task samples $\bm{l}$ for the entire imagine horizon H.
            \State Predict rewards $\mathbb{E}\left[q_{\bm{\theta}}\left(r_\tau \mid \bm{s}_{\tau}, \bm{l}\right)\right]$ and values $v_{\bm{\psi}}\left(\bm{s}_\tau, \bm{l}\right)$.
            \State Compute value estimates $V_\lambda(\bm{s}_\tau, \bm{l})$ via Equation 6 from \cite{dreamerv1}.
            \State Update actor parameters $\bm{\phi} \leftarrow \bm{\phi} + \alpha \nabla_{\bm{\phi}} \sum_{\tau=t}^{t+H} V_\lambda(\bm{s}_\tau, \bm{l})$.
            \State Update critic parameters $\bm{\psi} \leftarrow \bm{\psi} - \alpha \nabla_{\bm{\psi}} \sum_{\tau=t}^{t+H} \frac{1}{2} \|v_{\bm{\psi}}(\bm{s}_\tau, \bm{l}) - V_\lambda(\bm{s}_\tau, \bm{l})\|^2$.
        \EndFor
    
        \State \textbf{// Environment interaction}
        \State $o_1 \leftarrow \text{env.reset()}$.
        \State Collect initial context chunk $\bm{C}_{\bm{l}_1}$ using random actions.
        \For{time step $t = 1$ to $\frac{T}{R}$}
            \State Infer latent task posterior $p_{\bm{\theta}} \left( \bm{l}_t \mid \bm{C}_{\bm{l}_{t}} \right)$ and sample $ \bm{l}_t \sim p_{\bm{\theta}} \left( \bm{l}_t \mid \bm{C}_{\bm{l}_{t}} \right)$.
            \State Compute $\bm{s}_t \sim p_{\bm{\theta}}\left( \bm{s}_t \mid \bm{s}_{t-1}, \bm{a}_{t-1}, \bm{o}_t, \bm{l}_t\right)$ from history.
            \State Compute $\bm{a}_t \sim \pi_{\bm{\phi}}\left(\bm{a}_t \mid \bm{s}_t, \bm{l}_t\right)$ with the action model.
            \State Add exploration noise to action $\bm{a}_{t}$.
            \For{action repeat $k = 1..R$}
                \State $r_{t,k}, \bm{o}_{t,k+1} \gets \text{env.step}(\bm{a}_t)$
            \EndFor
            \State $r_t, \bm{o}_{t+1} \gets \sum_{k=1}^{R} \gamma^{k-1} r_{t,k}, \bm{o}_{t,R+1}$
            \State Add experience to dataset $\mathcal{D} \gets \mathcal{D} \cup \{\left(\bm{o}_{t}, \bm{a}_{t}, r_{t}, \bm{o}_{t+1}\right)\}$.
            \State Update context chunk $\bm{C}_{\bm{l}_{t}}$ in a sliding-window manner using $ \{\left(\bm{o}_{t}, \bm{a}_{t}, r_{t}, \bm{o}_{t+1}\right)\}$.
        \EndFor
        \State // \textbf{Agent evaluation}
        \If {T mod E == 0}:
            \State Evaluate agent by calculating the expected return from X episodes.
        \EndIf
    \EndWhile
    \end{algorithmic}
    \end{algorithm}

    The HiP-Dreamer algorithm operates in \textbf{four stages}: 
    
    \textbf{1. Dynamics learning phase} (Lines 6-14): Batch of $\mathcal{B}$ consecutive trajectory data chunks $\{\left(\bm{o}, \bm{a}, r, \bm{o\prime} \right)_{t}\}_{t=k}^{k+L} \sim \mathcal{D}$ is sampled uniformly at random from the replay buffer and used to train the world model, where k indicates a random start index uniformly sampled at random within the episode and is clipped to not exceed the episode length minus the training sequence length. Concurrently, $\mathcal{B}$ consecutive context chunks of  N previous interaction transitions are drawn $\bm{C_l} = \left\{ \left( \bm{o}, \bm{a}, r, \bm{o\prime} \right)_{t}\right\}_{t=k-N}^{k-1} \sim \mathcal{D}$. 
    If fewer than N previous transitions are available, the context is zero-padded at the beginning. 
    The context chunk $\bm{C_{l}}$ is used to infer the latent task posterior distribution $p_{\bm{\theta}} \left( \bm{l} \mid \bm{C_{l}} \right)$ in closed form using Equation 8 from \cite{volpp2021bayesian}.
    To approximate the outer context-dependent expectation of the objective from Section \ref{subsec:learning_adaptive_representations}, a reparameterized sample from the inferred latent task posterior distribution $\bm{l} \sim p_{\bm{\theta}} \left(\bm{l}\mid \bm{C_{l}} \right)$ is used, allowing gradients to flow through the sampling process. 
    The latent task posterior $p_{\bm{\theta}} \left(\bm{l}\mid \bm{C_{l}} \right)$ is inferred once at the start of each training iteration and used for the entire sequence length L.
    All model components (including the context set encoder) are trained end-to-end using Backpropagation-Through-Time (\textbf{BPTT}).

    \textbf{2. Behavior learning phase} (Lines 16-21): The actor chooses actions to predict imagined sequences of compact model states, while the critic accumulates the future predicted rewards beyond the planning horizon. 
    Both actor and critic use learned model states, benefiting from the world model’s representations. 
    Each posterior state inferred during model training serves as an initial state for the actor’s latent trajectory imagination. The latent task posterior $p_{\bm{\theta}} \left( \bm{l} \mid \bm{C_{l}} \right)$ remains consistent across the entire horizon length H = 15, a reasonable assumption for such short horizons.
    The actor aims to output actions that maximize the prediction of long-term future rewards made by the critic. The critic model is trained to regress the $\lambda-\textbf{returns}$, computed as in Equation 8 from \cite{dreamerv1}.
    The world model (including the set encoder) is fixed during behavior learning, so the actor and critic gradients do not affect its representations.

    \textbf{3. Environment interaction phase} (Lines 24-37): Initial context sequence chunk $\bm{C}_{\bm{l}_{1}}$ is collected using random actions and updated in a sliding window manner, incorporating every new collected transition $\left( \bm{o}, \bm{a}, r, \bm{o\prime} \right)_{t}$ to adapt to environmental changes as fast as possible. At each point in time, the context chunk $\bm{C}_{\bm{l}_{t}}$ is used to infer the latent task posterior $\bm{l}_t \sim p_{\bm{\theta}} \left( \bm{l}_{t} \mid \bm{C}_{\bm{l}_{t}}\right)$ and sample $\bm{l}_{t}$, followed by inferring an approximate latent state posterior $\bm{s}_{t} \sim p_{\bm{\theta}}(\bm{s}_t \mid \bm{s}_{t-1}, \bm{a}_{t-1}, \bm{o}_{t}, \bm{l}_{t})$. 
    An action trajectory is generated by conditioning the actor on the inferred latent state $\bm{s}_{t}$ and latent task $\bm{l}_{t}$, repeating the chosen action R times.

    \textbf{4. Evaluation phase} (Lines 39-43): Every E epoch the agent’s performance is assessed over X episodes to estimate the mean return. 
    Unlike data collection, the evaluation phase uses the mean $\bm{\mu}_{\bm{l}_t}$ of the latent task distribution $p_{\bm{\theta}} \left(\bm{l}_{t} \mid \bm{C}_{\bm{l}_{t}}\right)$ enabling deterministic behavior, followed by approximating the latent state posterior $p_{\bm{\theta}}\left(\bm{s}_t \mid \bm{s}_{t-1}, \bm{a}_{t-1}, \bm{o}_{t}, \bm{\mu}_{\bm{l}_t}\right)$, repeating the chosen action R times.
    
\clearpage
\subsection{Hyperparameters.}
\label{sec:hyperparameters}
    This section details only those hyperparameter values that differ from the original architectures or correspond to new architectural components, such as the set encoder. For additional hyperparameters related to all algorithm stages not explicitly mentioned here, we refer to \cite{dreamerv1}.
    
    \paragraph{Set encoder.}
        The set encoder includes a shared fully connected layer with 240 units, followed by two separate fully connected layers of 240 units each. One layer computes the latent task observation $\bm{x}_{n}^{\bm{l}}$, while the other computes the latent task variance $\bm{\sigma}_{n}^{\bm{l}}$. All set encoder layers use the ELU activation function. The latent task posterior $p_{\bm{\theta}} \left(\bm{l} \mid \bm{C}_{\bm{l}}\right)$ is modelled as a multivariate Gaussian with 20 dimensions. Due to computational and time constraints, extensive hyperparameter optimization was not conducted.

    \paragraph{Learning updates.}
        To train the world model, we use batches of 50 sequences of length 50, as in~\cite{hafner2019dream}. For the context data, we sample batches of 50 sequences from the replay buffer, each consisting of 20 prior interaction transitions.
        If fewer than 20 prior transitions are available, we use zero-padding to fill the context data before concatenating it with any available transition data. 
        The set encoder updates use the same learning rate as the world model.

    \paragraph{Objective and Critic.}
        In our objective we use the KL-Balancing technique introduced in \cite{hafner2021mastering}, which we found essential for stabilizing learning when using proprioceptive inputs. 
        Additionally, we experienced instability problems w.r.t. critic network in many environments. To stabilize the critic training further, we use target network to calculate the critic targets, using a soft-update, realized as $\bm{\theta}_{t+1} = \tau \bm{\theta}_{t} + (1 - \tau) \bm{\theta}_{t-1}$ with $\tau = 0.05$  for Gymnasium-based environments. For DMC multi-task benchmarks, we use a hard update, copying the critic’s weights every 100 gradient steps.

\section{Evaluation Details}
    Despite the prevalence of non-stationarity in real-world environments and the existence of some non-stationary benchmarks such as \cite{carl_benchmark} and \cite{dmc_real_world_benchmark}, a detailed categorization of environmental changes remains absent, along with analyses of model-based reinforcement learning agent's adaptability across various non-stationary scenarios.

    We categorize environmental changes into three primary types based on the temporal and structural aspects of the environment’s evolution:
    \begin{itemize}
        \item \textbf{Dynamical Changes:} Alterations affecting the robot’s system or the physical properties of the environment.
        \item \textbf{Objective Changes:} Modifications of objectives, such as changes in target velocity or task requirements. 
        \item \textbf{Combined Changes:} Concurrent occurrences of multiple dynamical changes or a combination of dynamical and objective changes.
    \end{itemize}
    Each category of change can occur either inter-episodically (between episodes) or intra-episodically (within episodes). The temporal dimension specifies when changes occur, while the structural dimension identifies which environmental aspects are affected. Addressing these types of changes enables the design of robust model-based reinforcement learning agents capable of adapting to a broad range of environmental dynamics.

    \subsection{Dynamical Changes}
    \label{subsec:dynamic_changes}
        We implemented various dynamical changes by modifying Gymnasium \cite{towers2024gymnasium} and DMC Control Suite \cite{tunyasuvunakool2020} environments, particularly focusing on Half Cheetah, Walker, and Hopper robots. The dynamical change values were chosen to ensure that MuJoCo’s models remained valid while still challenging pre-trained model-free and model-based agents.
        
        In environments with dynamical changes, the Oracle agent is conditioned on a vector that fully describes the changes. For instance, wind friction on Half Cheetah is represented by a vector of Cartesian forces acting on each body part. Joint perturbations are similarly represented, while actuator masking is indicated by a one-hot vector. In Hopper and Walker, the Oracle receives vectors representing body masses, inertia, and contact friction coefficients, respectively. Details for each environment are provided below:
        \paragraph{Half Cheetah}
        \begin{itemize}
            \item \textbf{Wind Friction}: Random 3D Cartesian forces are applied at the center of mass of each body part, directed either with or against the agent’s movement, sampled from the range $[ -10, 10 ] N$.
            \item \textbf{Joint Perturbation}: Random torques are added to joint torques resulting from the agent’s actions, sampled from the range $[ -20, 20 ] Nm$.
            \item \textbf{Actuator Deactivation}: One of the six actuators is randomly deactivated, meaning the agent’s actions do not affect that actuator.
        \end{itemize}

        \paragraph{Hopper}
            \begin{itemize}
                \item \textbf{Body Mass Inertia}: A random number of body parts is chosen that will be affected by the change and a random scaling factor from $[0.1, 2.5]$ is applied to both mass and inertia, resulting in varying weights and inertia for each affected part.
            \end{itemize}

        \paragraph{Walker}
            \begin{itemize}
                \item \textbf{Contact Friction}: Friction coefficients at ground contact points are modified by randomly selecting new values from $[0.1, 3.9]$. The default feet friction coefficient is 1.9.
            \end{itemize}

        The environmental changing values are sampled always randomly from the predefined value ranges. 
        For these types of dynamic changes, a vanilla agent may still adapt because the optimal behavior’s state region remains the same, what changes is how the agent reaches it.

\clearpage
\subsection{Objective changes}
\label{subsec:evaluation_objective_changes}

    \paragraph{Half Cheetah, Hopper, and Walker}
        \begin{itemize}
            \item Target Velocity: A target velocity is randomly chosen within the range $[-6, 6] \frac{m}{s}$.
        \end{itemize}

    In these environments, the Oracle agent receives the target velocity as a scalar.
    
    \paragraph{DMC Multi-Task Benchmarks}
        \begin{itemize}
            \item Objective Change: At the beginning of each episode, a new objective is selected, requiring the agent to acquire multiple skills within the same domain.  
        \end{itemize}

        We implement these multi-task benchmarks using pre-defined tasks from \cite{hansen2024tdmpc}, creating a variety of benchmarks across different domains and categorizing them by task count, as shown in Table \ref{table:dmc_multi_task_benchmarks}.

        In all DMC multitask environments, the Oracle agent receives a unique one-hot vector corresponding to each environment.
        
    \begin{table}[H]
        \caption{Multi-task benchmarks based on Cheetah, Walker, Ball-In-Cup, and Pendulum models.}
        \label{table:dmc_multi_task_benchmarks}
        \centering
        \setlength{\arrayrulewidth}{0.7mm}  
        \setlength{\tabcolsep}{8pt}        
        \renewcommand{\arraystretch}{1.0}  
    
        \begin{tabular}{| l | l | l | l |}
            \hline
            \rowcolor{headercolor}
             \textbf{Cheetah Tasks} & \textbf{Walker Tasks} & \textbf{Cup Tasks} & \textbf{Pendulum Tasks}\\
            \hline
            Stand Front      & Walk Backwards & Catch & Swingup\\
            Stand Back       & Run Backwards & Spin & Spin \\
            Run Front        & Walk & & \\
            Run Back         & Run & & \\
            \hline
            Run Backwards    & Walk Backwards & & \\
            Stand Front      & Arabesque & &  \\
            Stand Back       & Lie Down & &  \\
            Jump             & Legs Up & & \\
            Run Front        & Head Stand & & \\
            Run Back         & Walk & & \\
            Lie Down         & Flip & & \\
            Legs Up          & Backflip & & \\
            Flip             & & & \\
            Flip Backwards   & & & \\
            \hline
        \end{tabular}
    \end{table}

\clearpage
\subsection{Further Evaluation Under Dynamical Changes}
\label{appen:evaluation_dynamical_changes}
    This section presents additional evaluations of environments with both inter-episodic and intra-episodic dynamical changes.

    \begin{figure}[htbp!]
        \centering
        \includegraphics[width=1.0\linewidth]{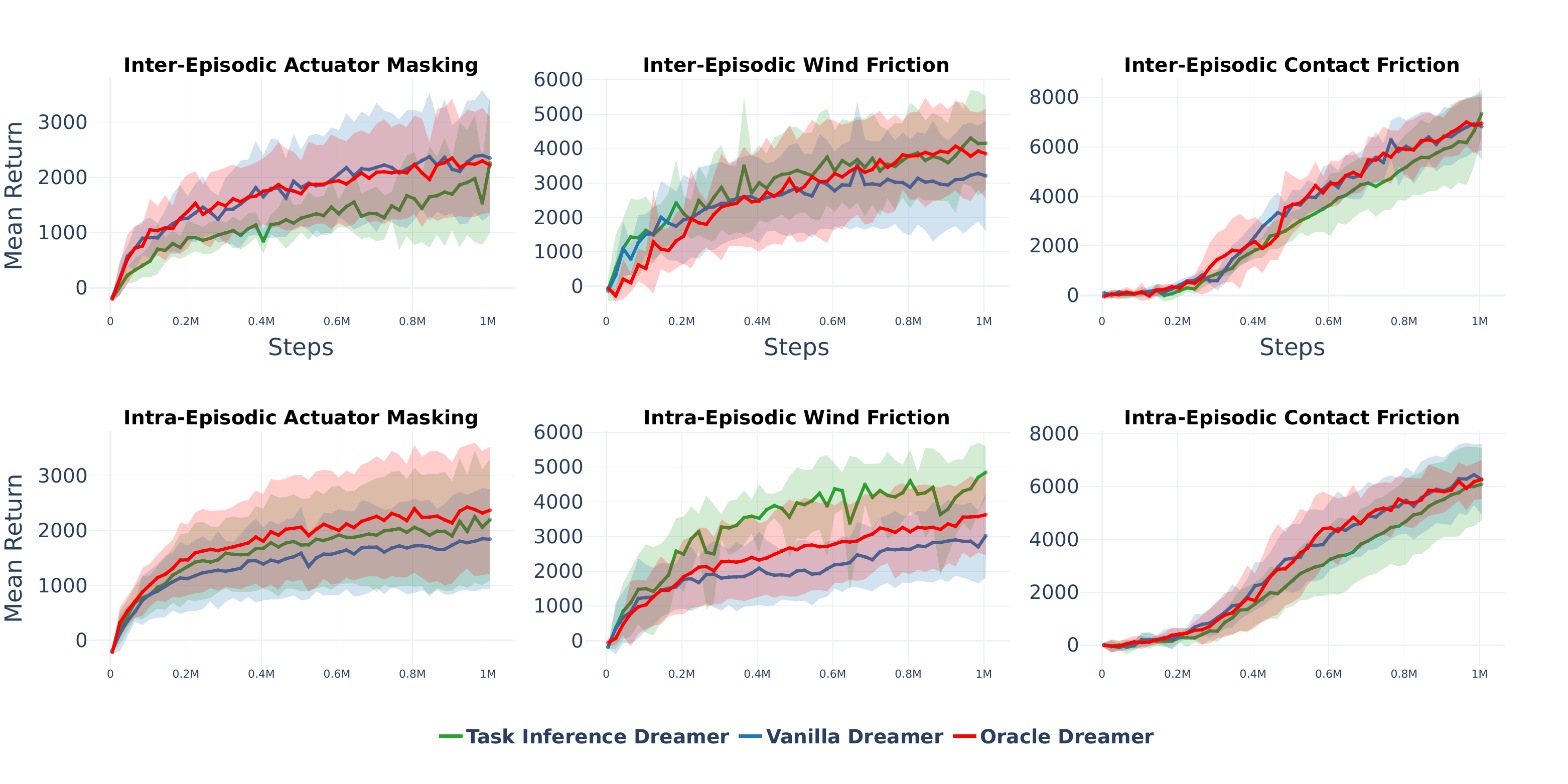}
        \caption{Dynamical changes on HalfCheetah and Walker. The first row shows inter-episodic dynamical changes, whereas the second row shows intra-episodic dynamical changes with a frequency of 200 environmental steps.}
        \label{fig:additional_inter_intra_halfcheetah_walker_dynamical_changes}
    \end{figure}
    
    Figure \ref{fig:additional_inter_intra_halfcheetah_walker_dynamical_changes} illustrates additional experiments involving dynamical changes in the environment. Across both inter- and intra-episodic changes, task-conditioned agents exhibit similar performance to the vanilla agent, with only minor differences observed. 

    On one hand, this result suggests that the vanilla agent is capable of adapting to various dynamic changes, indicating that approaches based on the POMDP formalism can indeed learn representations that support adaptive behavior. On the other hand, the HiP-POMDP formalism generally performs comparably or slightly better, highlighting the benefits and expressiveness of a learned latent task representation.

    Interestingly, in some cases, the task-inference agent even outperforms the Oracle agent, despite the former having to learn and utilize a task representation for each task, while the Oracle agent is directly provided with task change information. This advantage may be due to the learned task representation capturing additional task-relevant information beyond the task changes alone, or it could suggest that the ground-truth task representation provided to the Oracle agent is not optimally structured for adaptation. This observation warrants further investigation in future work. 

\newpage
\subsection{Further Evaluation Under Objective Changes}
\label{appen:objective_changes}

    In this section, we conduct additional experiments under various reward/objective changes and identify the points at which different algorithms begin to fail.
    
    \begin{figure}[htbp!]
        \centering
        \begin{subfigure}{0.6\textwidth}
            \centering
            \includegraphics[width=\linewidth]{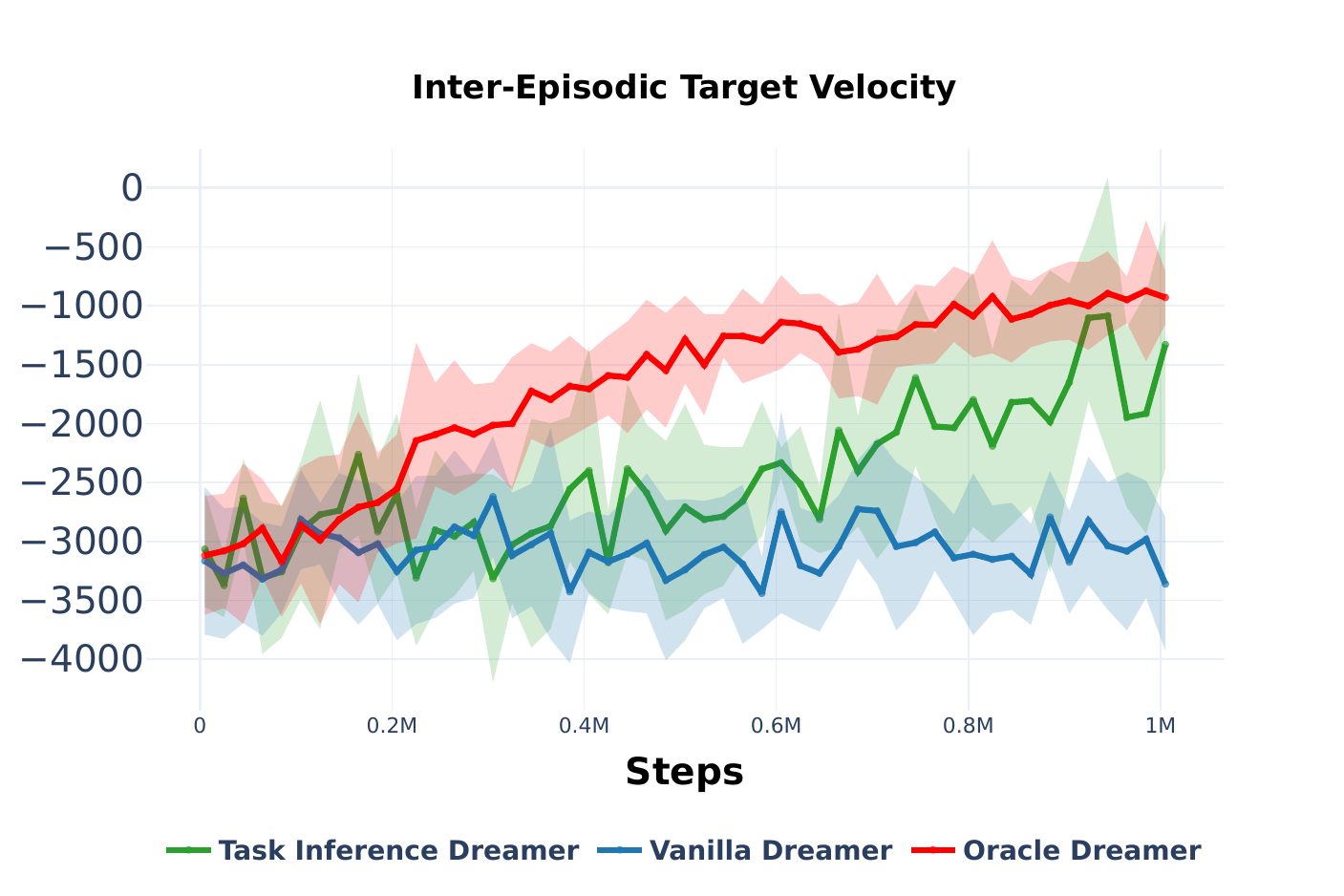}        
            \label{fig:inter_episodic_hopper_target_velocity_every_200}
        \end{subfigure}
    \caption{Inter-Episodic target velocity change on Hopper.}
    \label{fig:additional_inter_hopper_objective_changes}
    \end{figure}

    Figure \ref{fig:additional_inter_hopper_objective_changes} provides further empirical evidence that the vanilla agent struggles to handle objective changes, while task-conditioned agents demonstrate adaptive behavior, consistent with the results shown in Figure \ref{fig:inter_halfcheetah_walker_objective_changes}.

    \paragraph{Breaking point of Vanilla Dreamer under objective changes.}
    
    \begin{figure}[htbp!]
        \centering \includegraphics[width=0.6\linewidth]{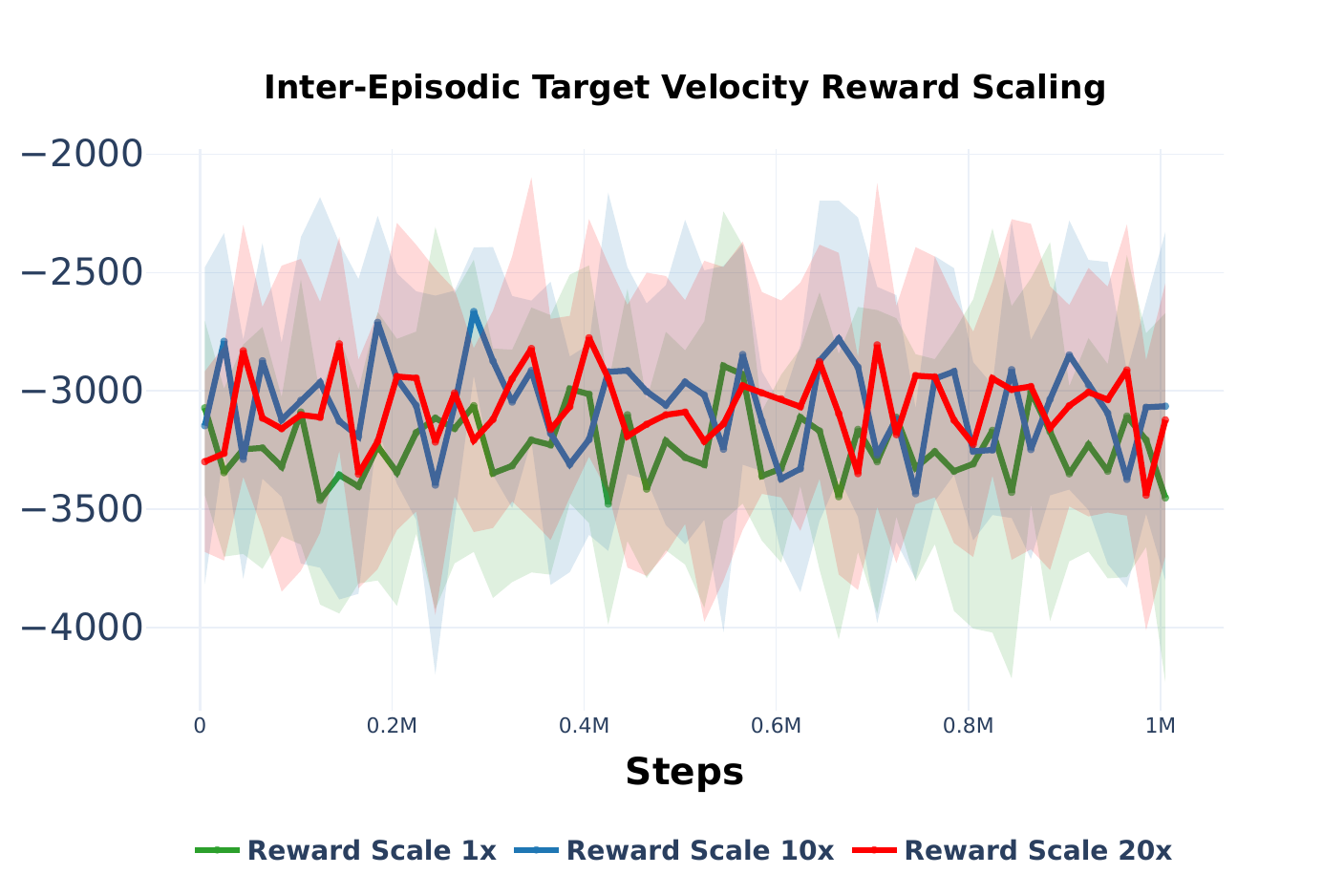}
        \caption{Vanilla Dreamer agent under inter-episodic reward change on Half Cheetah with different reward loss scaling factors.} \label{fig:additional_intra_halfcheetah_objective_changes} 
    \end{figure}

    To investigate the vanilla agent’s failures under target reward-changing experiments, we adjust the reconstruction loss by giving more weight to reward reconstruction to offset potential higher loss weights from the multi-dimensional observations. Figure \ref{fig:additional_intra_halfcheetah_objective_changes} illustrates the effect of scaling the reward reconstruction loss differently.

    Our observations show that increasing the reward reconstruction weight factor also raises the KL divergence between the prior and posterior distributions over the latent state, along with an increased observation reconstruction loss, ultimately degrading performance. These findings suggest the optimization process struggles to balance reconstruction and regularization loss terms, possibly leading to overfitting to specific rewards within each mini-batch.

    \newpage
    \paragraph{Breaking point of Task-inference Dreamer under objective changes.}
    
    \vspace{-20pt}
    \begin{figure}[htbp!]
        \centering
        \includegraphics[width=1.0\linewidth]{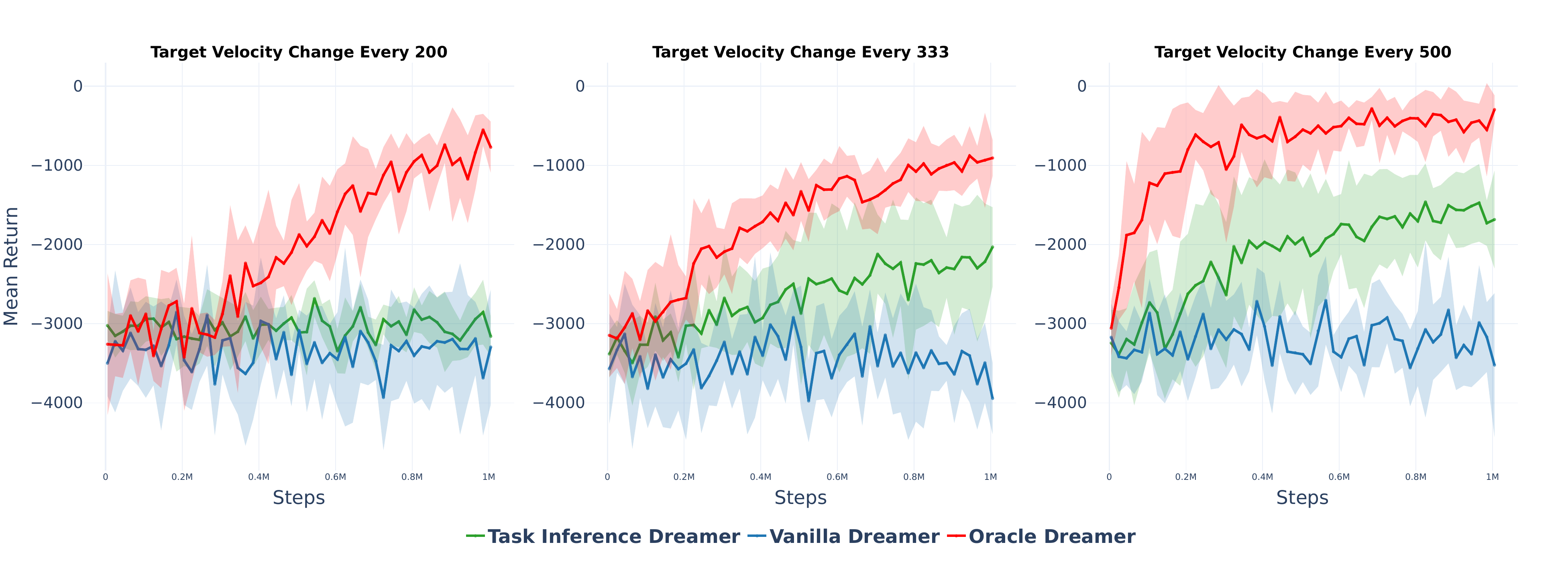}
        \caption{Intra-episodic target velocity change on Half Cheetah. The change frequency is reduced from left to right.}
        \label{fig:objective_frequency_change}
    \end{figure}

    Figure \ref{fig:objective_frequency_change} highlights a breaking point in the latent task inference mechanism when the target velocity changes every 200 environmental steps. We hypothesize that the standard latent task aggregation requires more time to infer a new task belief accurately, a necessity for effective adaptation, especially when target velocity undergoes drastic changes near boundary values.
    To address this, we are currently experimenting with a modified latent task update mechanism that reduces the influence of older transitions, introducing a “forgetting” effect that could help the agent adapt more rapidly to abrupt task shifts.

    \paragraph{DMC-Multitask benchmarks}
    We also assess all agents under more complex objective change settings, where each agent must learn multiple skills simultaneously. Figure \ref{fig:inter_episodic_dmc_multi_task_objective_changes} shows results on a variety of different custom-designed multi-task benchmarks.

    \vspace{-5pt}
    \begin{figure}[H]
        \centering
        \includegraphics[width=1.0\linewidth]{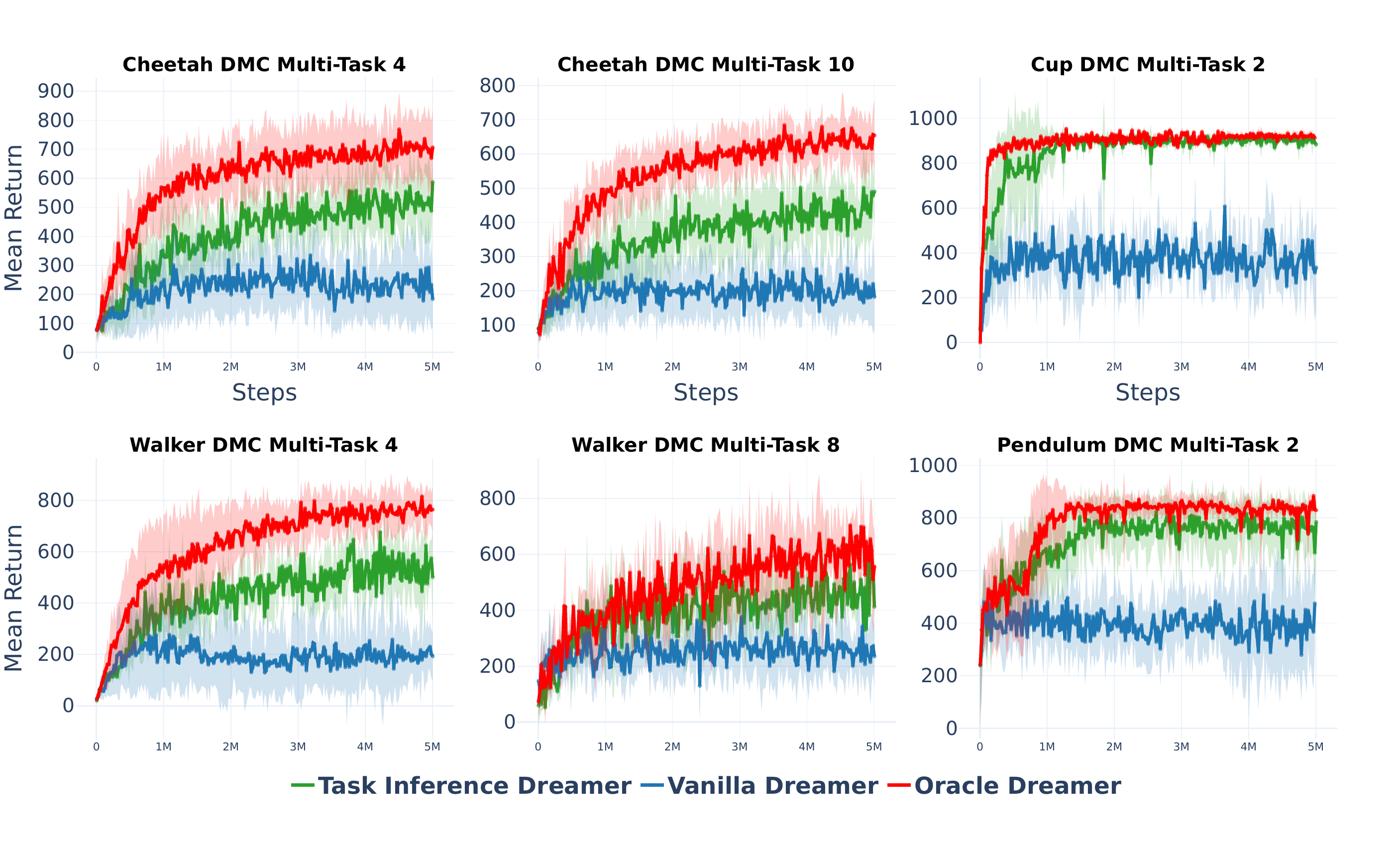}
        \vspace{-5pt}
        \caption{Inter-episodic objective changes on DMC Multi-task benchmarks on Half Cheetah, Walker, Cup, and Pendulum domains. Each number indicates the number of tasks in each experiment.}
        \label{fig:inter_episodic_dmc_multi_task_objective_changes}
    \end{figure}
    \vspace{-10pt}

    The vanilla agent struggles to learn multiple skills simultaneously. We hypothesize this limitation arises from multiple interferences in the learned latent state, resulting in a task-independent latent space structure. 
    In such cases, the reward predictor cannot estimate rewards accurately, leading to suboptimal performance.
    However, providing task information enables better multi-skill learning across all benchmarks, potentially creating a more structured latent space. Additional evidence for these hypotheses is presented in Appendix \ref{appen:latent_state_task_space_visualizations}.

\subsection{Evaluation under Combined Environmental Changes}
\label{appen:combined_changes}
In this section, we evaluate all agents in the most challenging setting, involving both multiple dynamical changes and combined dynamical and objective changes. Figure \ref{fig:inter_multi_modal_changes} illustrates the evaluation results for all agents on the Half Cheetah task, where changes evolve in an inter-episodic manner.

    \begin{figure}[H]
        \centering
        \includegraphics[width=1.0\linewidth]{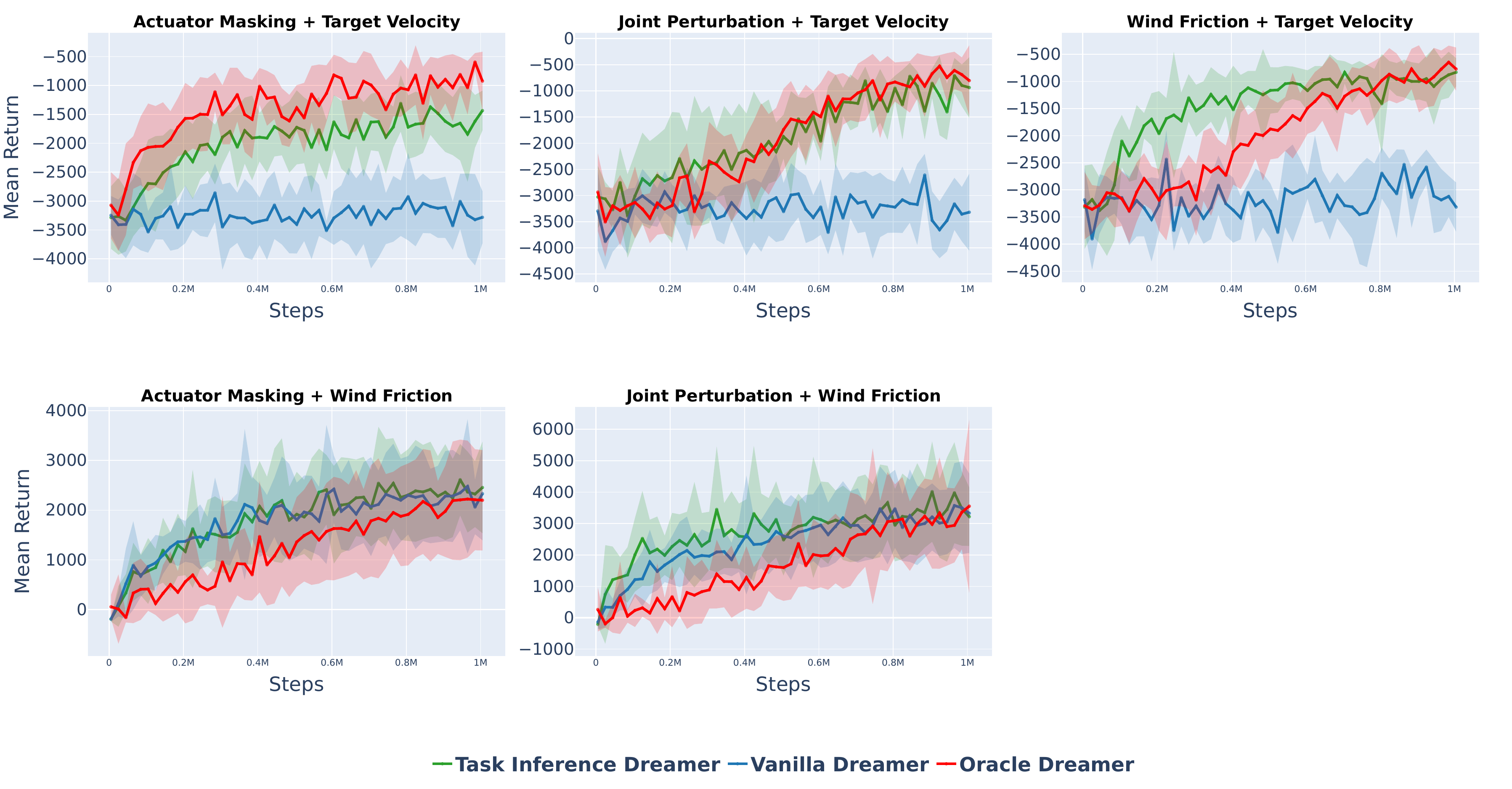}
        \caption{Multiple changes on HalfCheetah. The first row shows combined dynamical and objective changes, whereas the second row shows combined dynamical changes.}
        \label{fig:inter_multi_modal_changes}
    \end{figure}

    As seen in Figure \ref{fig:inter_multi_modal_changes}, the vanilla agent demonstrates adaptability under multi-modal dynamical changes, however, it struggles with combined dynamical and objective changes, largely due to its limited ability to handle shifts in objectives. 
    
    Introducing a task representation notably enhances performance, as both the task-inference and oracle agents successfully adapt across all combined change scenarios. However, under scenarios involving only combined dynamical changes, the oracle agent exhibits slower learning compared to the vanilla agent. This slower adaptation raises questions about the potential influences of the task representation, warranting further investigation.

\subsection{Latent Space Visualizations}
\label{appen:latent_state_task_space_visualizations}

    This section presents 2D projections of both the world model's latent state and the latent task spaces. The visualizations are generated by recording trajectories of posterior latent states (incorporating both deterministic and stochastic components) and latent task representations during evaluations across various tasks, with a final 2D projection achieved using t-SNE \cite{JMLR:v9:vandermaaten08a}.

    The primary objective of these visualizations is to explore two key questions: (1) Does conditioning the MBRL agent on either a ground truth or learned task representation introduce any structural changes in the latent state space? (2) How does the structure of the latent state space correlate with the agent’s performance?
        
    \subsubsection{Dynamics Changes}
    \label{appen:latent_visualizations_dynamical_changes}
        Figure \ref{fig:latent_visualizations_dynamical_changes} displays 2d projections of the latent state space under different dynamic environmental conditions.
        
        \begin{figure}[h]
            \centering
            \begin{subfigure}{\textwidth}
                \centering
                \begin{minipage}[b]{0.27\textwidth}
                    \centering
                    \scalebox{0.6}{\parbox{1.5\linewidth}{\centering \textbf{Task Inference Dreamer \\ Latent Task Space}}}
                    \vspace{-0.5em}  
                    \includegraphics[width=\textwidth]{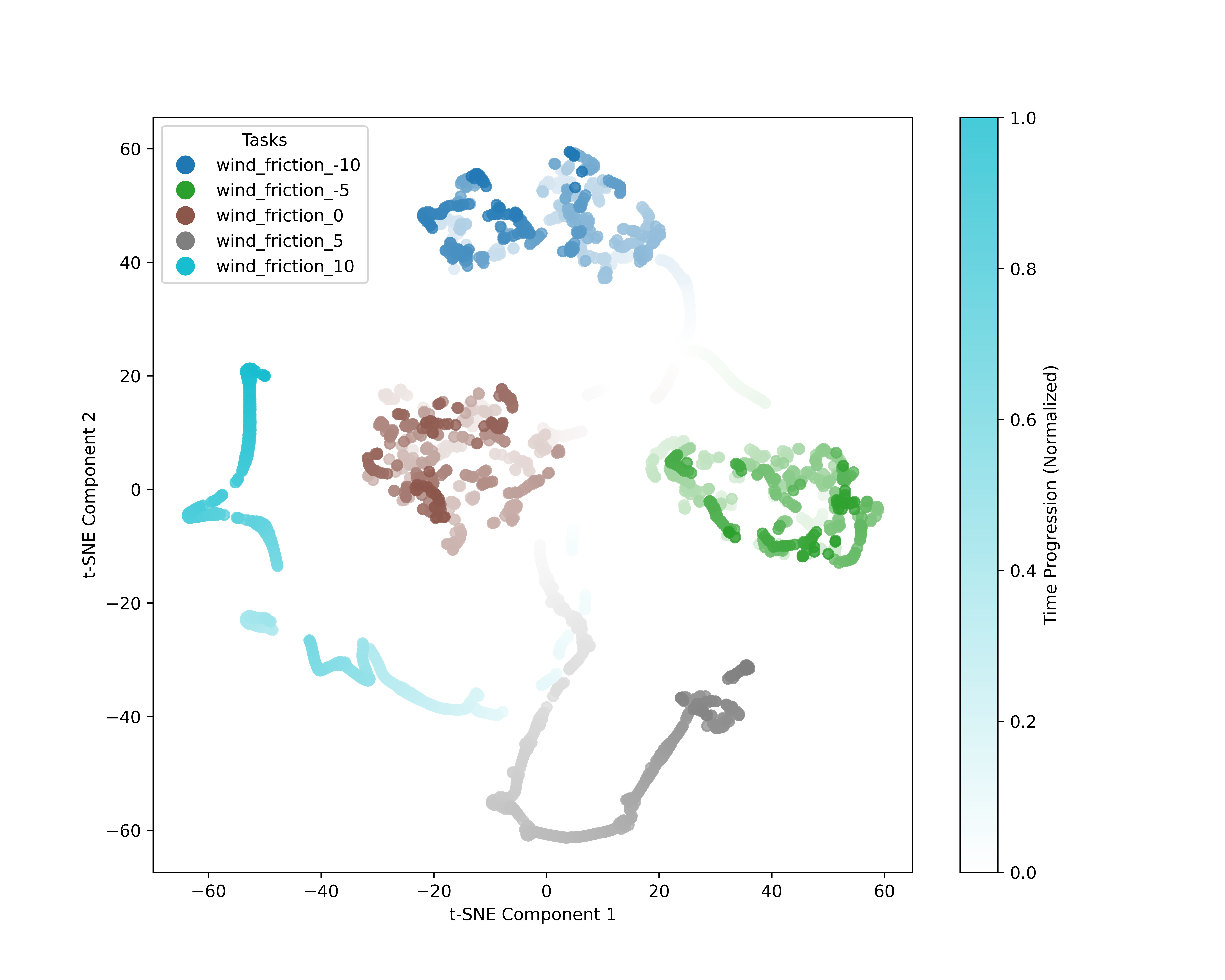}
                \end{minipage}
                \hspace{-18pt}
                \begin{minipage}[b]{0.27\textwidth}
                    \centering
                    \scalebox{0.6}{\parbox{1.5\linewidth}{\centering \textbf{Task Inference Dreamer \\ Latent State Space}}}
                    \vspace{-0.5em}  
                    \includegraphics[width=\textwidth]{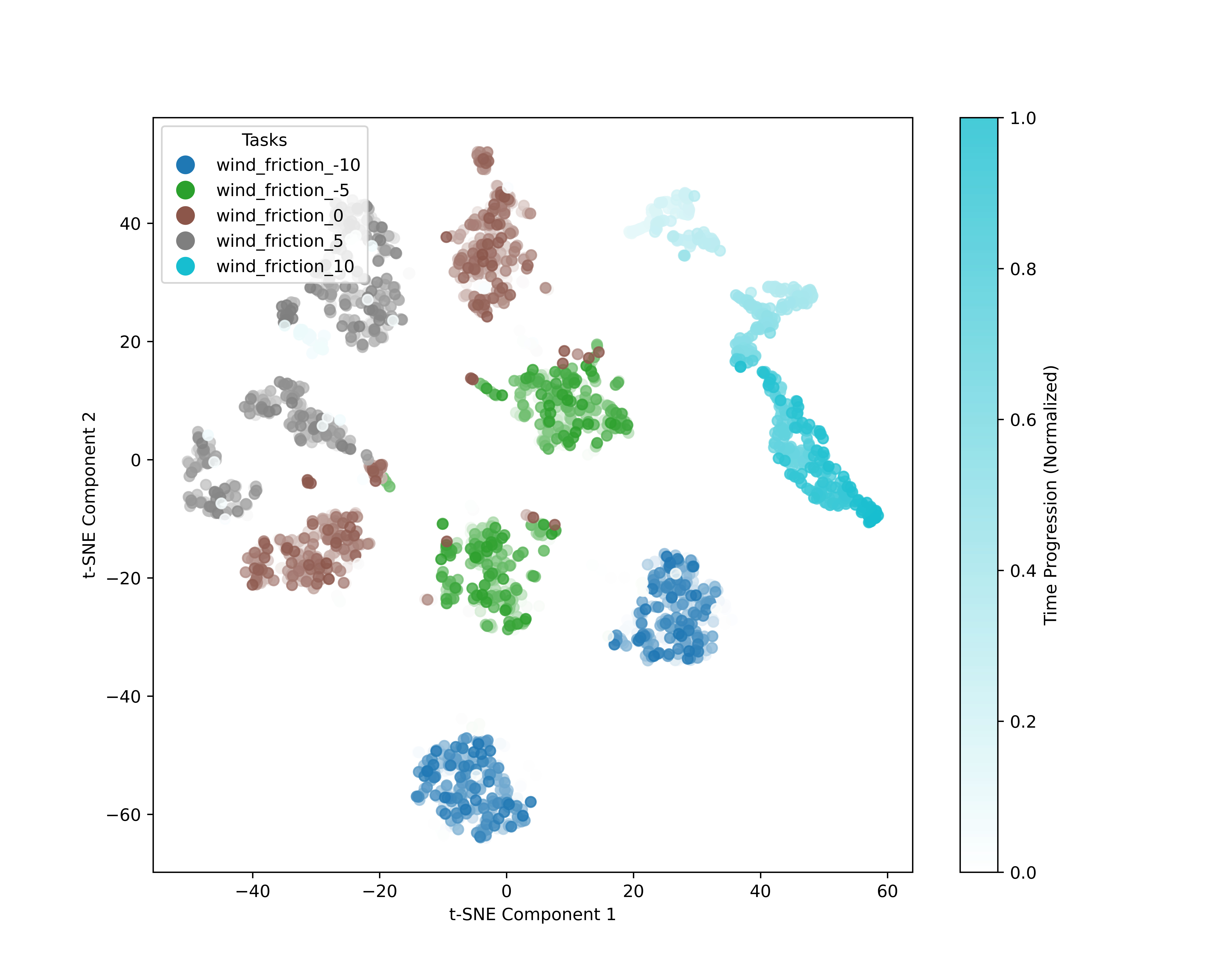}
                \end{minipage}
                \hspace{-18pt}
                \begin{minipage}[b]{0.27\textwidth}
                    \centering
                    \scalebox{0.6}{\parbox{1.5\linewidth}{\centering \textbf{Vanilla Dreamer \\ Latent State Space}}}
                    \vspace{-0.5em}  
                    \includegraphics[width=\textwidth]{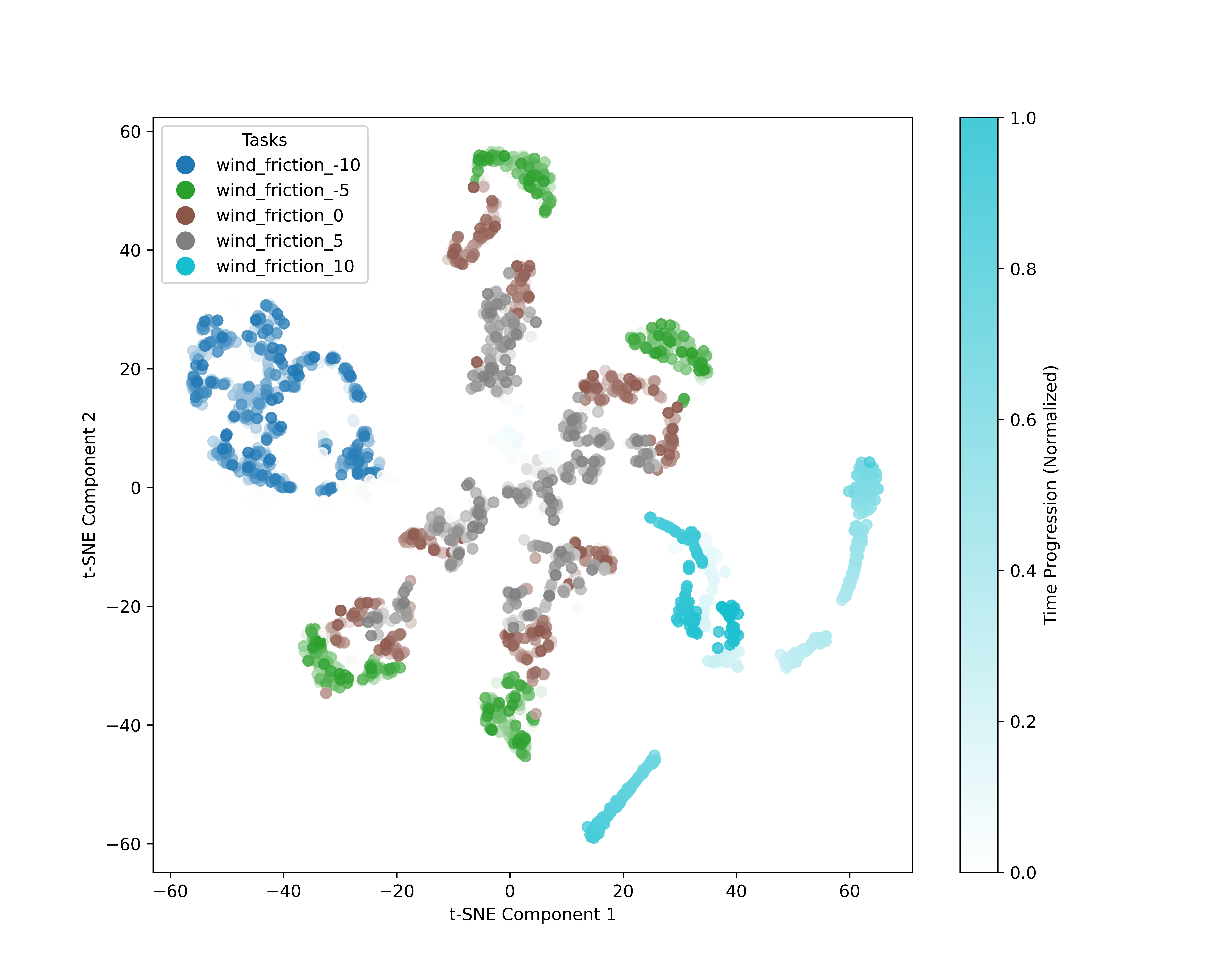}
                \end{minipage}
                \hspace{-18pt}
                \begin{minipage}[b]{0.27\textwidth}
                    \centering
                    \scalebox{0.6}{\parbox{1.5\linewidth}{\centering \textbf{Oracle Dreamer \\ Latent State Space}}}
                    \vspace{-0.5em}  
                    \includegraphics[width=\textwidth]{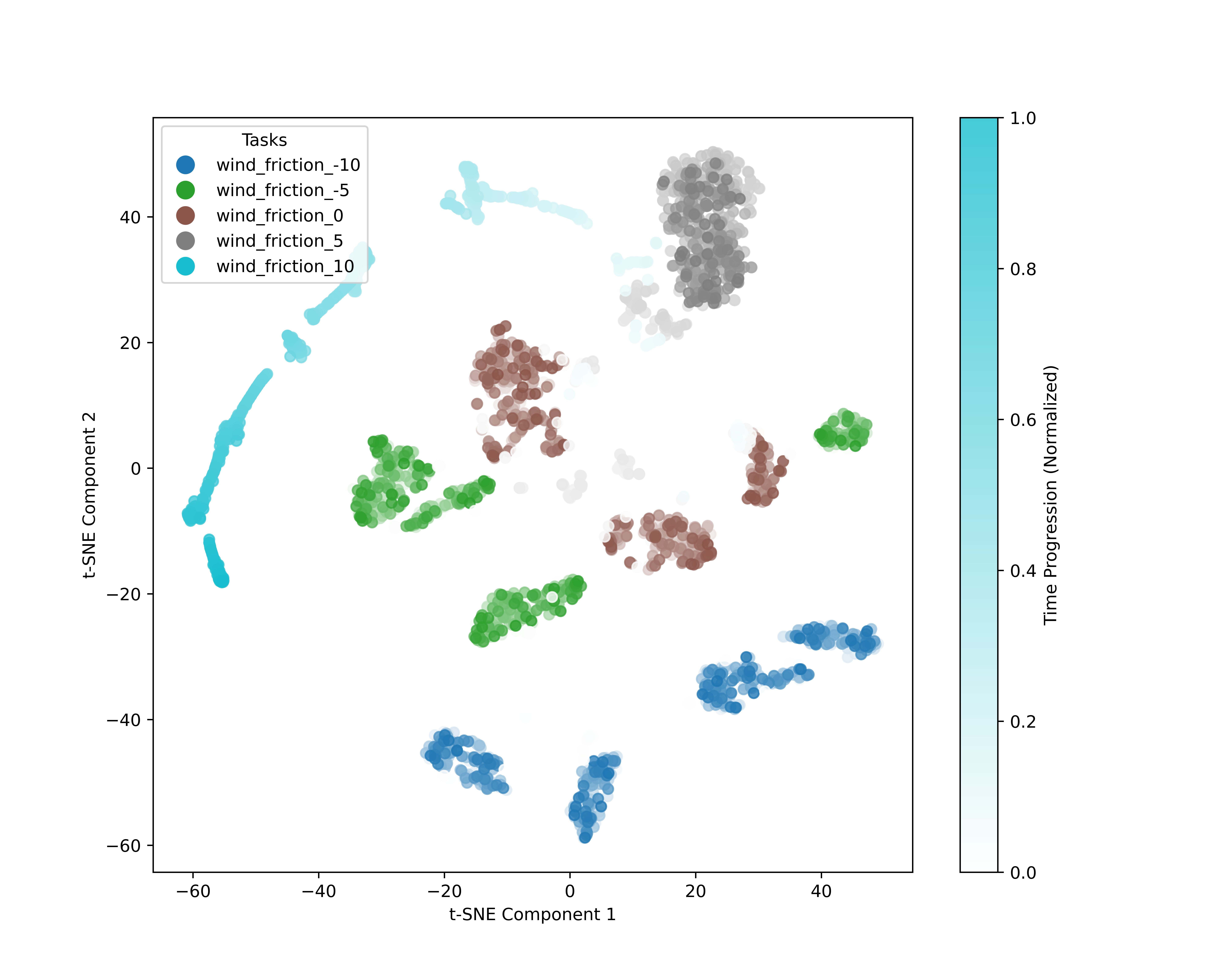}
                \end{minipage}
                \caption{ Half Cheetah: wind friction.}
            \end{subfigure}
            
            \begin{subfigure}{\textwidth}
                \centering
                \begin{minipage}[b]{0.27\textwidth}
                    \centering
                    \includegraphics[width=\textwidth]{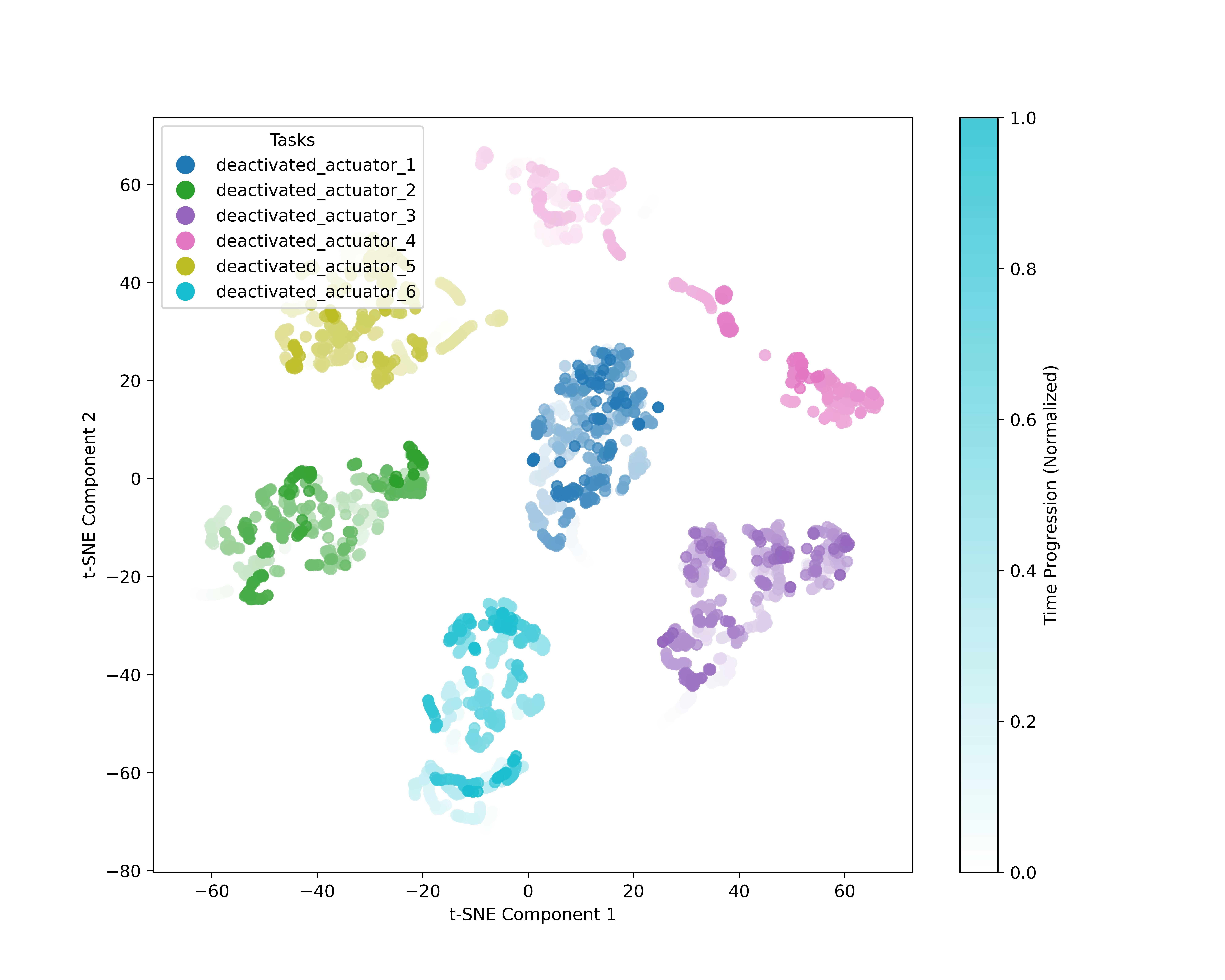}
                \end{minipage}
                \hspace{-18pt}
                \begin{minipage}[b]{0.27\textwidth}
                    \centering
                    \includegraphics[width=\textwidth]{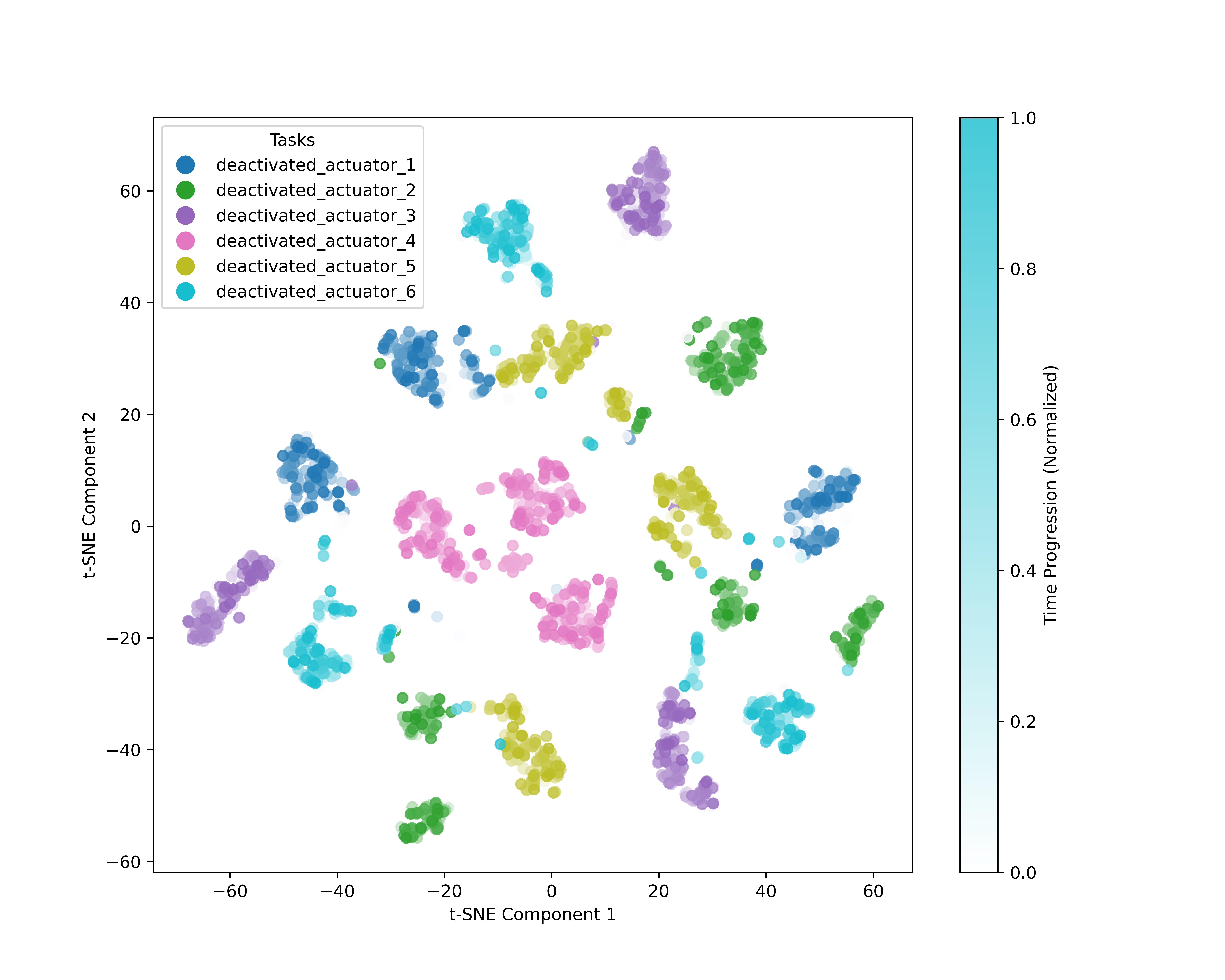}
                \end{minipage}
                \hspace{-18pt}
                \begin{minipage}[b]{0.27\textwidth}
                    \centering
                    \includegraphics[width=\textwidth]{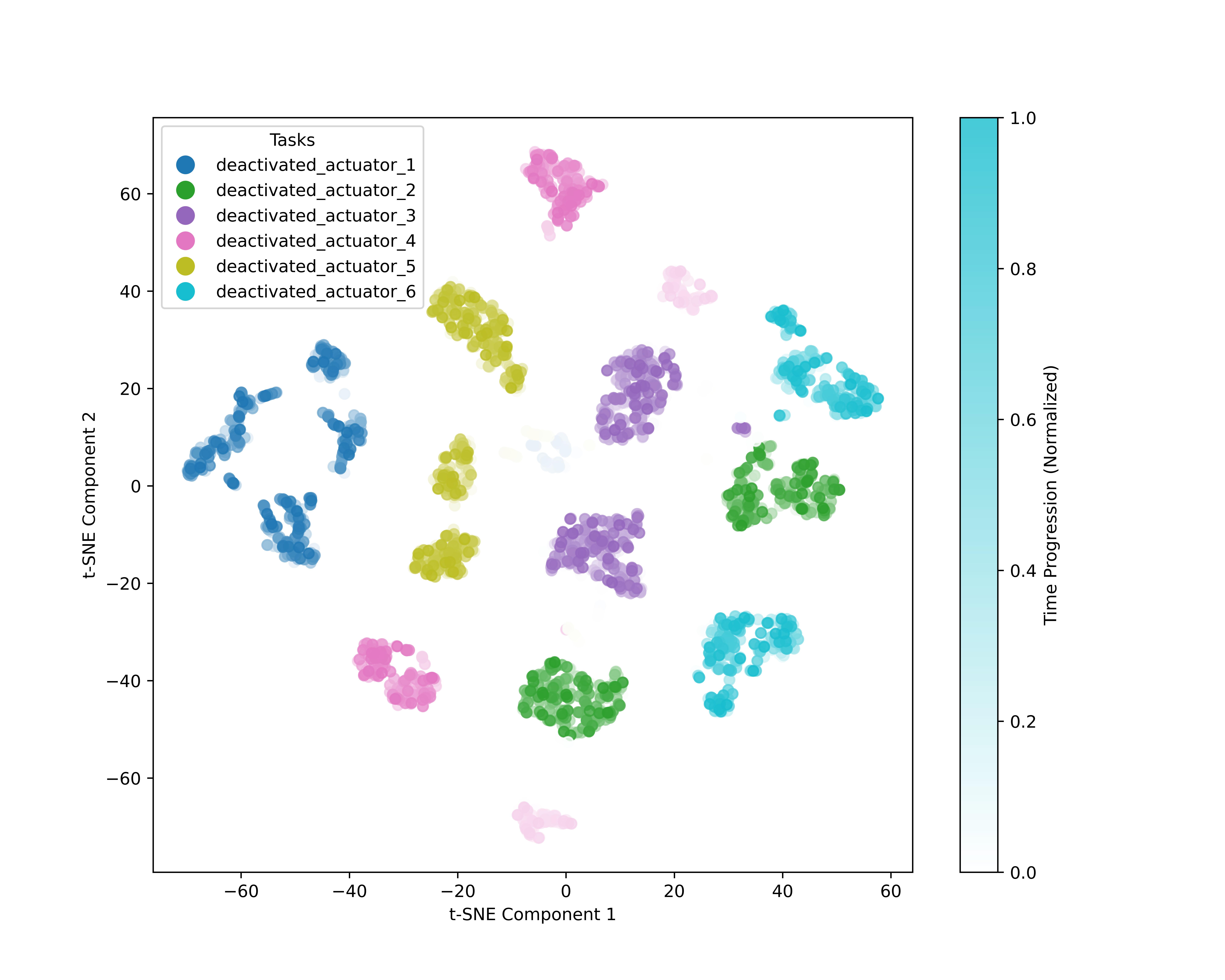}
                \end{minipage}
                \hspace{-18pt}
                \begin{minipage}[b]{0.27\textwidth}
                    \centering
                    \includegraphics[width=\textwidth]{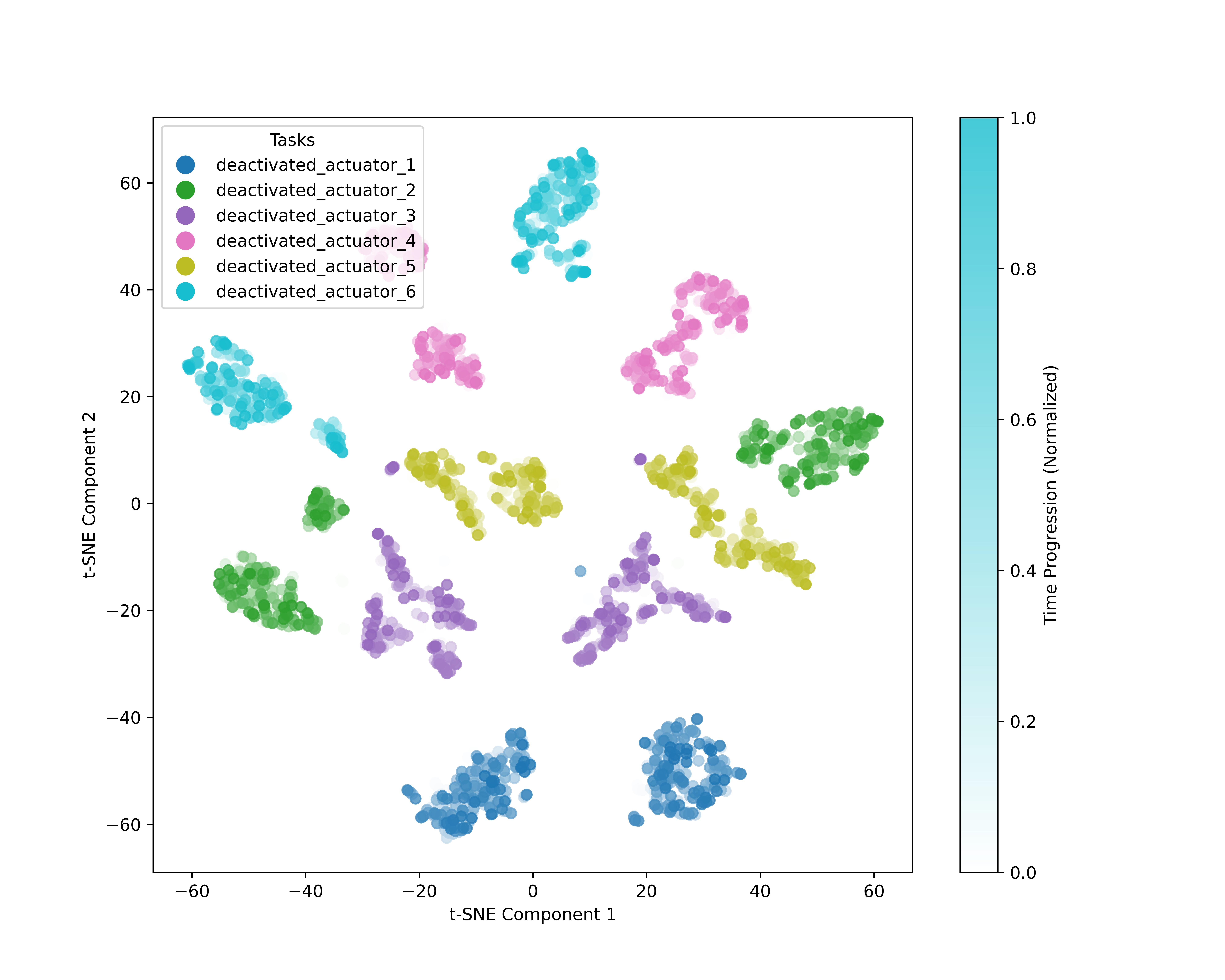}
                \end{minipage}
                \caption{Half Cheetah: actuator masking.}
            \end{subfigure}
                    
            \begin{subfigure}{\textwidth}
                \centering
                \begin{minipage}[b]{0.27\textwidth}
                    \centering
                    \includegraphics[width=\textwidth]{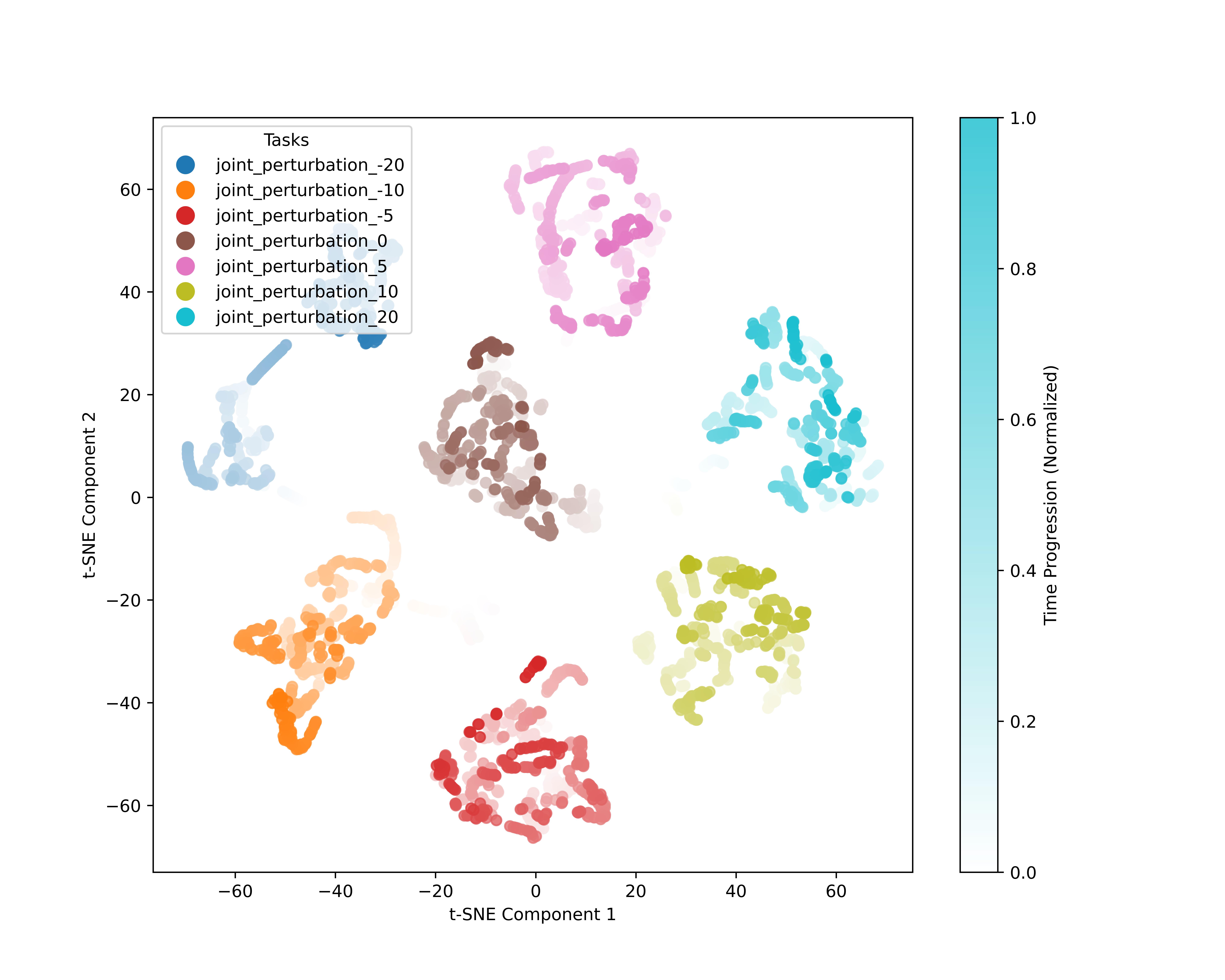}
                \end{minipage}
                \hspace{-18pt}
                \begin{minipage}[b]{0.27\textwidth}
                    \centering
                    \includegraphics[width=\textwidth]{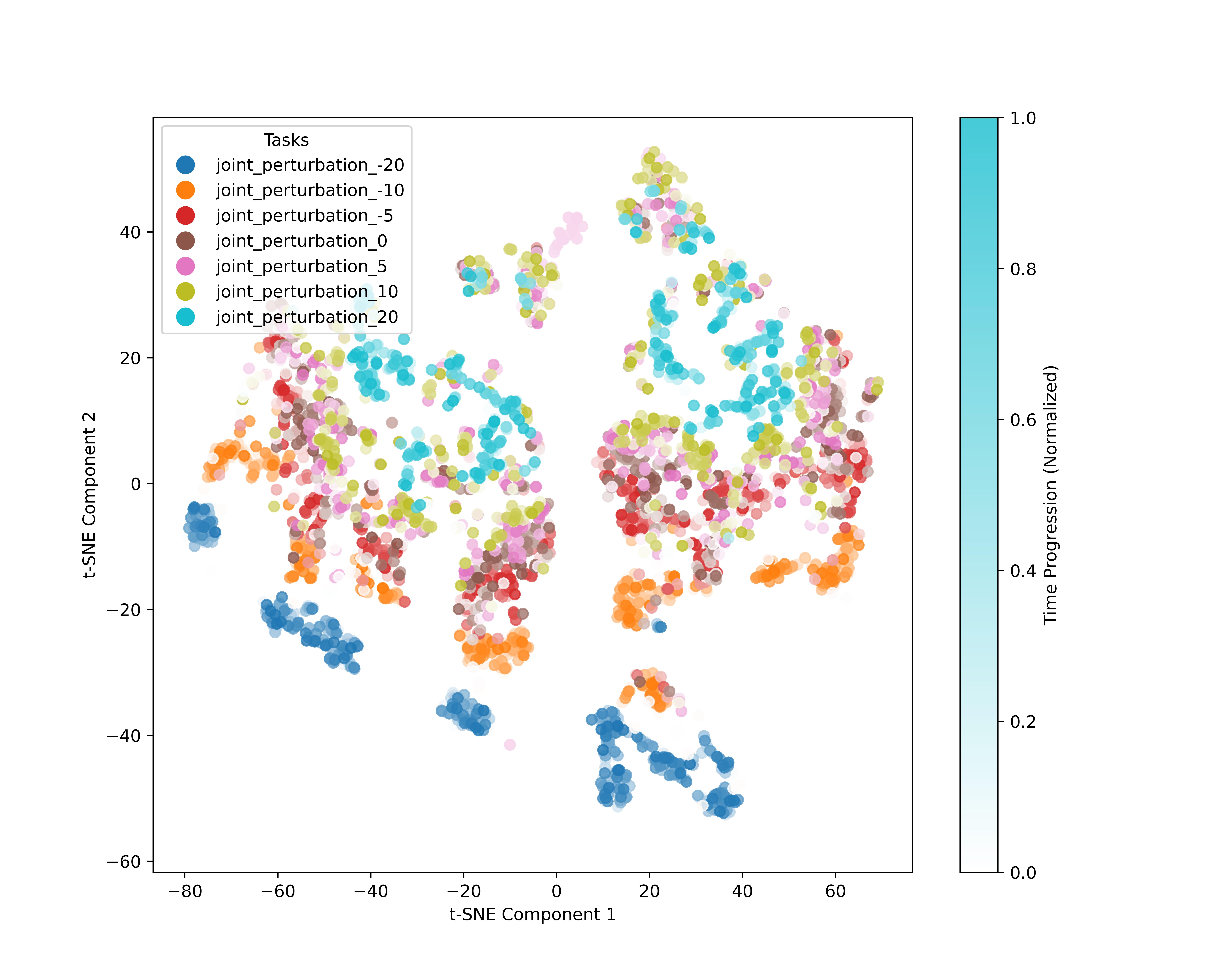}
                \end{minipage}
                \hspace{-18pt}
                \begin{minipage}[b]{0.27\textwidth}
                    \centering
                    \includegraphics[width=\textwidth]{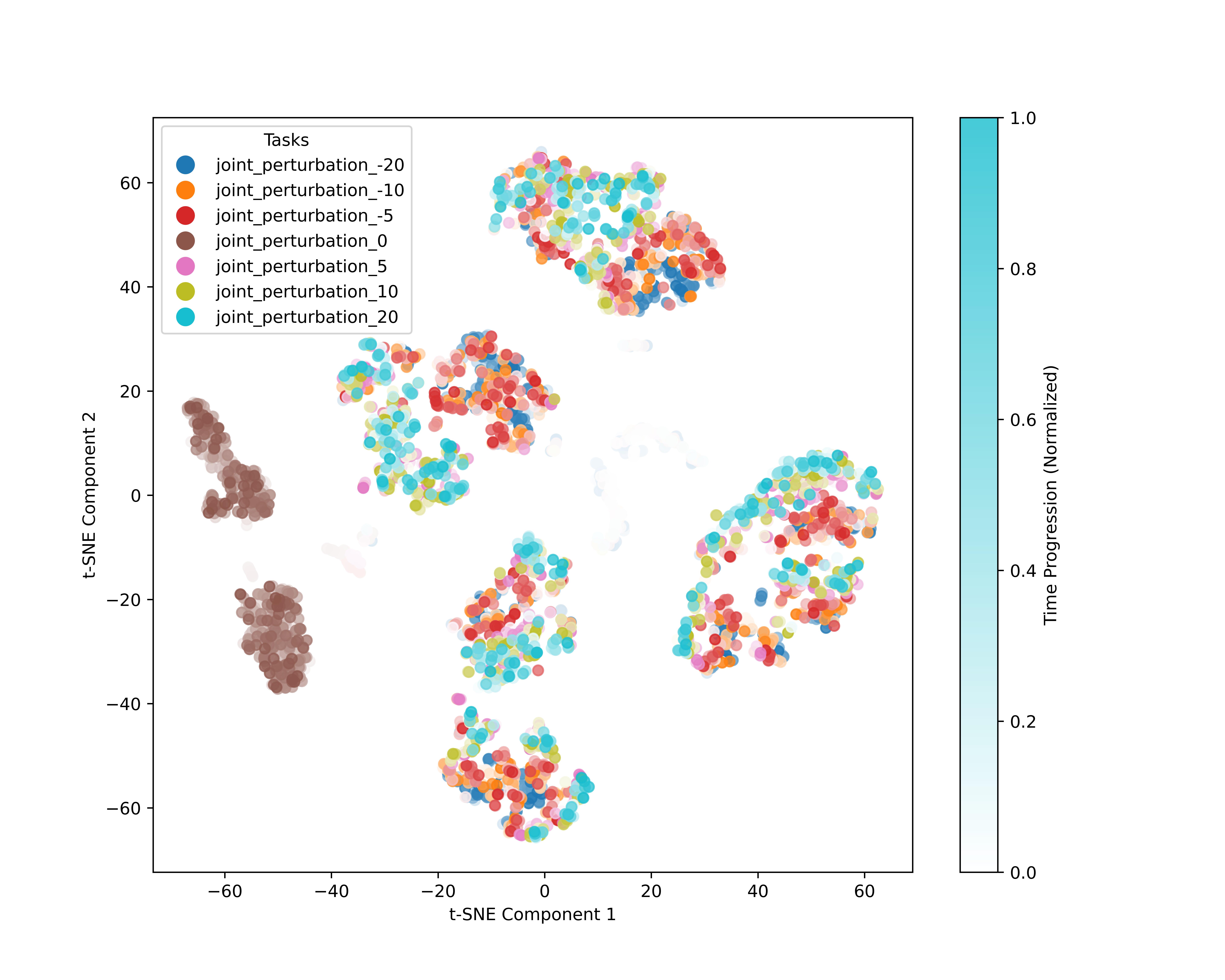}
                \end{minipage}
                \hspace{-18pt}
                \begin{minipage}[b]{0.27\textwidth}
                    \centering
                    \includegraphics[width=\textwidth]{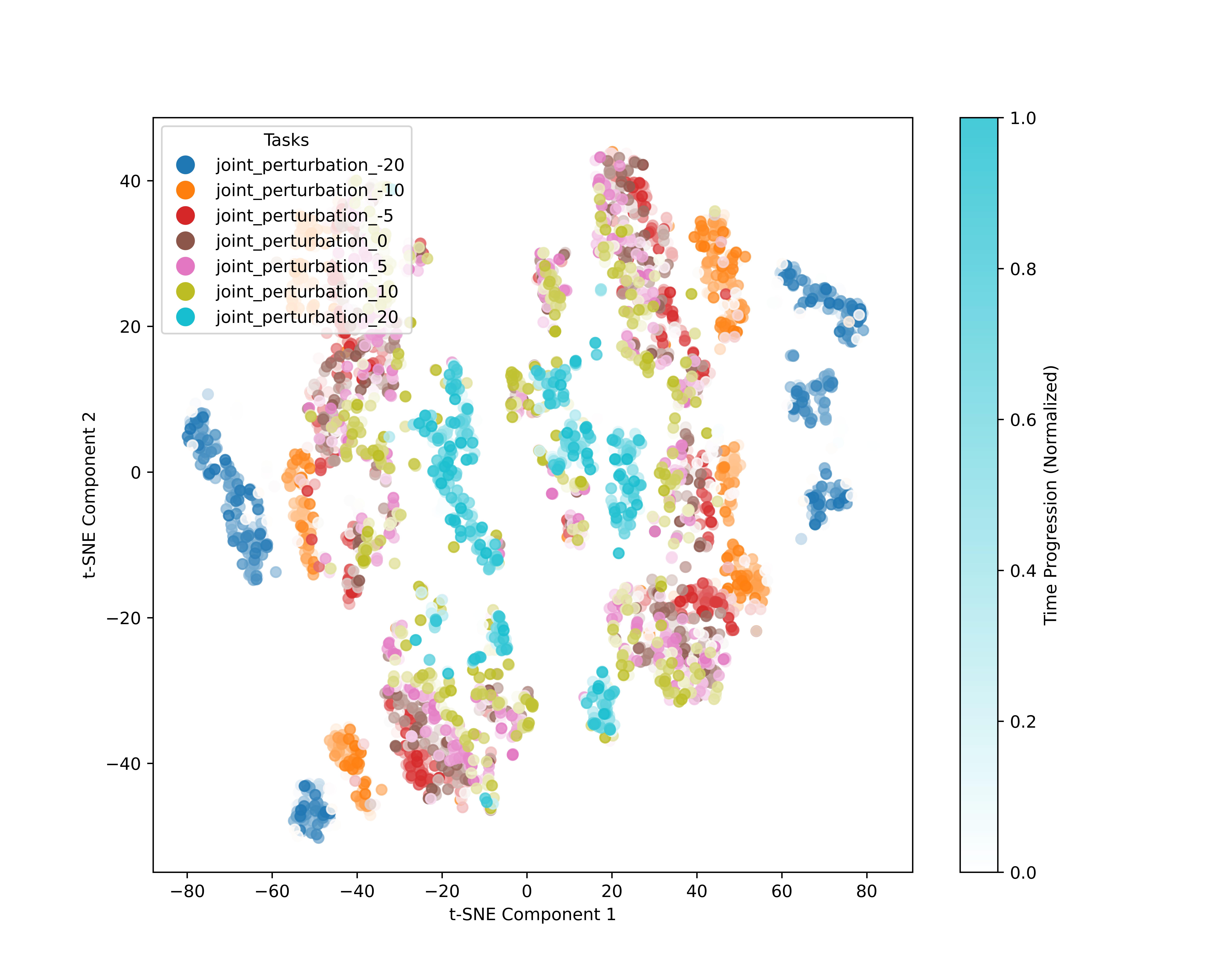}
                \end{minipage}
                \caption{Half Cheetah: joint perturbation.}
            \end{subfigure}
                    
            \begin{subfigure}{\textwidth}
                \centering
                \begin{minipage}[b]{0.27\textwidth}
                    \centering
                    \includegraphics[width=\textwidth]{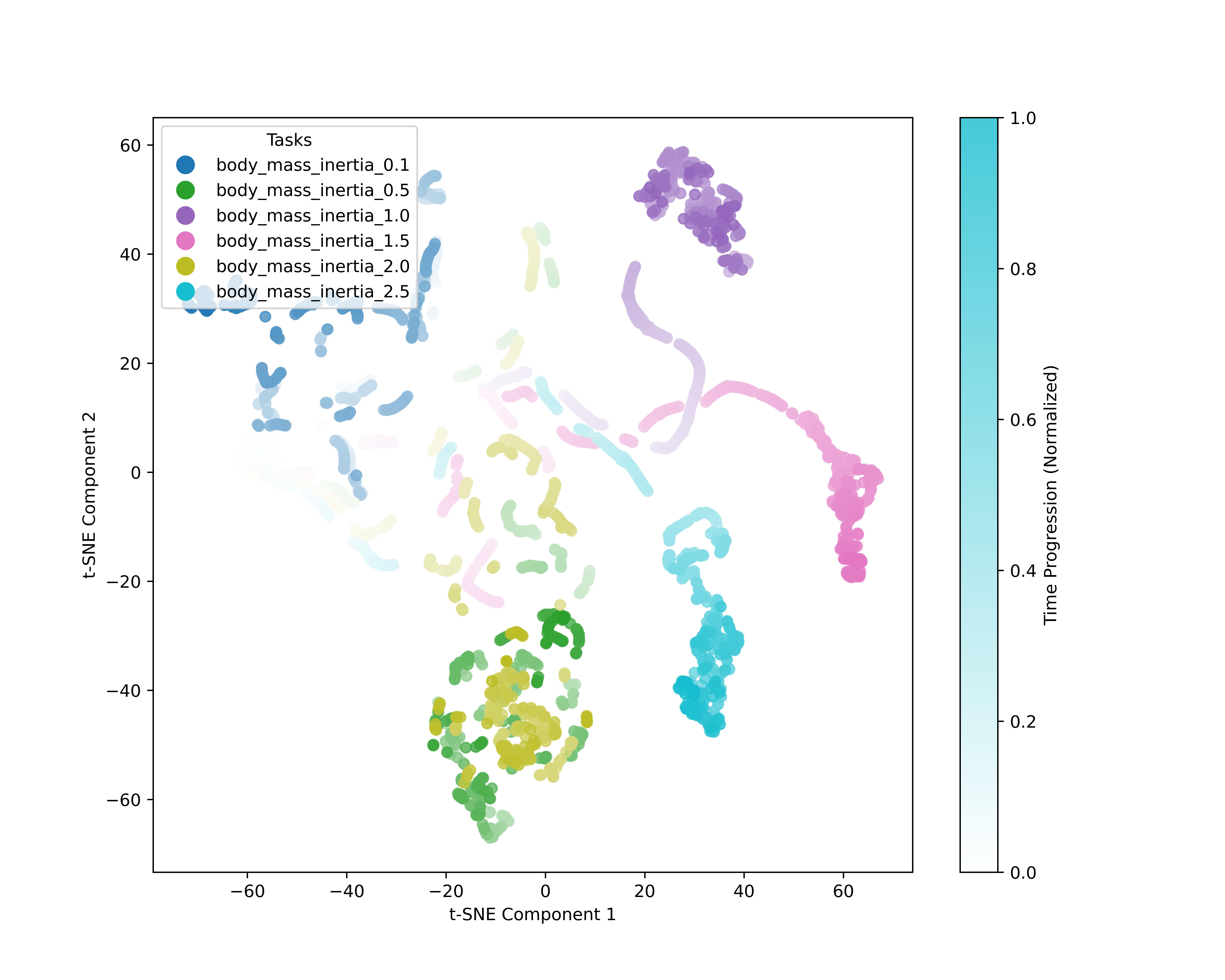}
                \end{minipage}
                \hspace{-18pt}
                \begin{minipage}[b]{0.27\textwidth}
                    \centering
                    \includegraphics[width=\textwidth]{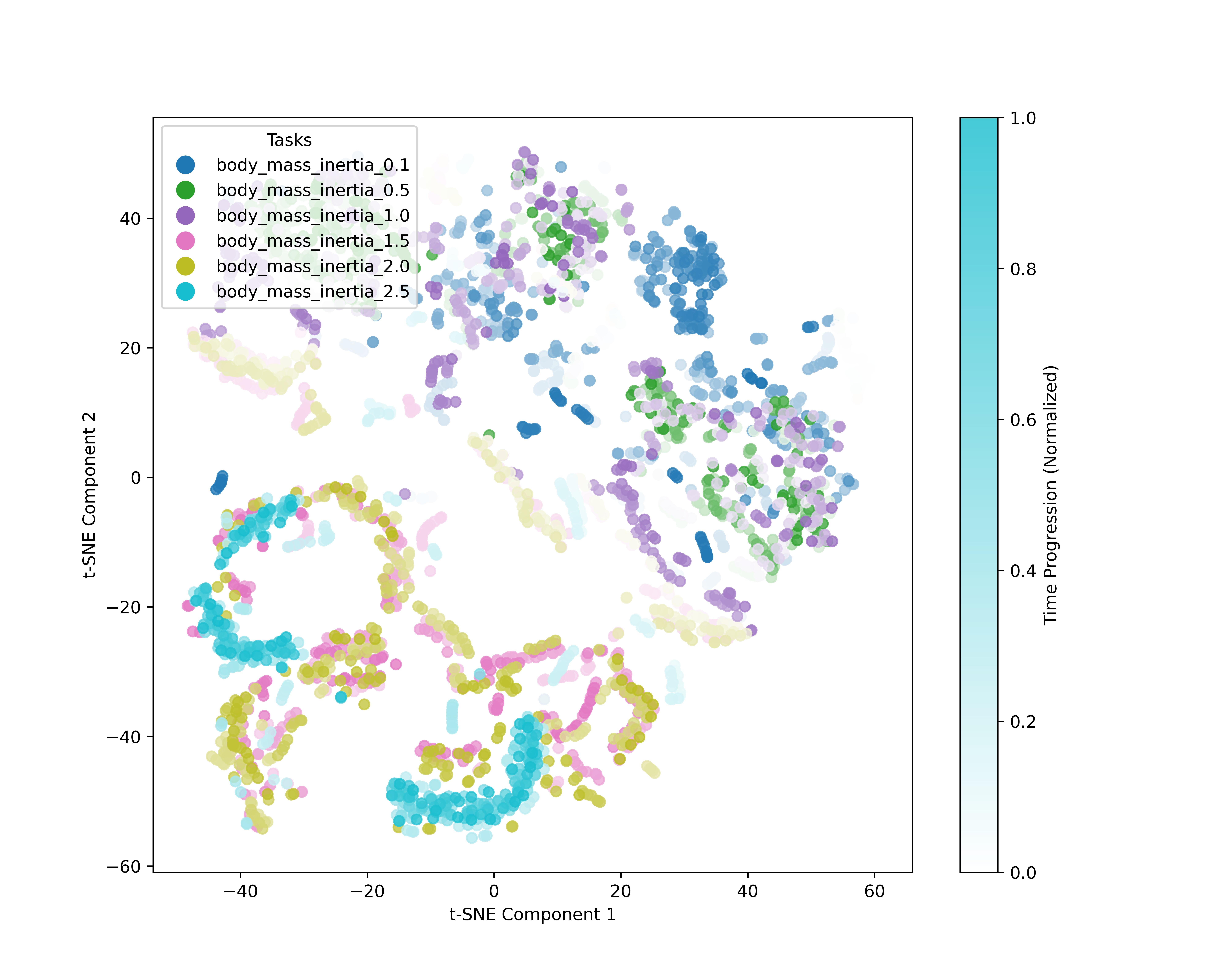}
                \end{minipage}
                \hspace{-18pt}
                \begin{minipage}[b]{0.27\textwidth}
                    \centering
                    \includegraphics[width=\textwidth]{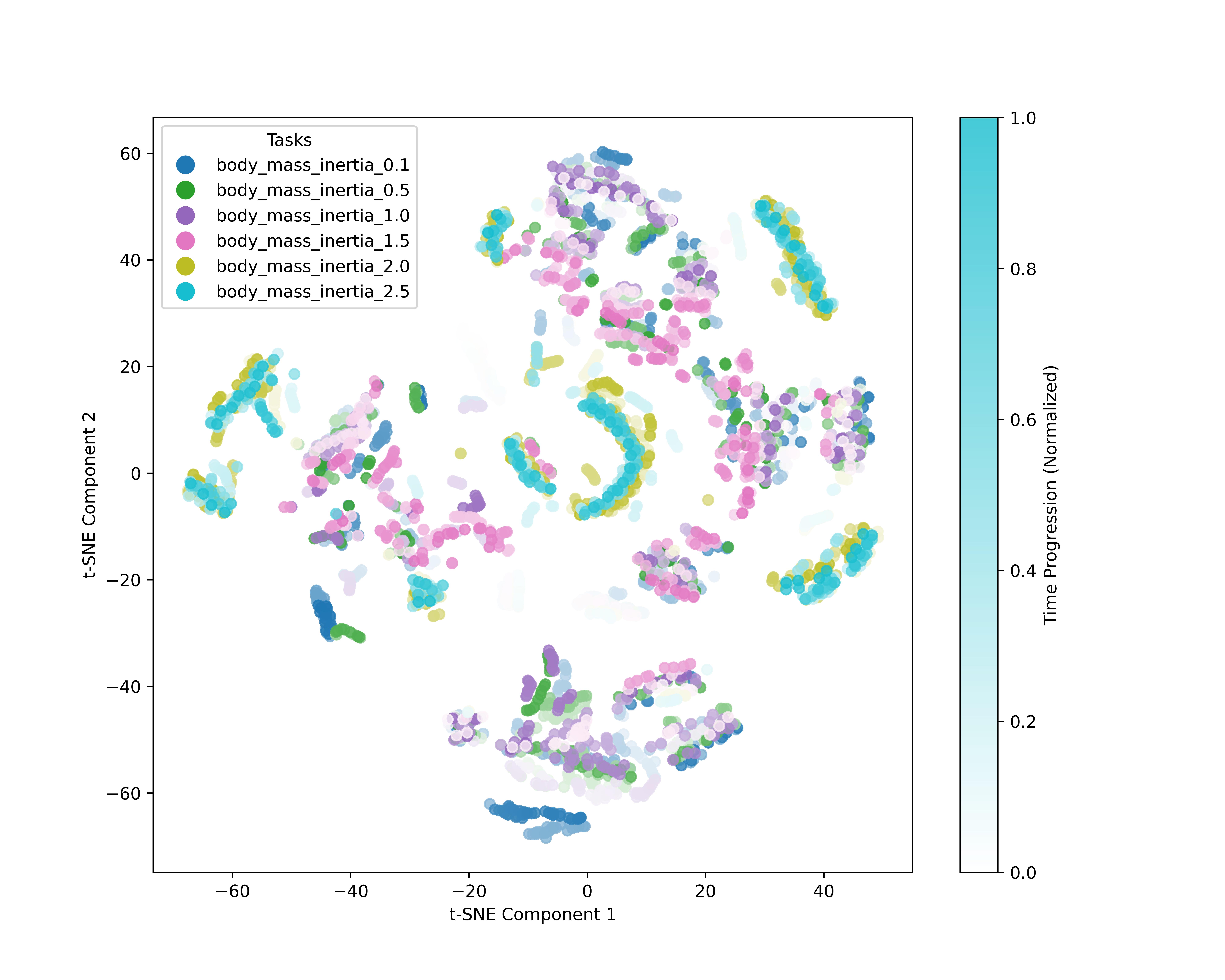}
                \end{minipage}
                \hspace{-18pt}
                \begin{minipage}[b]{0.27\textwidth}
                    \centering
                    \includegraphics[width=\textwidth]{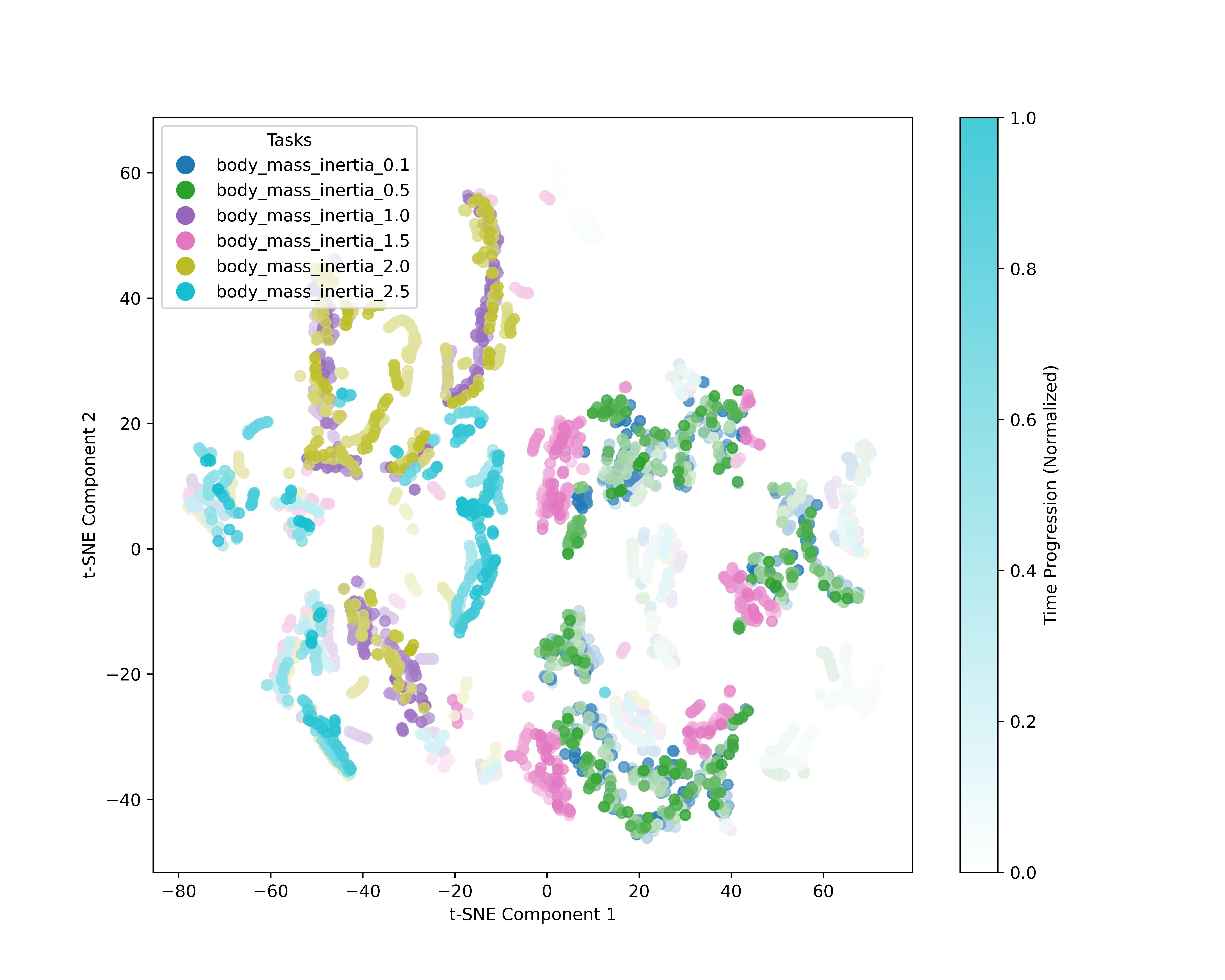}
                \end{minipage}
                \caption{Hopper: body mass and inertia.}
            \end{subfigure}
            \caption{Comprehensive latent state and task space visualizations across various environments experiencing inter-episodic dynamical changes.}
            \label{fig:latent_visualizations_dynamical_changes}
        \end{figure}

        In the visualizations, the vanilla agent—grounded in the POMDP formalism—demonstrates task-dependent latent space structuring under different dynamic changes. Examining these projections along with empirical results from Figure \ref{fig:additional_inter_intra_halfcheetah_walker_dynamical_changes} reveals a positive correlation between the structured latent space and the agent's performance. Notably, conditioning the agent on task representations produces an even more task-specific latent space structure.
        
        Task conditioning appears to provide two main advantages: (1) a more structured latent space enables enhanced data modeling, as task-specific representations minimize interferences, thus improving data reconstruction; and (2) task conditioning helps disambiguate overlapping latent states, especially when a state recurs across tasks. This disambiguation contributes to better data modeling and aids in achieving adaptive behaviors.
        
    \subsubsection{Objective Changes}

        Figure \ref{fig:latent_visualizations_skill_learning} presents 2D projections of latent task and state spaces during evaluations in reward-altered environments, where the objective is to reach various target velocities or master multiple skills.

        \begin{figure}[ht]
            \centering
            \begin{subfigure}{\textwidth}
                \centering
                \begin{minipage}[b]{0.27\textwidth}
                    \centering
                    \scalebox{0.6}{\parbox{1.5\linewidth}{\centering \textbf{Task Inference Dreamer \\ Latent Task Space}}}
                    \vspace{-0.5em}  
                    \includegraphics[width=\textwidth]{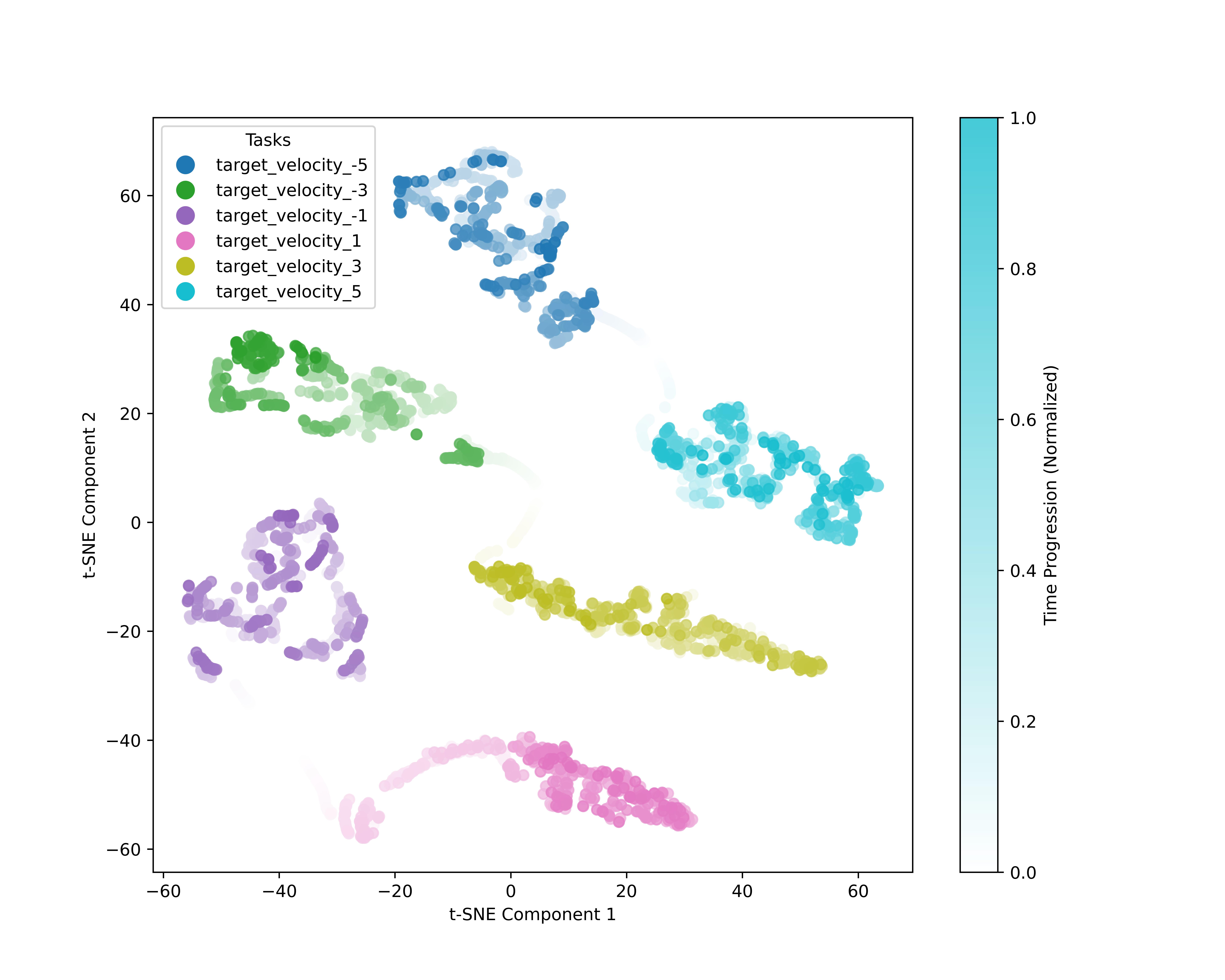}
                \end{minipage}
                \hspace{-18pt}
                \begin{minipage}[b]{0.27\textwidth}
                    \centering
                    \scalebox{0.6}{\parbox{1.5\linewidth}{\centering \textbf{Task Inference Dreamer \\ Latent State Space}}}
                    \vspace{-0.5em}  
                    \includegraphics[width=\textwidth]{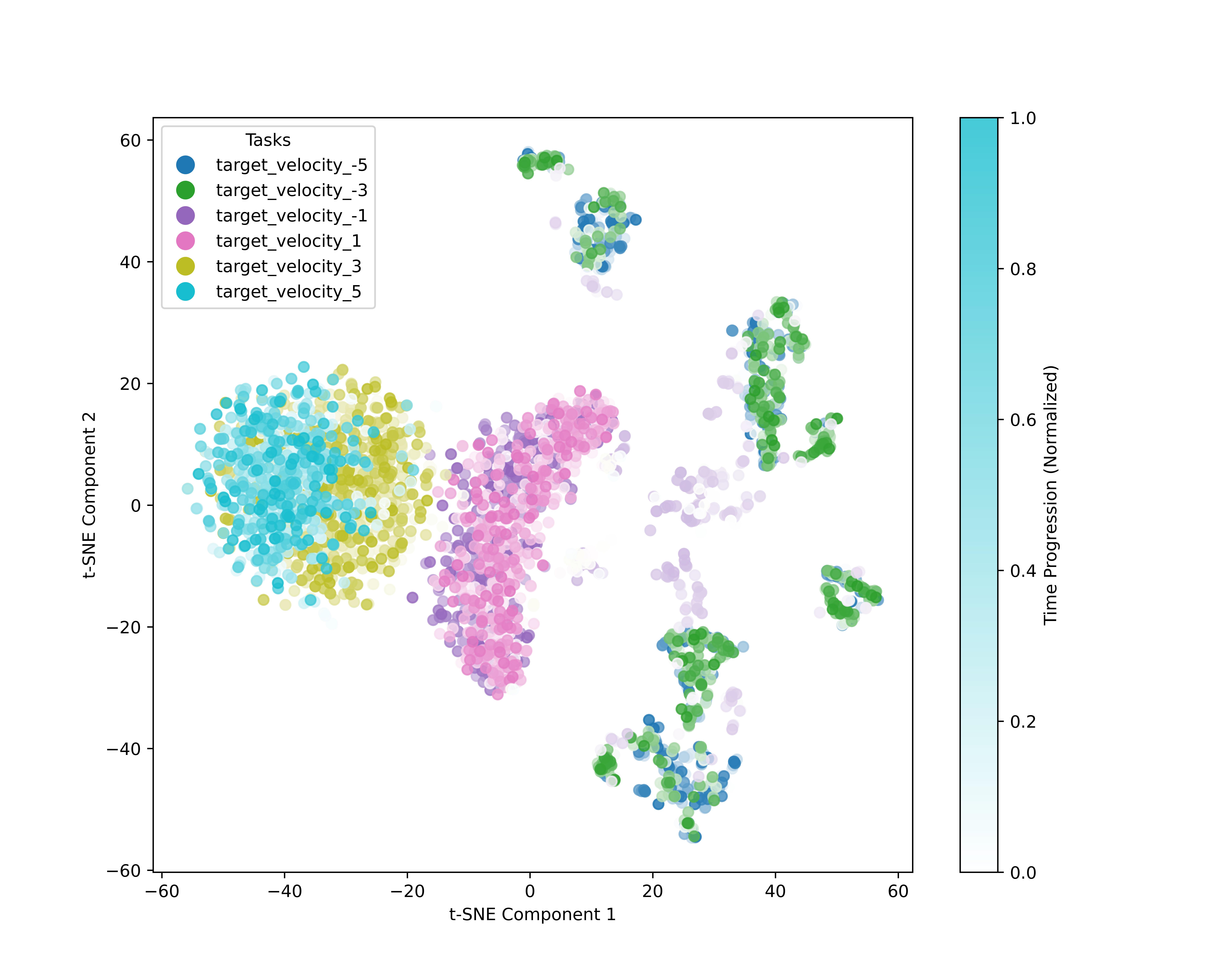}
                \end{minipage}
                \hspace{-18pt}
                \begin{minipage}[b]{0.27\textwidth}
                    \centering
                    \scalebox{0.6}{\parbox{1.5\linewidth}{\centering \textbf{Vanilla Dreamer \\ Latent State Space}}}
                    \vspace{-0.5em}  
                    \includegraphics[width=\textwidth]{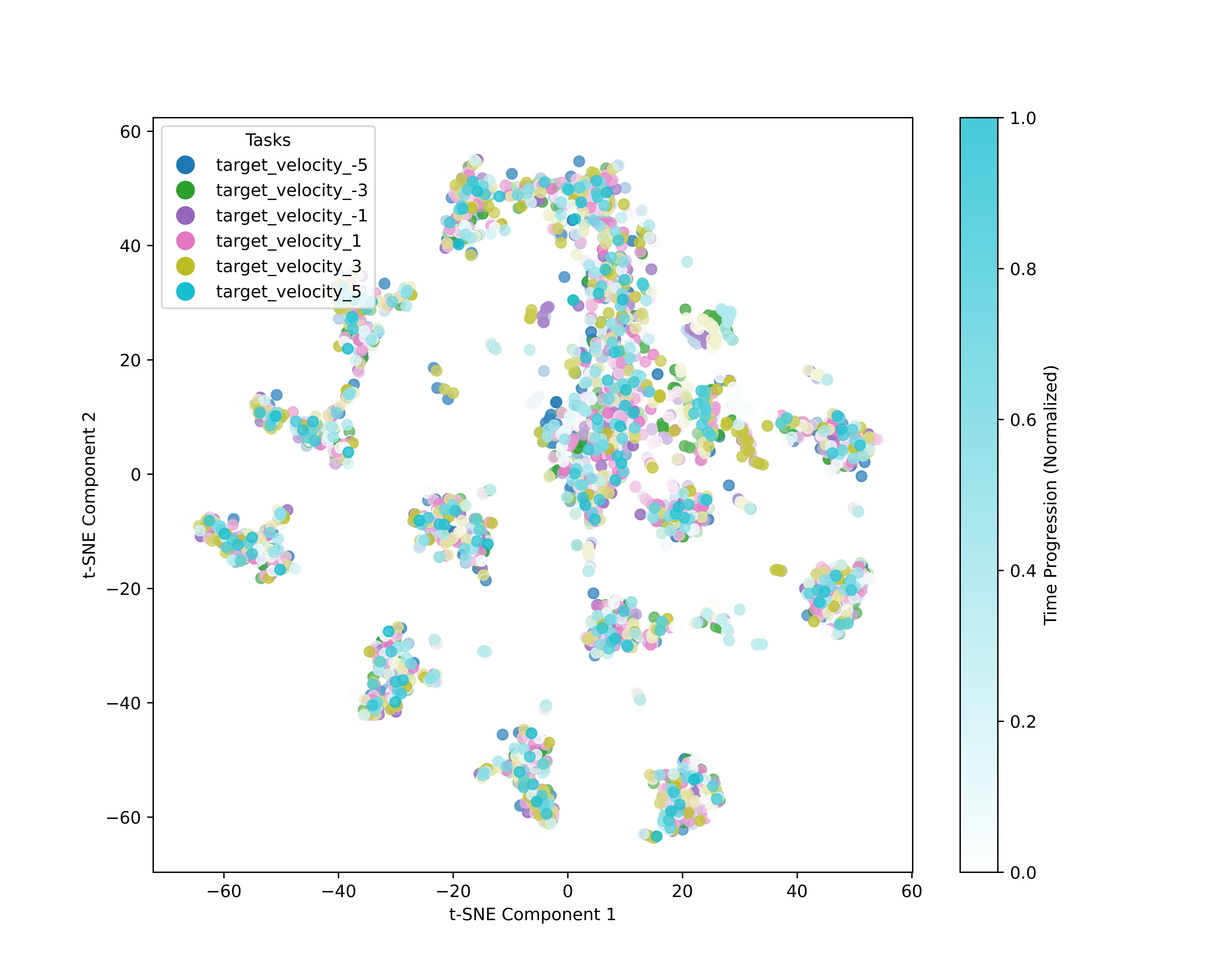}
                \end{minipage}
                \hspace{-18pt}
                \begin{minipage}[b]{0.27\textwidth}
                    \centering
                    \scalebox{0.6}{\parbox{1.5\linewidth}{\centering \textbf{Oracle Dreamer \\ Latent State Space}}}
                    \vspace{-0.5em}  
                    \includegraphics[width=\textwidth]{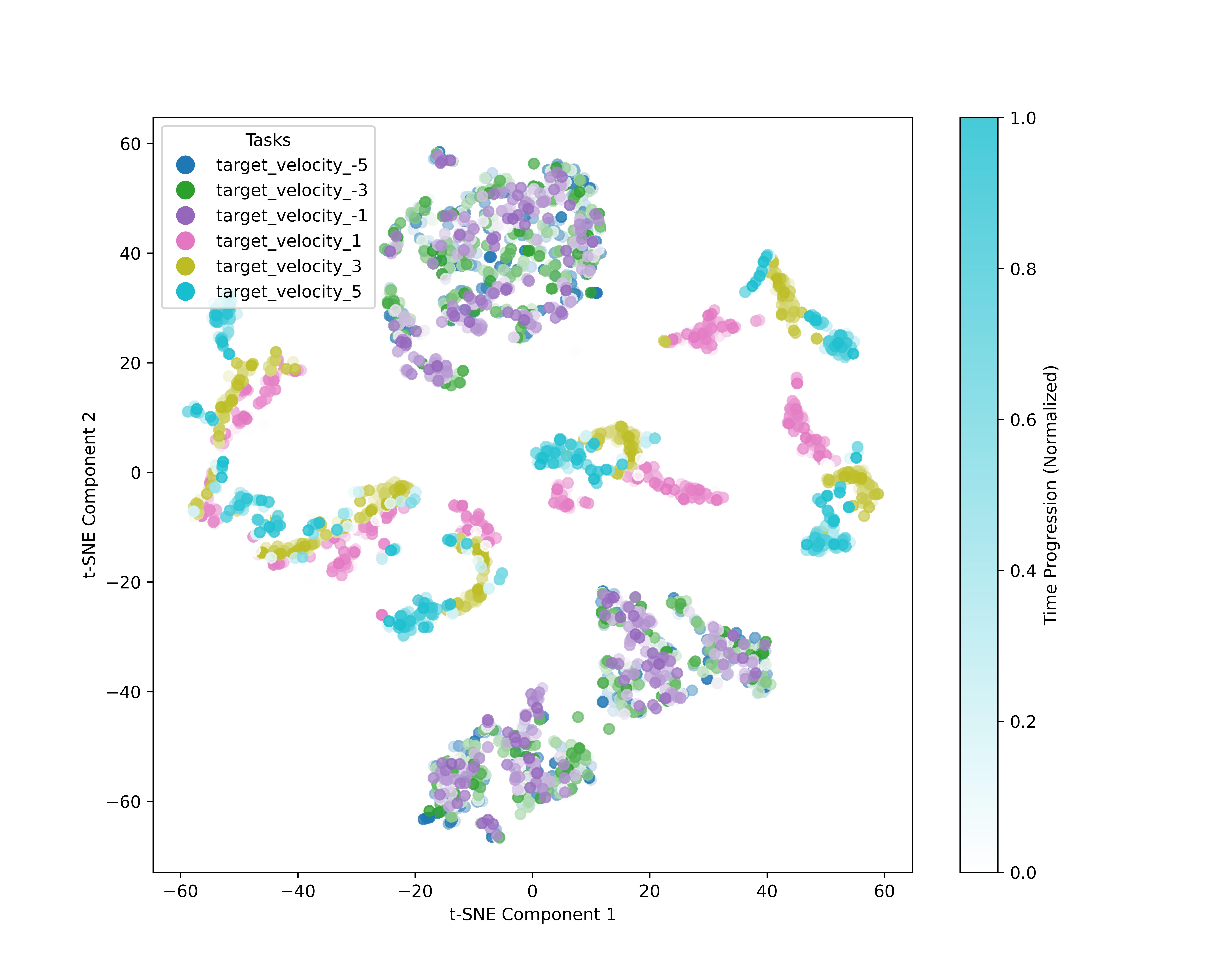}
                \end{minipage}
                \caption{ Cheetah: target velocity.}
            \end{subfigure}
            
            \begin{subfigure}{\textwidth}
                \centering
                \begin{minipage}[b]{0.27\textwidth}
                    \centering
                    \includegraphics[width=\textwidth]{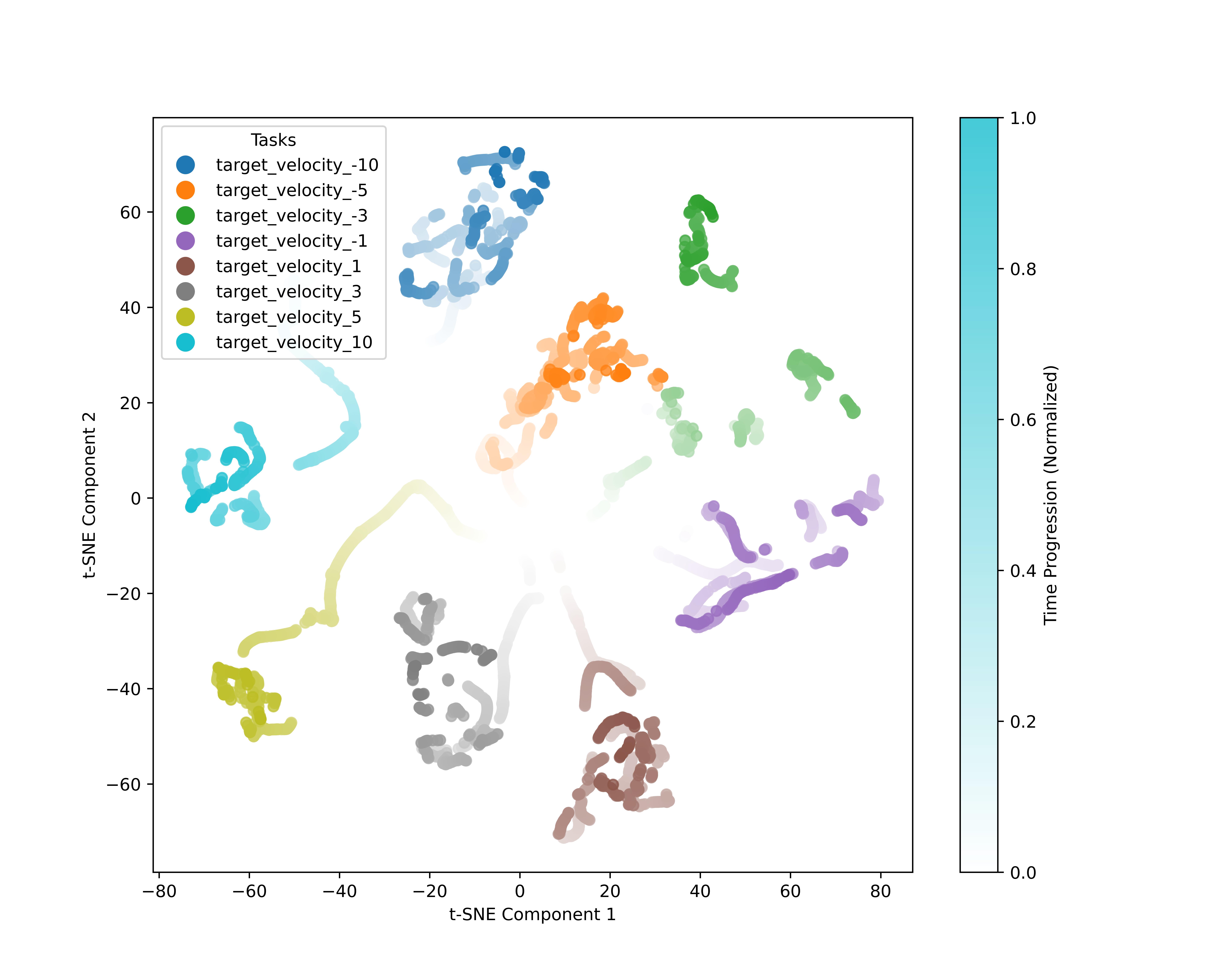}
                \end{minipage}
                \hspace{-18pt}
                \begin{minipage}[b]{0.27\textwidth}
                    \centering
                    \includegraphics[width=\textwidth]{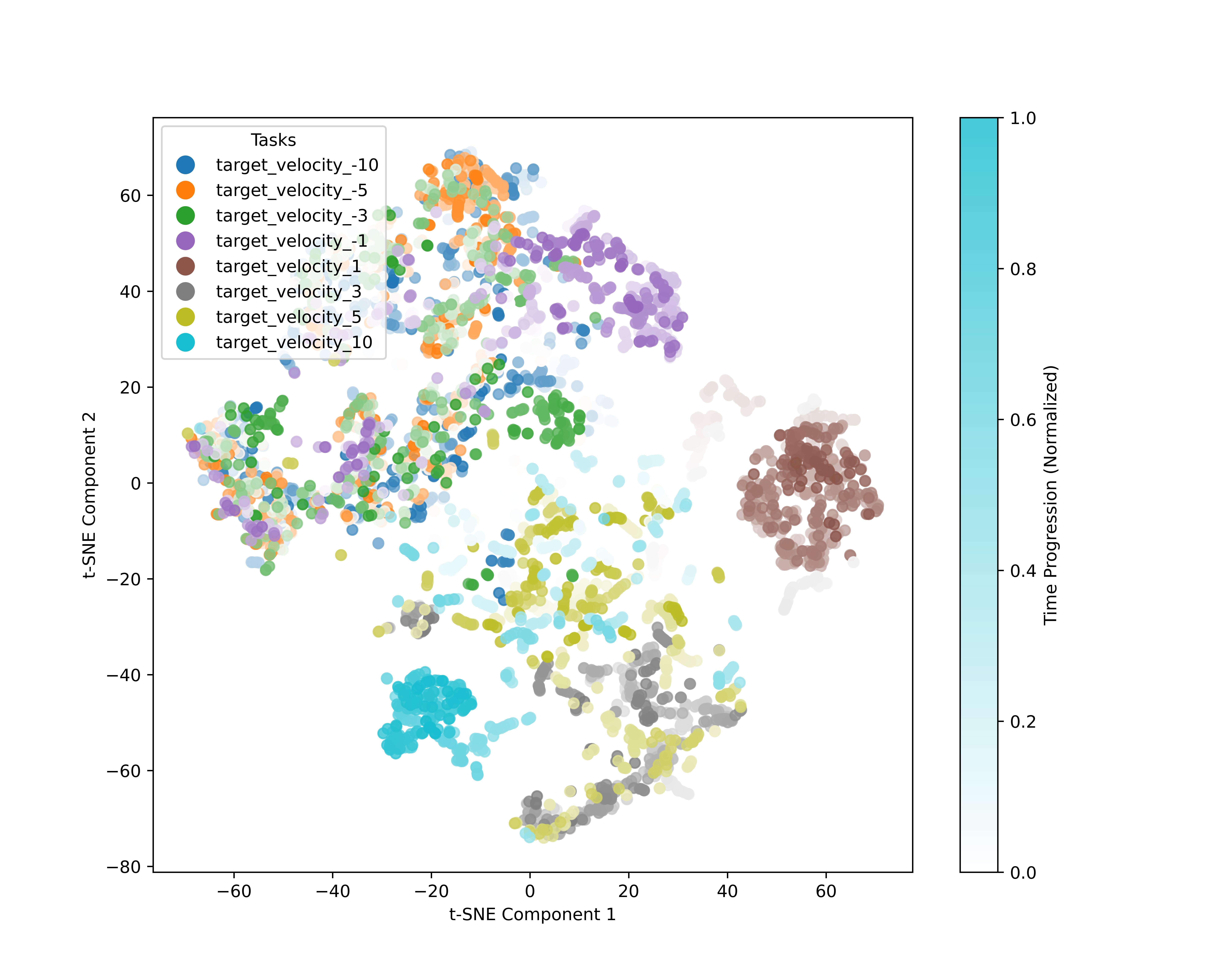}
                \end{minipage}
                \hspace{-18pt}
                \begin{minipage}[b]{0.27\textwidth}
                    \centering
                    \includegraphics[width=\textwidth]{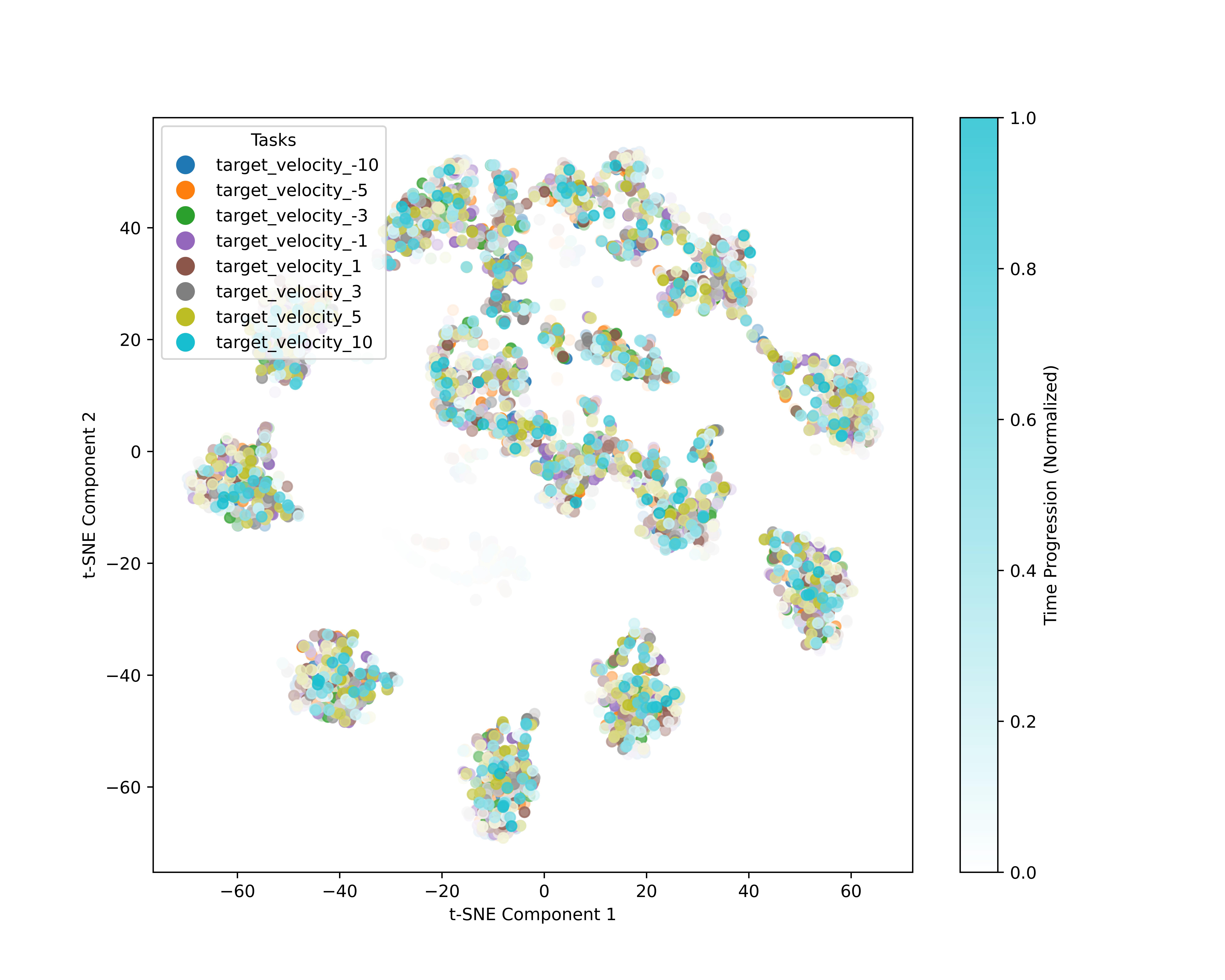}
                \end{minipage}
                \hspace{-18pt}
                \begin{minipage}[b]{0.27\textwidth}
                    \centering
                    \includegraphics[width=\textwidth]{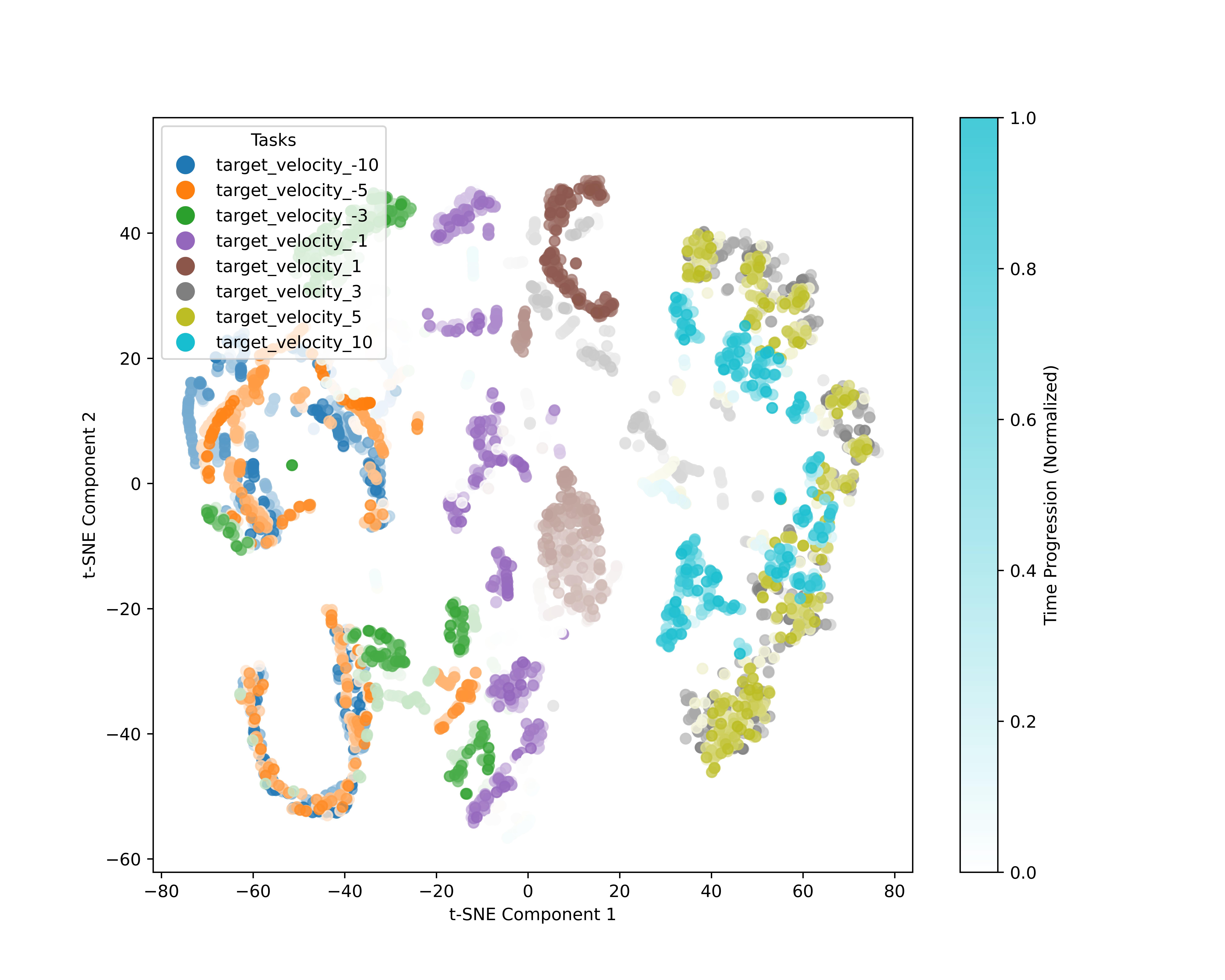}
                \end{minipage}
                \caption{Walker: target velocity.}
            \end{subfigure}
                    
            \begin{subfigure}{\textwidth}
                \centering
                \begin{minipage}[b]{0.27\textwidth}
                    \centering
                    \includegraphics[width=\textwidth]{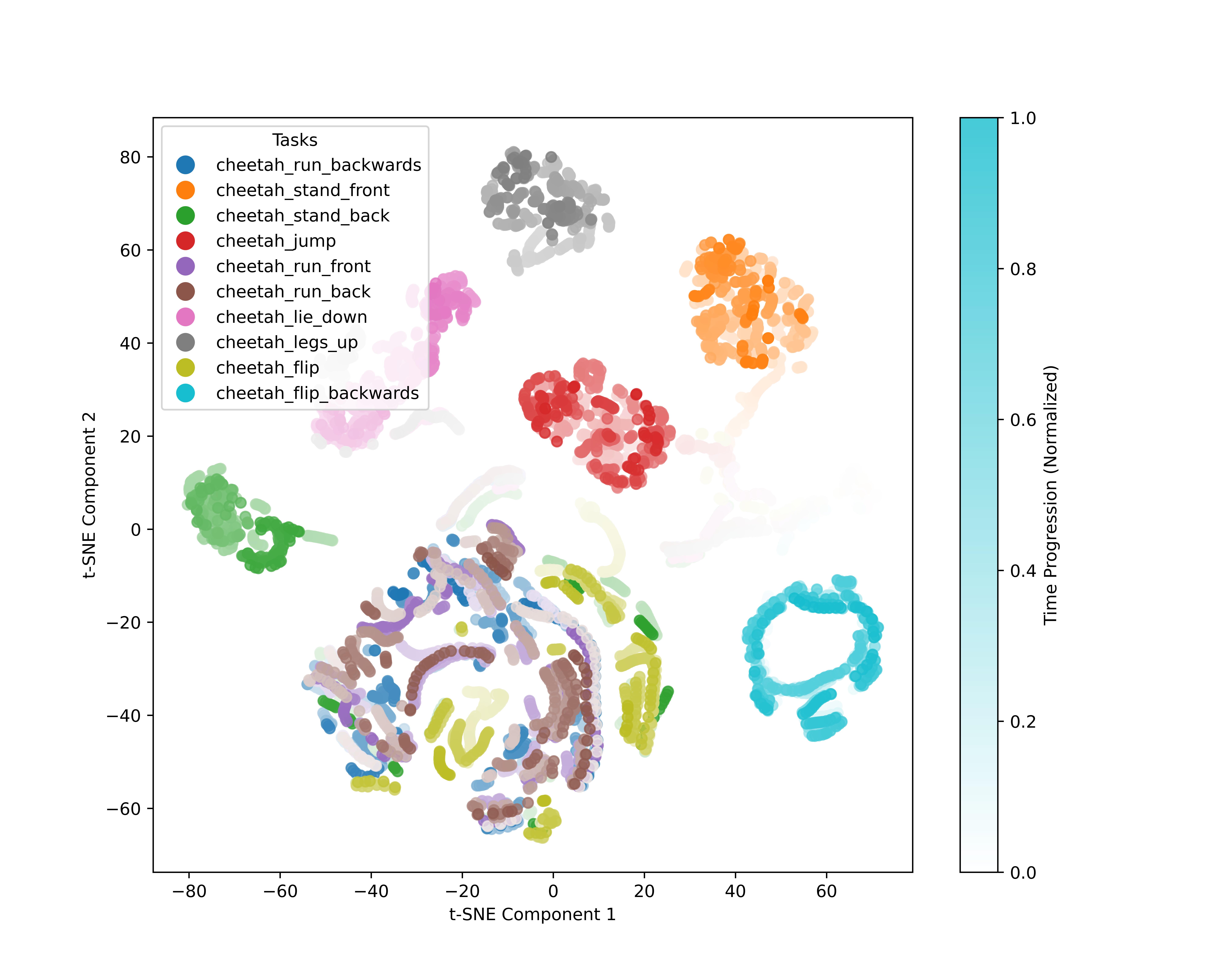}
                \end{minipage}
                \hspace{-18pt}
                \begin{minipage}[b]{0.27\textwidth}
                    \centering
                    \includegraphics[width=\textwidth]{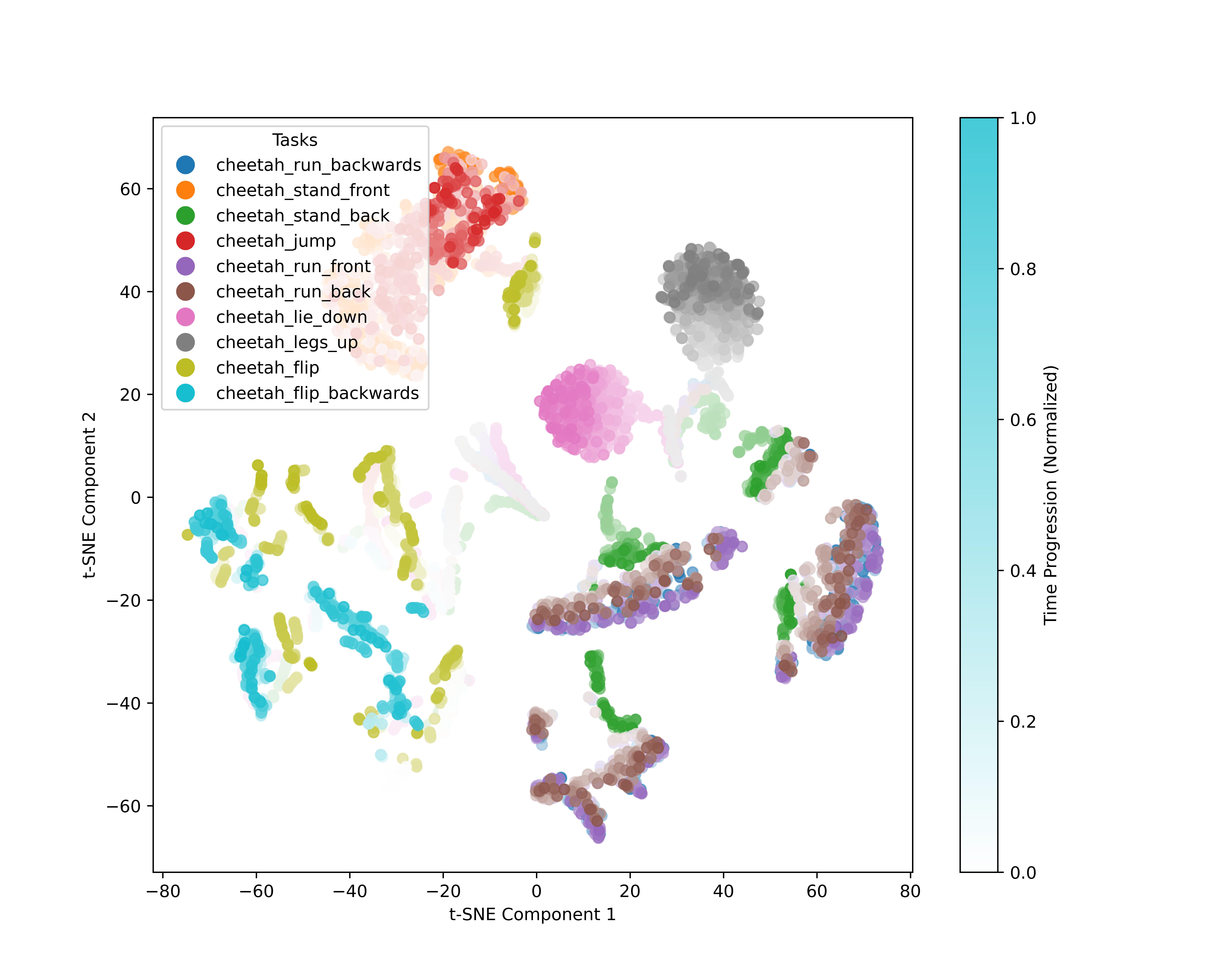}
                \end{minipage}
                \hspace{-18pt}
                \begin{minipage}[b]{0.27\textwidth}
                    \centering
                    \includegraphics[width=\textwidth]{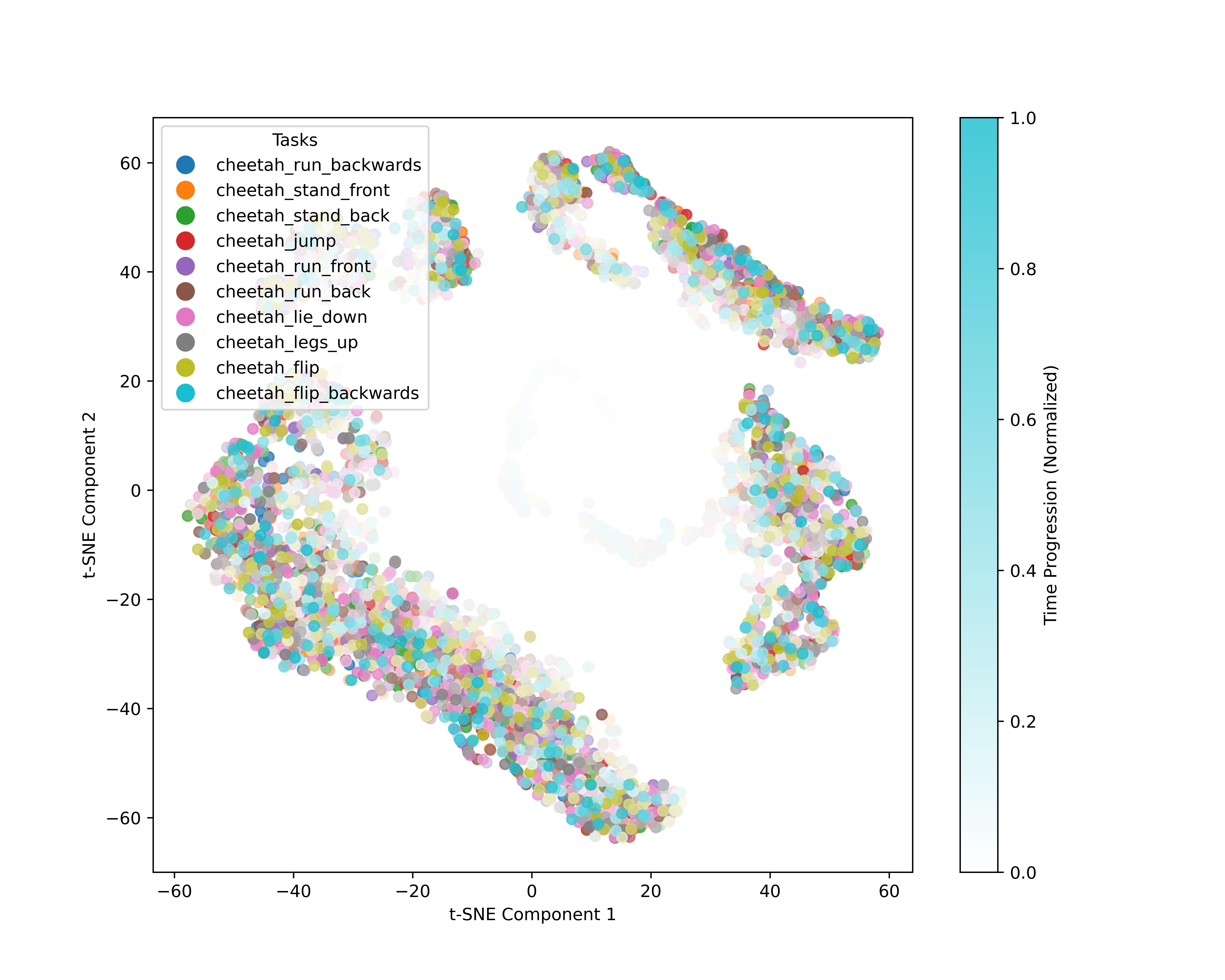}
                \end{minipage}
                \hspace{-18pt}
                \begin{minipage}[b]{0.27\textwidth}
                    \centering
                    \includegraphics[width=\textwidth]{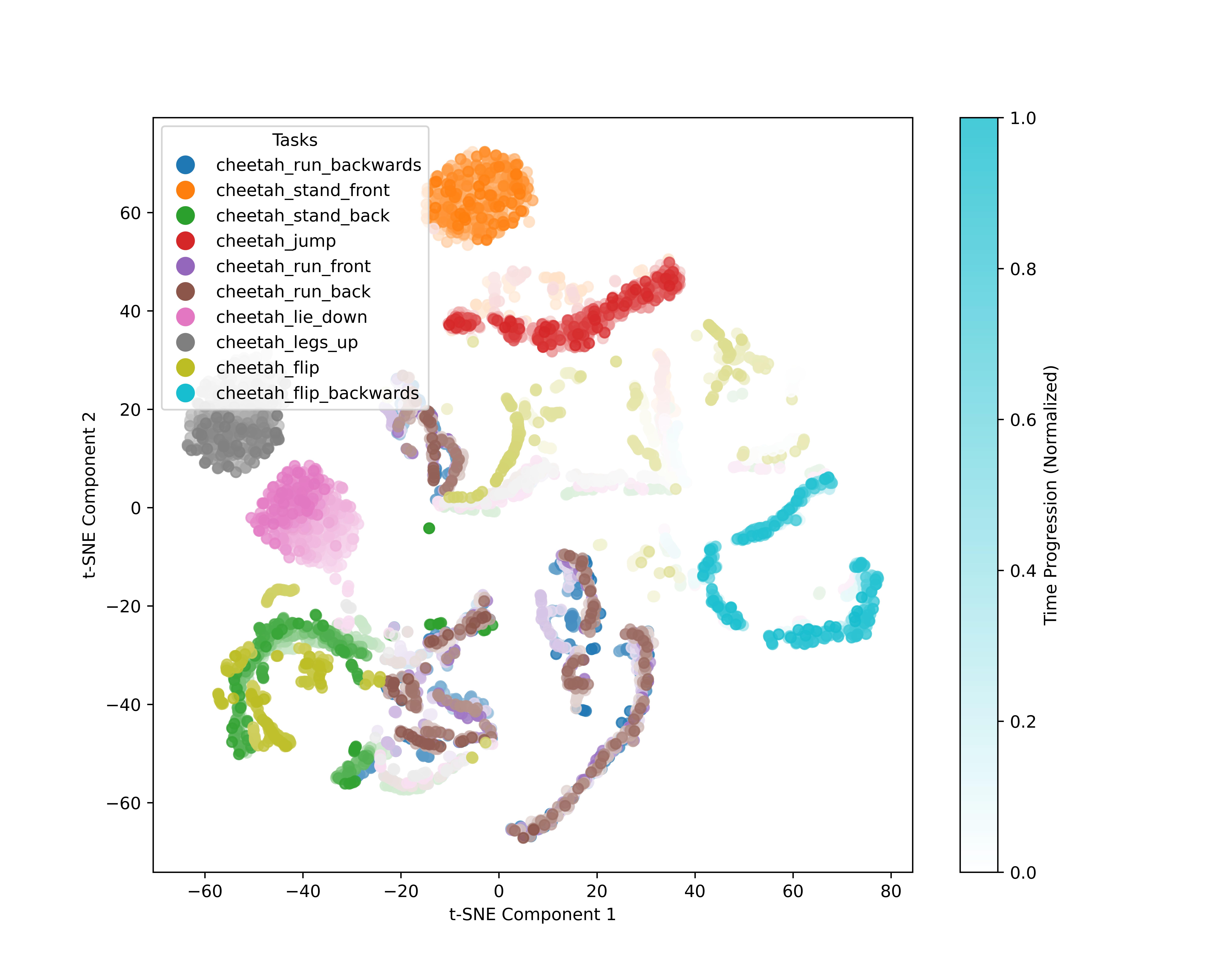}
                \end{minipage}
                \caption{Cheetah: 10 skills.}
            \end{subfigure}

            \begin{subfigure}{\textwidth}
                \centering
                \begin{minipage}[b]{0.27\textwidth}
                    \centering
                    \includegraphics[width=\textwidth]{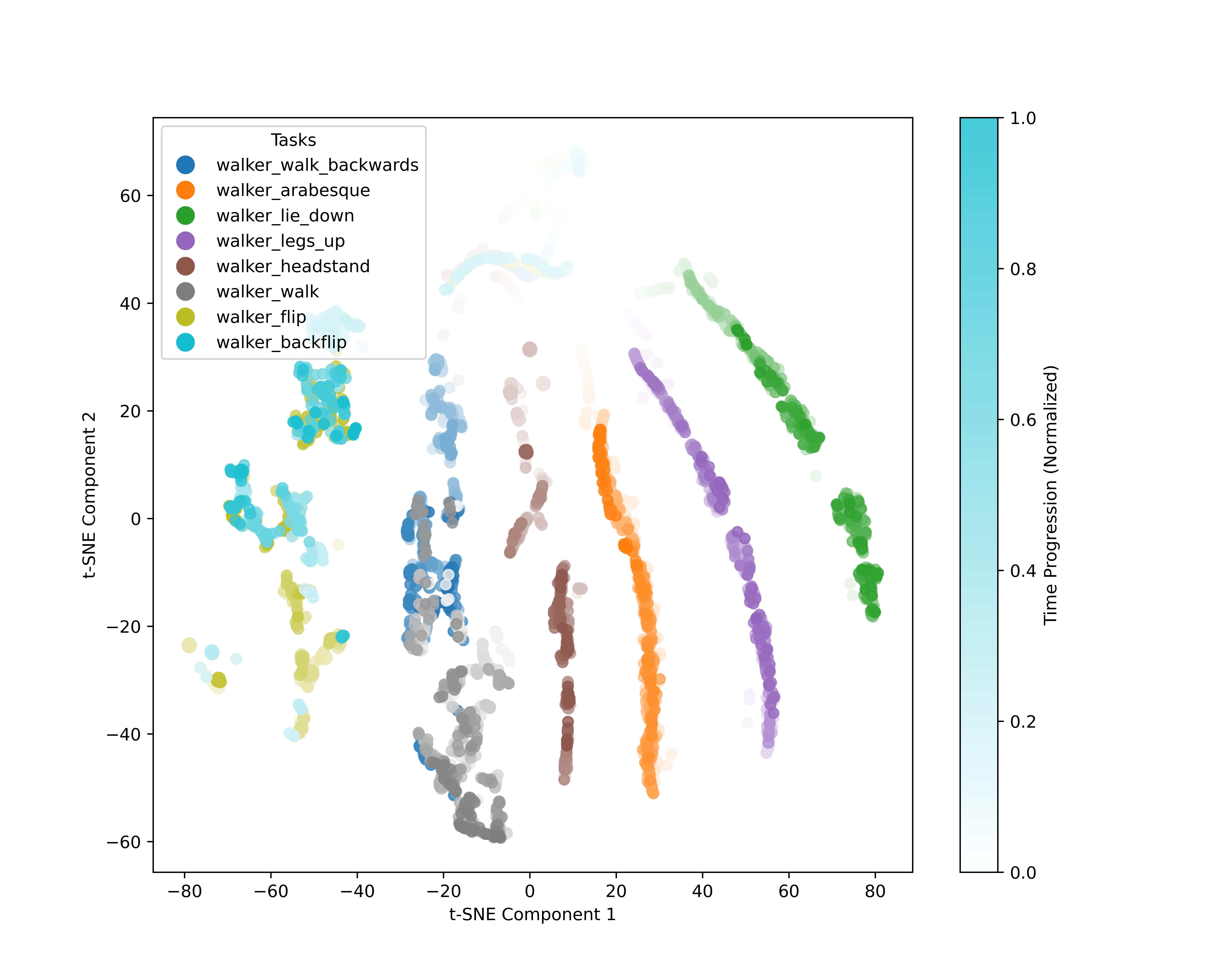}
                \end{minipage}
                \hspace{-18pt}
                \begin{minipage}[b]{0.27\textwidth}
                    \centering
                    \includegraphics[width=\textwidth]{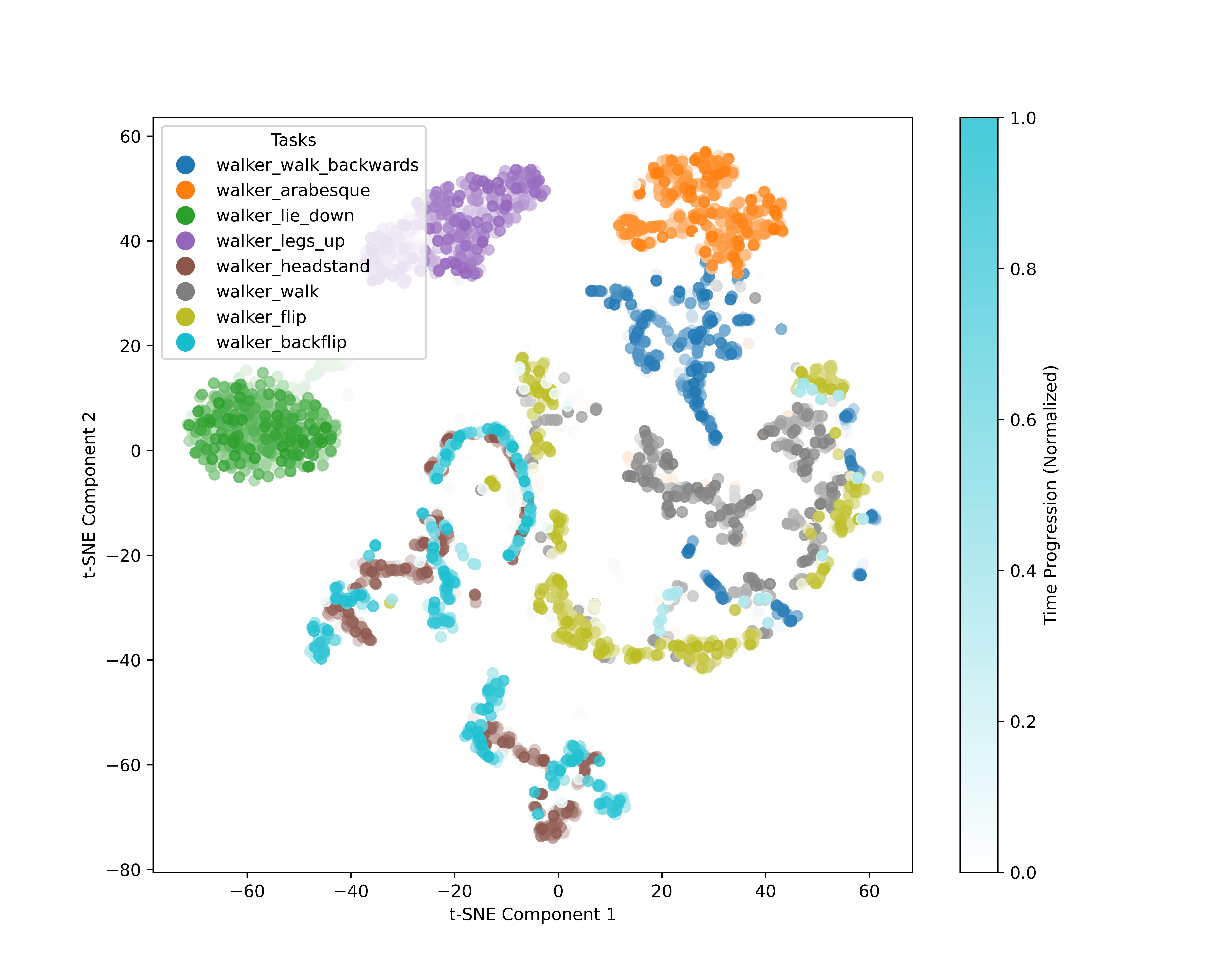}
                \end{minipage}
                \hspace{-18pt}
                \begin{minipage}[b]{0.27\textwidth}
                    \centering
                    \includegraphics[width=\textwidth]{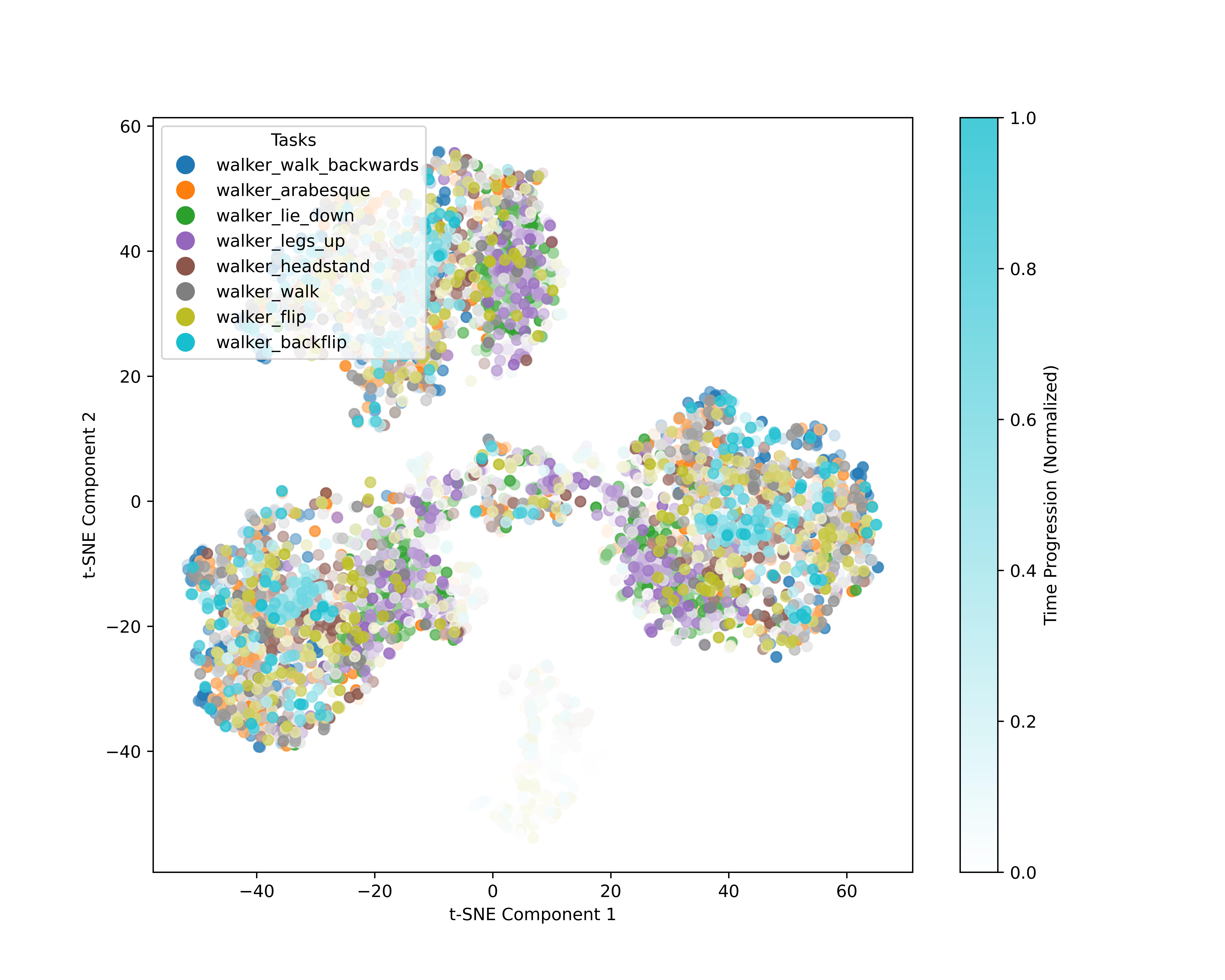}
                \end{minipage}
                \hspace{-18pt}
                \begin{minipage}[b]{0.27\textwidth}
                    \centering
                    \includegraphics[width=\textwidth]{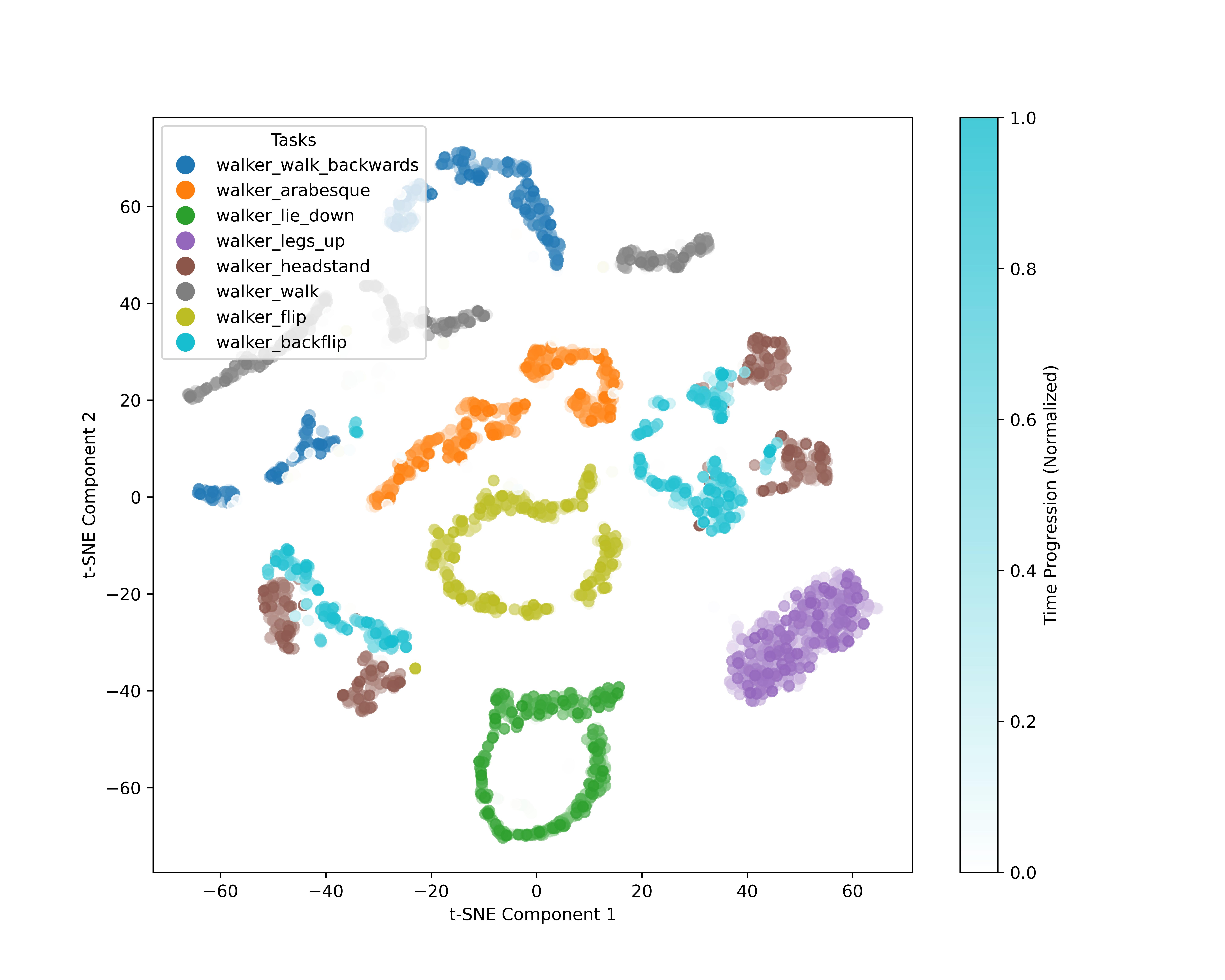}
                \end{minipage}
                \caption{Walker: 8 skills.}
            \end{subfigure}
            \caption{Comprehensive latent state and task space visualizations for agents learning to adapt to varying objectives, including different target velocities or skills.}
            \label{fig:latent_visualizations_skill_learning}
        \end{figure}

    Figure \ref{fig:latent_visualizations_skill_learning} reveals two critical observations. When the reward's structure changes, the vanilla agent’s latent space does not organize itself by task, leading to substantial interference and making it challenging for all agent’s components to infer task information, contributing to suboptimal performance as evidenced in Figure \ref{fig:inter_episodic_dmc_multi_task_objective_changes} for changing rewards. 

    In contrast, conditioning the agent on task representations produces a latent space that is not only more task-aware but also better suited for concurrent multi-task learning. Similarly to the findings in Section \ref{appen:latent_visualizations_dynamical_changes} under dynamics changes, a positive correlation emerges between the structured latent space and the agent’s final performance. However, for the most challenging skill-learning experiments, the task inference approach organizes latent space with some tasks represented jointly, which can hinder unique task identification and contribute to the observed performance gap in skill-learning tasks as shown in Figure \ref{fig:inter_episodic_dmc_multi_task_objective_changes}.

\end{document}

%% file: figures/tikz_graphs.tex
\newcommand{\tikzHiPDreamer}{
\begin{tikzpicture}[
 thick,
node distance = 0.5cm and 0.5cm,
minimum size=0.8cm,
invisible_node/.style={draw=none},
]

\node[det]                                      (h1) {${h_1}$};
\node[det, right=of h1, xshift=0.6cm]           (h2) {${h_2}$};
\node[det, right=of h2, xshift=0.6cm]           (h3) {${h_3}$};
\node[invisible_node, right=of h3, xshift=0.6cm]  (hf) {};

\node[latent, below=of h1]                      (s1) {${s_1}$};
\node[latent, below=of h2]                      (s2) {${s_2}$};
\node[latent, below=of h3]                      (s3) {${s_3}$};

\node[obs, below=of s1]          (o1) {${o_1, r_1}$};
\node[obs, below=of s2]          (o2) {${o_2, r_2}$};
\node[obs, below=of s3]          (o3) {${o_3, r_3}$};

\node[latent, above=of h1, xshift=1.0cm, yshift=-0.5cm]                      (a1) {${a_1}$};
\node[latent, above=of h2, xshift=0.7cm, yshift=-0.5cm]                      (a2) {${a_2}$};
\node[latent, above=of h3, xshift=1.0cm, yshift=-0.5cm]                      (a3) {${a_3}$};

\node[latent, draw=red, above=of h2, yshift=2.0cm]                   (l)  {${l}$};
\node[obs, draw=red, above=of l]                    (c)  {$\mathcal{C}_{{l}}$};

\edge{h1}{h2}
\edge{h1}{s1}
\edge{h2}{s2}
\edge{h2}{s2}
\edge{h2}{h3}
\edge{h3}{s3}
\edge{h3}{hf}

\edge{s1}{h2}
\edge{s1}{o1}

\edge{s2}{h3}
\edge{s2}{o2}

\edge{s3}{hf}
\edge{s3}{o3}

\path[->] (l) edge[red, thick, bend right=35] (h1);
\path[->] (l) edge[red, thick, bend left=35] (h3);
\path[->] (l) edge[red, thick, bend left=40] (hf);
\path[->] (l) edge[red, thick, bend right=60] (o1);
\path[->] (l) edge[red, thick, bend right=20, in=210] (o2);
\path[->] (l) edge[red, thick, bend left=15, out=30, in=195] (o3);

\path[->] (o1) edge[dotted, black, thick, bend right=30] (s1);
\path[->] (o2) edge[dotted, black, thick, bend right=30] (s2);
\path[->] (o3) edge[dotted, black, thick, bend right=30] (s3);
\path[->] (h1) edge[dotted, black, thick, bend left=30] (s1);
\path[->] (h2) edge[dotted, black, thick, bend left=30] (s2);
\path[->] (h3) edge[dotted, black, thick, bend left=30] (s3);
\path[->] (h1) edge[black, thick, bend right=20] (o1);
\path[->] (h2) edge[black, thick, bend right=20] (o2);
\path[->] (h3) edge[black, thick, bend right=20] (o3);

\edge{a1}{h2}
\edge{a2}{h3}
\edge{a3}{hf}

\edge[red]{l}{h2}
\edge[red]{l}{c}
\edge{h3}{hf}

\end{tikzpicture}
 }

\newcommand{\tikzLatentTaskInference}{
\begin{tikzpicture}[
very thick,
node distance = 0.5cm and 0.5cm,
invisible_node/.style={draw=none},
context_node/.style={circle, 
                     draw=black,
                     fill=gray!20,
                     very thick,
                     minimum size=1cm, 
                     },
r_node/.style={circle, 
               draw=black,
               fill=gray!20,
               very thick,
               minimum size=1cm,
               },
l_node/.style={circle, 
               draw=black,
               very thick,
               minimum size=1cm,
               },
o_node/.style={circle, 
               draw=black,
               fill=gray!20,
               very thick,
               minimum size=1cm,
               },
a_node/.style={circle, 
               draw=black,
               fill=gray!20,
               very thick,
               minimum size=1cm,
               },
]

\node[context_node]         (x)                                                       { \Large ${x^l_n}$};     
\node[context_node]   (a_t)    [below = of x, xshift=-0.75cm]                                   {\Large ${a}_{n}$};
\node[context_node]   (o_t)    [left = of a_t]                                  {\Large ${o}_{n}$};
\node[r_node] (r_t) [right = of a_t] {\Large ${r}_{n}$};
\node[invisible_node] (inv)    [above = of r_t, xshift=1cm, yshift=0.8cm]                {$n=\{1..N\}$};
\node[a_node]   (o_tp1)  [right = of r_t]                                 {\Large ${o}_n^{'}$};
\node[l_node] (l) [above = of x, yshift=1.0cm]                                {\Large ${l}$};

\draw[-{Latex[open]}] (o_t.north east) -- (x.south west);
\draw[-{Latex[open]}] (a_t.north) -- (x.south);
\draw[-{Latex[open]}] (r_t.north) -- (x.south);
\draw[-{Latex[open]}] (o_tp1.north west) -- (x.south east);

\draw[draw=black, rounded corners] (-3, 1.0) rectangle ++(6.0,-3.3);

\edge{l}{x}

\end{tikzpicture}
}